%% file: main.tex
\documentclass[10pt,twocolumn,letterpaper]{article}

\usepackage{cvpr}              % To produce the CAMERA-READY version
\input{preamble}
\definecolor{cvprblue}{rgb}{0.21,0.49,0.74}
\usepackage[pagebackref,breaklinks,colorlinks,allcolors=cvprblue]{hyperref}

\usepackage{wrapfig}
\usepackage{booktabs}
\usepackage{array}
\usepackage[table]{xcolor}
\usepackage{adjustbox}
\usepackage{colortbl}
\usepackage{enumitem}

\usepackage{caption}               % subplots
\usepackage{subcaption}
\usepackage{graphicx}
\usepackage{threeparttable}        % table note
\usepackage{wrapfig}               % Partial Table
\usepackage{multirow}
\usepackage{xcolor}
\usepackage{bbding}                % XSolidBrush
\usepackage{lipsum}                % Dummy Text 
\usepackage[cjk]{kotex}            % TeX Liver (Korean)
\usepackage{amsmath}               % center function
\usepackage{graphicx}              % left vector
\usepackage{accents}               % left vector hat
\usepackage{multirow}              % table
\usepackage{mathtools}             % math bracket
\usepackage{footnote}
\usepackage{tablefootnote}
\usepackage{subfiles}
\usepackage{colortbl}              % CHECK
\usepackage{wrapfig}               % row figs
\usepackage{amssymb}               % 수식
\usepackage{comment}               % comment
\usepackage[capitalize]{cleveref}              % cref
\usepackage{pifont}
\usepackage{makecell}
\usepackage[stable,multiple]{footmisc}
\usepackage{bold-extra}

\usepackage{float}
\usepackage{hhline}

\usepackage[dvipsnames]{xcolor}
\definecolor{darkergreen}{RGB}{21, 152, 56}
\definecolor{ForestGreen}{RGB}{21, 152, 56}
\definecolor{red2}{RGB}{252, 54, 65}
\definecolor{Gray}{gray}{0.6}
\definecolor{LavenderBlush}{rgb}{1.0, 0.94, 0.96}
\definecolor{lightgray}{gray}{0.93} 

\newcommand{\yesmark}{\textcolor{darkergreen}{\ding{52}}}
\newcommand{\nomark}{\textcolor{red2}{\ding{56}}}

\definecolor{color1}{HTML}{006EB8}
\usepackage{url} 
\definecolor{topThreeBlue}{HTML}{08306B} % deep-blue  (Top-3)
\definecolor{topOneBlue}{HTML}{90C5E0}   % sky-blue   (Top-1)
\definecolor{Baseline}{HTML}{6EACDA} % paleblue
\definecolor{Ours}{HTML}{fa8787}
\definecolor{mygray}{gray}{0.65}
\def\paperID{} % *** Enter the Paper ID here
\def\confName{CVPR}
\def\confYear{2026}

\title{Real-Time Long Horizon Air Quality Forecasting\\
via Group-Relative Policy Optimization}

\author{
    Inha Kang\textsuperscript{1}, 
    Eunki Kim\textsuperscript{1}, 
    Wonjeong Ryu\textsuperscript{1}, 
    Jaeyo Shin\textsuperscript{1}, 
    Seungjun Yu\textsuperscript{1} \\
    Yoon-Hee Kang\textsuperscript{2}, 
    Seongeun Jeong\textsuperscript{2}, 
    Eunhye Kim\textsuperscript{3}, 
    Soontae Kim\textsuperscript{2}, 
    Hyunjung Shim\textsuperscript{1} \\[0.5em]
    \textsuperscript{1}KAIST AI \qquad 
    \textsuperscript{2}Ajou University \qquad 
    \textsuperscript{3}Kunsan University \\
    {\tt\small \{rkswlsj13, eunkikim, petac, jaeyo\_shin, seungjunyu, kateshim\}@kaist.ac.kr} \\
    {\tt\small \{ykang, atmos1214, soontaekim\}@ajou.ac.kr \qquad ekim@kunsan.ac.kr}
}

\begin{document}
\maketitle
\input{sec/0_abstract}    
\input{sec/1_intro}

\input{sec/2_related}
\input{sec/3_dataset}

\input{sec/4_method}
\input{sec/5_exp}
\input{sec/6_conc}

%\newpage
{
    \small
    \bibliographystyle{ieeenat_fullname}
    \bibliography{main}
}

\input{sec/X_suppl}

\end{document}

%% file: sec/0_abstract.tex
\begin{abstract}

Accurate long horizon forecasting of particulate matter (PM) concentration fields is essential for operational public health decisions. However, achieving reliable forecasts remains challenging in regions with complex terrain and strong atmospheric dynamics such as East Asia. While foundation models such as Aurora offer global generality, they often miss region-specific dynamics and rely on non–real-time inputs, limiting their practical utility for localized warning systems.
To address this gap, we construct and release the real-world observations and high-resolution CMAQ--OBS dataset for East Asia, reducing regional error by 59.5\% and enabling real-time 48--120 hour forecasts critical for public health alerts.
However, standard point-wise objectives cannot reflect asymmetric operational costs, where false alarms deteriorate public trust while missed severe events endanger populations. This cost mismatch causes SFT models to over-predict and yield high False Alarm Rates. We introduce Group-Relative Policy Optimization (GRPO) with class-wise rewards and curriculum rollout to align predictions with operational priorities.
Experimental results demonstrate that our framework significantly improves the reliability of the forecast. Compared to the SFT-only baseline, our model reduces the False Alarm Rate by 47.3\% while achieving a competitive F1-score, proving its effectiveness for practical, real-world air quality forecasting systems on long lead time scenarios. Code and dataset are publicly available at \url{https://github.com/kaist-cvml/FAKER-Air}.

\end{abstract}

%% file: sec/1_intro.tex
\section{Introduction}
\label{sec:intro}

\begin{figure}[t!]
  \centering
  \vspace{-2em}
    \includegraphics[width=0.9\linewidth]{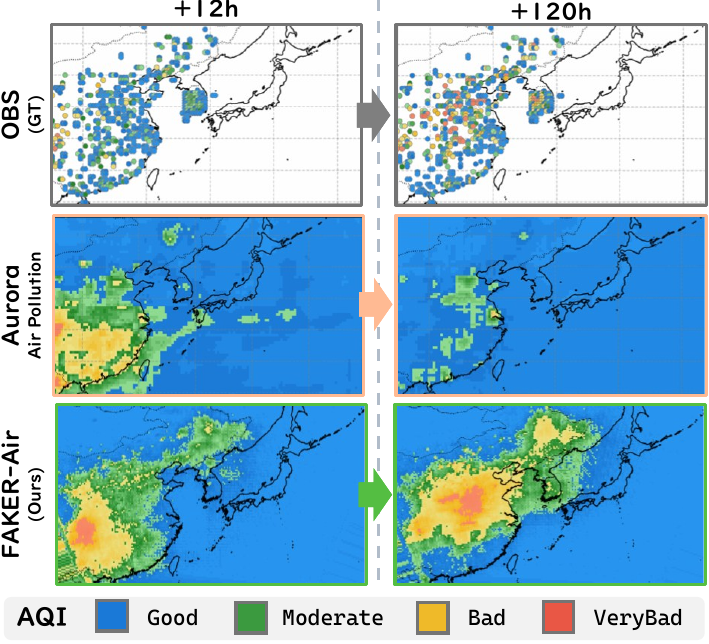}
    \vspace{-0.5em}
  \caption{\textbf{Illustration of PM\textsubscript{2.5} predictions.}
Our method effectively captures the dynamic temporal variations in PM concentration over time, whereas Aurora~\citep{bodnar2025foundation} fails to reflect such changes.}
  \label{fig:aurora_vs_ours}
  \vspace{-1.5em}
\end{figure}

Atmospheric particulate matter, such as PM\textsubscript{2.5} and PM\textsubscript{10}, poses serious public health risks by aggravating respiratory and cardiovascular diseases, and disrupting urban transportation, industrial operations, and power demand. Accurate long lead time forecasts of PM concentrations (48--120 hours) are essential for issuing public alerts, enforcing emission controls, and protecting vulnerable populations~\cite{LEE2023131678,Reichstein2025,Newell2017,Anderson2012-eb,acp-20-2319-2020}. However, forecast accuracy deteriorates rapidly beyond short horizons due to complex spatio-temporal interactions among meteorology, emissions, and terrain~\citep{Zaini2022}.  This degradation is particularly severe in East Asia, where global models increasingly fail to detect severe pollutions~\citep{Wang2021,Wei2023-rh,modelingAirPollution}, as in~\Cref{fig:aurora_vs_ours}. Our objective is to achieve stable long horizon PM forecasting with reliable air quality classification suitable for real-time operational warning systems at the target region.

Recent advances in deep learning have produced data-driven weather and air quality models that rival numerical weather prediction (NWP) systems in accuracy and efficiency~\cite{hurrell2013community,shi2025deep,du2019deep}. 
Foundation models such as Aurora, GraphCast, and Pangu-Weather learn global atmospheric dynamics from large-scale reanalyses including ERA5, GFS, and CAMS~\cite{bodnar2025foundation,lam2023learning,bi2023accurate,hersbach_era5_2018,noaa_gfs_2024,ecmwf_cams_2024}. 
Notably, Aurora is the only open-source PM forecasting model, making it our baseline.
At the regional scale, hybrid convolutional-transformer architectures capture spatio-temporal pollutant transport for PM forecasting~\cite{zhang2023pmMultistep,gul2022pmMultistep}.

However, when evaluated against ground-based observations (OBS) in East Asia, reanalysis products exhibit large systematic biases and limited operational usability. 
For instance, CAMS~\cite{ecmwf_cams_2024}, a key training source for foundation models, deviates from ground observations by an average of 52.66~$\mu$g/m$^3$ in China and Korea, and suffers from update latencies of a few days. 
Such limitations, (1) low regional accuracy and (2) lack of real-time availability, stem from data imbalance and physical divergence: East Asia comprises less than 15\% of global training coverage, yet contributes over 60\% of severe PM exposure. This clearly causes global models to underfit local dynamics.

\begin{figure}[t!]
  \centering
  \vspace{-2em}
    \includegraphics[width=0.95\linewidth]{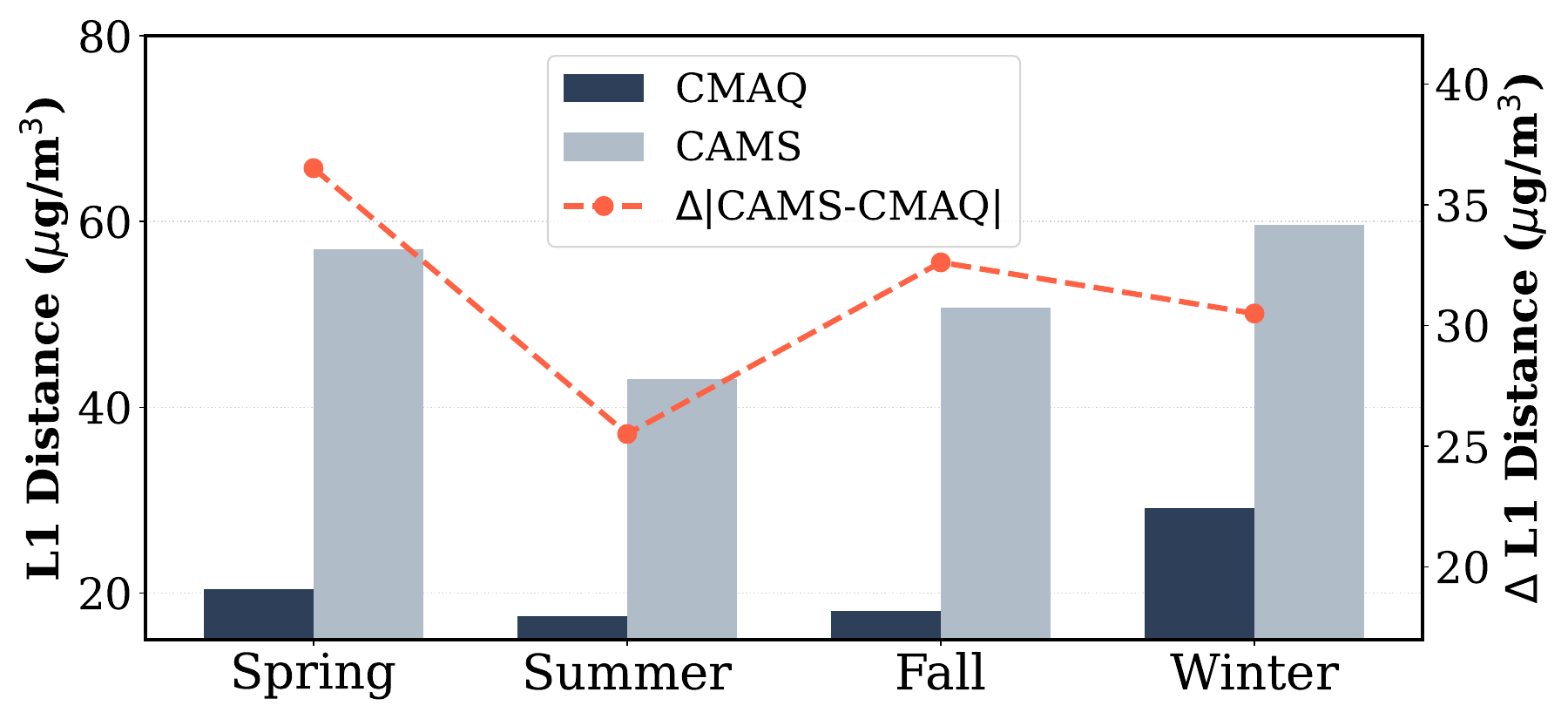}
    \vspace{-1em}
  \caption{\textbf{The seasonal L1 distance error compared with OBS ($\downarrow$ is better).}
Unlike global datasets (CAMS), our locally developed datasets (CMAQ) show low error with real observation.}
  \label{fig:cmaq_vs_cams}
  \vspace{-1.5em}
\end{figure}

To address both issues, we release a regional dataset spanning 2016--2023 (8+ years) that pairs real-world \textbf{OBS} from 532 Korean and 1,290--1,781 Chinese stations and high resolution Community Multiscale Air Quality (\textbf{CMAQ}) reanalysis at 27 km resolution for Korea and China. Our OBS data provides ground-truth measurements at 6~hour intervals, enabling accurate model supervision. CMAQ, tailored for East Asian meteorology and emissions, reduces error to 21.33~$\mu$g/m$^3$, a 59.5\% improvement over CAMS. Critically, CMAQ-driven forecasts can be initialized within hours from local observations, eliminating the 5~day update delay of global datasets such as CAMS. Training on this unified CMAQ--OBS dataset yields region-aware and operationally real-time ready models.

Once our regional CMAQ--OBS dataset ensures regional accuracy and real‑time availability, we establish a strong baseline by supervised training an Aurora-based 3D encoder-decoder on localized data. Although this baseline captures East Asian aerosol dynamics, it faces two limitations. First, long-range forecasting requires sequential 6-hour rollouts, yet teacher forcing exposes the model only to ground-truth states. This causes early inference errors to propagate across future steps. Second, the squared error objective treats all deviations uniformly and fails to capture the asymmetric costs of air-quality decisions, resulting in frequent false alarms despite high predictive accuracy. To address these limitations, we introduce \textbf{FAKER‑Air} (\textbf{F}orecast \textbf{A}lignment via \textbf{K}nowledge‑guided \textbf{E}xpected‑\textbf{R}eward), a two stage framework built on three core innovations.

First, we introduce a temporal accumulation loss to enforce temporal consistency during supervised finetuning~(SFT). Long horizon forecasting proceeds in sequential 6-hour steps. Each step consumes the previous prediction. Teacher forcing feeds only ground truth during training, so the model never sees its own rollouts, creating a train-test mismatch known as exposure bias~\cite{bengio2015scheduledsampling,ranzato2015sequence}. At inference, small early errors propagate across steps. We supervise N-step trajectories with the temporal accumulation loss. The loss penalizes stepwise errors along the rollout and reduces exposure bias by stabilizing long-range forecasts.

Temporal accumulation loss still optimizes squared error and remains agnostic to operational costs. Our confusion analysis of the SFT baseline shows many false alarms in clean air levels and strong detection for severe pollution cases. This pattern reflects the symmetric penalty of regression objectives and the heavy influence of large residuals at peaks, which shifts predictions upward in uncertain cases.

To mitigate overestimate problems, we adopt Group Relative Policy Optimization (GRPO)~\citep{shao2024deepseekmath} with a class-wise Air Quality Index (AQI)~\citep{Jung2022} reward to align decisions with operational costs. The reward assigns stronger penalties to false alarms in \texttt{Good} and \texttt{Moderate}. It assigns higher gains for true positives and stronger penalties for misses in \texttt{Bad} and \texttt{VeryBad} to preserve recall. GRPO updates the policy by ranking multiple rollouts for the same input and by increasing the likelihood of higher reward trajectories.

Finally, we employ curriculum rollout scheduling to stabilize long lead optimization during GRPO training. Direct training with a long horizon increases the variance of return estimates and weakens credit assignment. Later states depend on the model’s own predictions, so early errors move the state distribution off manifold. Training starts with short horizons such as 6 hours and extends to 24 hours as learning progresses. The curriculum reduces gradient variance in early training and yields stable updates at long lead times.

Our novel framework improves operational reliability by introducing policy optimization into spatio-temporal forecasting for the first time. Experiments show that FAKER-Air reduces the FAR by 47.3\%, while maintaining strong F1-scores for all AQI classes. These advances validate the system's value for real-world air quality warnings.\\

\noindent Our work enables reliable localized long horizon PM forecasting with the following key contributions:
\begin{itemize}
\item \textbf{Regional Dataset for Real-Time Forecasting.} We release the first CMAQ--OBS dataset for East Asia, reducing errors by 59.5\% compared to CAMS and supporting real-time initialization from 1,822+ monitoring stations.
\item \textbf{Two-Stage Training Framework.} We combine SFT with multi-step temporal accumulation loss for temporal consistency and introduce GRPO with curriculum rollout and class-wise rewards for decision-aware optimization.
\item \textbf{Operational Reliability.} Our model improves F1-score by 3.5× over Aurora and reduces the FAR by 47.3\%, achieving balanced performance when evaluate on OBS.
\end{itemize}

%% file: sec/2_related.tex
\section{Related Work}
\subsection{Earth system models and datasets}
Conventional numerical weather prediction (NWP) and earth system models (ESM) \citep{hurrell2013community} couple atmospheric, oceanic, and land processes through primitive equations, but they remain computationally expensive, sensitive to initial conditions, and often inconsistent across modeling systems \citep{balaji2022general}. 
The Coupled Model Intercomparison Project (CMIP) \citep{meehl2000coupled} provides a standardized framework to evaluate these discrepancies and has generated large datasets that have enabled recent data-driven approaches.

Deep learning has accelerated progress in this direction. 
Large-ensemble models \citep{weyn2021sub} have demonstrated subseasonal forecasting skill, while generative models for short-term precipitation prediction \citep{ravuri2021skilful} capture local spatio-temporal variability. 
Subsequent advances include FourCastNet \citep{pathak2022fourcastnet} using Fourier Neural Operators for global forecasts and GraphCast \citep{lam2023learning} employing graph neural networks for accurate medium-range prediction. 
Transformer-based models such as ClimaX \citep{nguyen2023climax} and Pangu-Weather \citep{bi2023accurate} further unified diverse forecasting tasks, and physics-informed approaches such as ClimODE \citep{verma2024climode} improved temporal consistency. 
Aurora \citep{bodnar2025foundation} extends this trend through large scale pretraining on multimodal geoscientific data. Critically, Aurora is the only open-source foundation model that explicitly includes PM forecasting, making it the sole publicly available baseline for comparative evaluation.

Building on these developments, our work targets regional scale, real-time PM forecasting. 
Unlike global models, our focus is on capturing fine-grained spatial variability and local temporal dynamics for operational applications.

\subsection{Alignment Training}

Recent advances in alignment training offer an alternative to purely supervised objectives by optimizing task- or preference-aware rewards. 
Reinforcement learning from human feedback (RLHF) \cite{christiano2017rlhf} formalizes this using the Bradley-Terry (BT) model \cite{btmodel} to represent pairwise preferences, with policy optimization methods such as PPO maximizing expected reward under a KL constraint. 
However, online RL remains unstable and computationally expensive, motivating direct alignment methods.

Direct Preference Optimization (DPO) \cite{rafailov2023direct} removes the need for an explicit reward model by deriving a closed-form likelihood objective from the BT model, effectively treating preference learning as a classification task. 
Recent work links DPO to an MDP interpretation in which the model implicitly defines a policy $\pi_\theta$, and the objective approximates optimizing $Q(s,a)$ toward the optimal $Q^*(s,a)$ under Bellman consistency \cite{rafailov2024r,bellman1957markovian}. 
This perspective connects reinforcement-style value learning with SFT.

Several Direct Alignment Algorithms (DAA) build on this idea. ORPO \cite{hong2024orpo} aligns policies via preference-weighted likelihood without a reference model, and SimPO \cite{meng2025simpo} simplifies alignment through implicit reward estimation. Group-Relative Policy Optimization (GRPO) \cite{shao2024deepseekmath}, used in our framework, replaces absolute rewards with relative rankings of multiple rollouts from the same context, improving stability without critics or expensive reward modeling. In our setting, GRPO aligns predictions with operational goals such as reducing false alarms and improving detection of severe pollution, bridging supervised learning and decision-centered policy optimization.

%% file: sec/3_dataset.tex
\begin{figure}[t!]
\vspace{-2em}
  \centering
    \includegraphics[width=\linewidth]{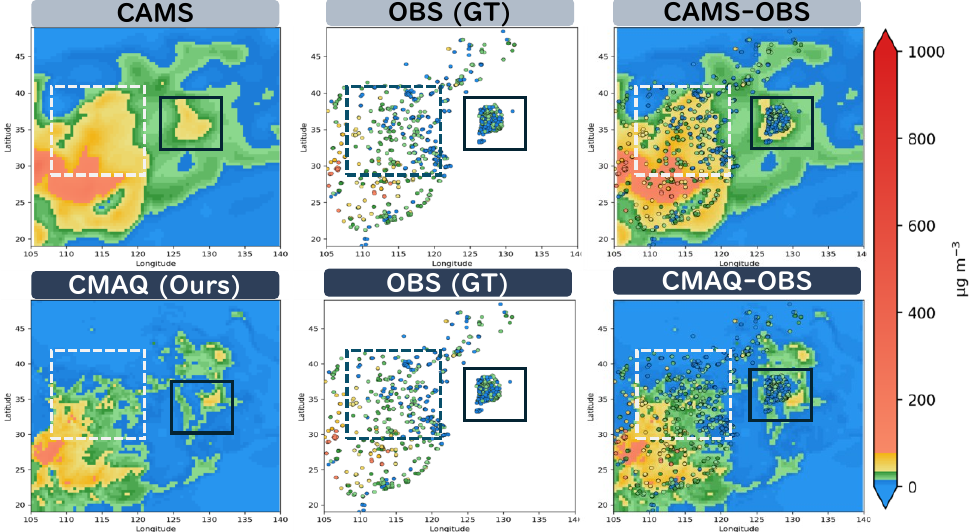}
    \vspace{-2em}
  \caption{\textbf{Comparison of CAMS and CMAQ~(ours) with OBS.} CMAQ achieves lower regional error and near real-time availability, enabling stable long horizon forecasting.}
  \label{fig:cmaq_vs_cams}
  \vspace{-1.5em}
\end{figure}

\section{Dataset}
\label{sec:dataset}

We construct and release the \textbf{CMAQ--OBS Regional Air Quality Dataset}, which jointly achieves (1) observation-validated regional accuracy and (2) real-time operational readiness. Existing global reanalysis datasets lack the capacity to satisfy both requirements simultaneously. Global datasets such as CAMS suffer from large regional biases due to coarse emissions and chemistry representations, and their multi-day update latencies prevent timely alert issuance. By integrating sparse ground-observation values with dense physics-driven regional CMAQ reanalysis across China and Korea, our dataset overcomes both limitations, establishing the foundation for reliable long horizon air quality forecasting in East Asia. Details on each dataset are in the supplementary materials.

\begin{figure*}[!t]
  \centering
  \vspace{-2.5em}
    \includegraphics[width=0.98\textwidth]{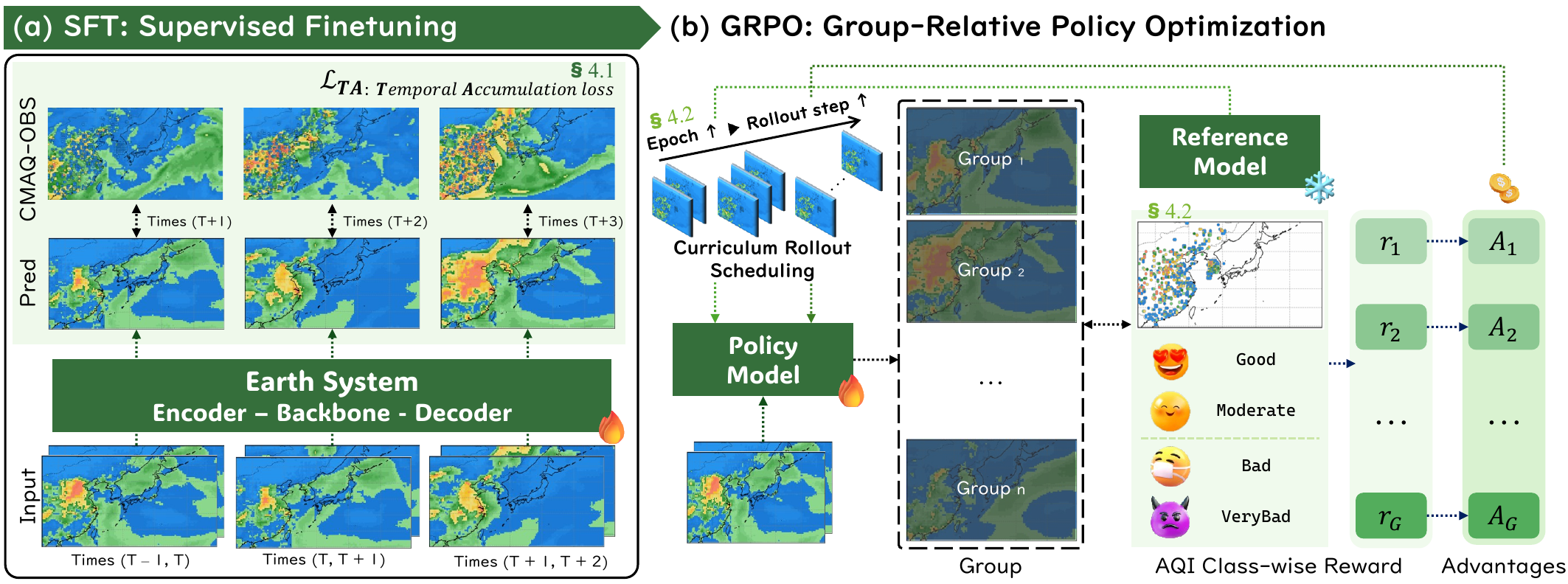}
  \caption{\textbf{Overall pipeline of FAKER-Air.} Our two-stage training framework begins with supervised fine-tuning (SFT) with rollout loss, followed by Group-Relative Policy Optimization (GRPO). During GRPO, multiple trajectory groups are evaluated using AQI-based rewards to guide policy updates, with the rollout horizon gradually increasing to enable long-term predictions.}
  \vspace{-1.5em}
  \label{fig:overview}
\end{figure*}

\paragraph{Station-based Observations (OBS).}
OBS dataset comprises 6~hour ground-level measurements of PM\textsubscript{2.5}, PM\textsubscript{10}, and O\textsubscript{3} concentrations. As shown in ~\Cref{fig:cmaq_vs_cams}, the data is sourced from 532 stations in Korea through the AirKorea network and 1,290--1,781 stations in China via the Airquality monitoring system, spanning the period from January 2016 to June 2024. The station-based point observations are collected at 6-hour temporal resolution.
For our modeling framework, we spatially interpolate these observations onto the CMAQ grid at 27 km resolution, producing structured fields aligned with our modeling domain.
Each gridded OBS field provides surface-level pollutant concentrations. These fields serve as both input features and ground truth targets, split into training and test sets. Critically, they capture real-time atmospheric conditions without the multi-day update delays inherent in global reanalysis products.
The Korean observations are complemented by meteorological measurements from 94 ASOS sites and 349 AWS stations and Chinese meteorological observations from Wyoming.

\paragraph{CMAQ.} Community Multiscale Air Quality (CMAQ) produces a spatially continuous field spanning 2016--2023 that covers the entire model domain, generating predictions for every grid cell at each time step and providing complete regional coverage beyond individual station locations, as shown in~\Cref{fig:cmaq_vs_cams}. Unlike point-based OBS, CMAQ delivers domain-wide gridded fields that fill spatial gaps with physics-consistent aerosol dynamics, enabling the model to learn transport, boundary conditions, and dispersion patterns. When validated against ground truth, CMAQ tailored to East Asian meteorology and emissions achieves an average error of 21.33~$\mu$g/m$^3$, representing a 59.5\% improvement over CAMS (52.66~$\mu$g/m$^3$). It also supports initialization within hours from locally available observations, satisfying real-time operational requirements. Besides, its multivariate spatial continuity provides physically grounded priors that mitigate distribution shift during auto-regressive rollouts and prevent the long horizon degradation common in teacher-forced training. By supplying coherent structural signals across seasons and regions, CMAQ enhances generalization under the sparse and variable pollution patterns of East Asia.

%% file: sec/4_method.tex
\section{Method}
\label{sec:method}

This section introduces our two stage training framework, \textbf{FAKER-Air} (\textbf{F}orecast \textbf{A}lignment via \textbf{K}nowledge-guided \textbf{E}xpected-reward \textbf{R}einforcement learning), for long horizon air quality forecasting. As illustrated in \Cref{fig:overview}, we first describe Stage 1, which establishes fundamental predictive capacity via SFT with temporal accumulation loss, then present Stage 2, which further improves operationally reliable forecasting by directly optimizing decision-centered rewards. This second stage complements SFT by addressing its remaining limitations related to over-prediction bias and asymmetric operational costs.

\subsection{Stage 1: SFT with temporal accumulation loss}
\label{sec:stage1}

We begin with SFT to train a base model capable of learning regional aerosol dynamics from historical CMAQ and OBS data. 
As illustrated in ~\Cref{fig:overview}\textcolor{cvprblue}{(a)}, given a sequence of input grids $\mathbf{x}_{1:T}$ and the next-step target $\mathbf{y}_{T+1}$, 
the model $f_\theta$ learns to minimize the Mean Squared Error (MSE) loss:
\begin{equation}
    \mathcal{L}_{\mathrm{SFT}} = 
    \mathbb{E}_{\mathbf{x}, \mathbf{y}} \big[\| f_\theta(\mathbf{x}_{1:T}) - \mathbf{y}_{T+1} \|_2^2\big].
\end{equation}
This supervised stage captures the coarse-scale spatio-temporal structure of pollutant transport and builds a strong baseline for downstream optimization. It allows the model to produce physically consistent concentration fields over multi hour horizons by leveraging the spatial encoder-decoder backbone. Although SFT provides a straightforward way to adapt the pretrained Aurora-like model to a regional domain, 
we observe a critical limitation when applying it to long horizon PM forecasting.

\paragraph{Addressing error accumulation with Temporal Accumulation (TA) loss.}
Teacher-forced supervised training treats each lead independently and does not expose the model to its own rollout errors, which creates a train-test mismatch known as exposure bias~\citep{bengio2015scheduledsampling}. This causes small early errors to compound over the horizon and degrade multi-step forecasts~\cite{schmidt2019generalizationgenerationcloserlook,ranzato2015sequence,rennie2017selfcriticalsequencetrainingimage,zhou2025taming}.

To address exposure bias, we adopt temporal accumulation loss that supervises auto-regressive forecast trajectories over $H$ lead times rather than single-step predictions. 
At each step $i \in \{1, \ldots, H\}$, the model generates predictions $\hat{\mathbf{y}}_{T+i}$ conditioned on its own previous outputs in an auto-regressive manner:
\begin{equation}
    \hat{\mathbf{y}}_{T+i} = f_\theta\big(\mathbf{x}_{1:T}, \hat{\mathbf{y}}_{T+1:T+i-1}\big), \quad i = 1, \ldots, H.
\end{equation}
The step-wise loss $\ell_i(\theta)$ at each lead time $i$ is computed as a weighted MSE across multiple variable groups. 
To progressively emphasize longer forecast horizons, we employ linearly increasing step weights $w_i = b + (1-b)\frac{i-1}{H-1}$ for $H>1$, where $b \in (0,1]$. 
The final temporal accumulation loss aggregates normalized step losses:
\begin{equation}
    \mathcal{L}_{\mathrm{TA}}(\theta) = \sum_{i=1}^{H} \tilde{w}_i \cdot \ell_i(\theta), 
    \quad \text{where} \quad 
    \tilde{w}_i = \frac{w_i}{\sum_{j=1}^{H} w_j}.
\end{equation}
By exposing the model to multi-step error accumulation during training, the temporal accumulation loss reduces the distribution shift between teacher-forced training and auto-regressive inference, thereby improving temporal consistency over extended forecast horizons.

%%%%%%%%%%%%%%%%%%%%%%%%%%%%%%%%%%%%%%%%%%%%%%%%%%%%%%%%%%%%%%%%%%%%
%%%%%%%%%%%%%%%%%%%% STAGE 2
%%%%%%%%%%%%%%%%%%%%%%%%%%%%%%%%%%%%%%%%%%%%%%%%%%%%%%%%%%%%%%%%%%%%
\subsection{Stage 2: Group-Relative Policy Optimization}
\label{sec:stage2}

\paragraph{Motivation: Decision-cost mismatch.}
While temporal accumulation improves long horizon stability, the MSE objective remains misaligned with operational decisions. 
Quadratic loss over-weights large residuals, biases predictions upward in uncertain regimes \cite{jadon2024comprehensive}, and can over-smooth spatial structure \cite{ledig2017photo}. 
Critically, squared error is cost-symmetric, whereas air quality operations are not: missing a severe event (\texttt{Bad} or \texttt{VeryBad}) is more costly than issuing a false alarm under clear air (\texttt{Good} or \texttt{Moderate}) \cite{elkan2001cost}. 
We therefore add a second stage that directly optimizes a verifiable, cost-sensitive reward that encodes asymmetric penalties for false alarms and misses, aligning policy updates with real-world alerting priorities.

\paragraph{Policy optimization formulation.}
We cast forecasting as a policy optimization problem. The model $f_\theta$ defines a stochastic policy 
$\pi_\theta(a_t \mid s_t)$, where the state $s_t$ encodes the spatiotemporal inputs and the action $a_t$ denotes the predicted concentration field at time $t{+}1$. 
The optimization objective is to maximize the expected task reward:
\begin{equation}
    \mathcal{J}(\theta)
    = \mathbb{E}_{\pi_\theta}\!\left[\sum_{t=1}^{T} r(s_t,a_t)\right],
\end{equation}
where $r(s_t,a_t)$ is a verifiable AQI-based reward. 
This reward is used only to evaluate forecast quality. In GRPO, it is converted into a relative advantage signal rather than being applied directly to the policy gradient.

\paragraph{GRPO mechanism.}
GRPO~\cite{shao2024deepseekmath} extends policy gradient methods by replacing absolute returns with group-wise relative comparisons. 
Starting from the SFT-trained policy, we generate $G$ trajectories by sampling actions from a Gaussian policy 
$a = \mu + \sigma\epsilon$, $\epsilon\sim\mathcal{N}(0,I)$, using antithetic pairs for variance reduction. 
Each trajectory produces a reward $r_t^{(g)}$, forming a group
$\mathcal{G}_t = \{(a_t^{(g)}, r_t^{(g)})\}_{g=1}^G$.

Rather than using raw rewards, GRPO converts them into softmax-normalized weights:
\begin{equation}
    A_g = \frac{\exp(r_t^{(g)} / \tau)}{\sum_{j=1}^G \exp(r_t^{(j)} / \tau)},
\end{equation}
where $A_g$ functions as a relative advantage indicating the performance of action $a_t^{(g)}$ compared with other rollouts under the same state. 
The policy is updated by weighting log-likelihoods with these advantage-like coefficients:
\begin{equation}
    \mathcal{L}_{\mathrm{GRPO}}
    = -\mathbb{E}_{(a_t^{(g)}, r_t^{(g)}) \in \mathcal{G}_t}\!\left[ A_g\, \log \pi_\theta(a_t^{(g)} \mid s_t ) \right].
\end{equation}

This ranking-based formulation increases the likelihood of actions with higher relative advantage while suppressing unreliable ones, effectively steering the policy toward trajectories that outperform their group peers. 
As illustrated in \Cref{fig:overview}\textcolor{cvprblue}{(b)}, GRPO evaluates multiple action rollouts for each state, converts their AQI rewards into relative advantages through groupwise normalization, and adjusts the policy accordingly. 
% This process encourages forecast behaviors that are operationally reliable, notably reducing false alarms in clean air conditions while preserving high recall for severe pollution events.
This process steers the policy toward operationally reliable predictions, reducing false alarms in clean air conditions while preserving high recall for severe pollution events.

\input{tables/sft_ablation}

\paragraph{Class-wise reward design.}
To encode asymmetric operational costs, we adopt a class-wise reward function $R(a_t, y_t)$ defined over discrete AQI categories $\mathcal{C} = \{\texttt{Good}, \texttt{Moderate}, \texttt{Bad}, \texttt{VeryBad}\}$. 
Let $\hat{c}_t = \text{AQI}(a_t)$ and $c_t = \text{AQI}(y_t)$ denote the predicted and true AQI classes, respectively. 
The reward follows the binary scheme used in RLVR (Reinforcement Learning with Verifiable Rewards)~\citep{shao2024deepseekmath}, which provides a verifiable and stable signal for policy comparison:
\begin{equation}
R(a_t, y_t) =
\begin{cases}
1, & \text{if } \hat{c}_t = c_t,\\[4pt]
0, & \text{otherwise.}
\end{cases}
\end{equation}
This straightforward binary structure allows GRPO to rank trajectories based on verifiable correctness rather than continuous value estimation, simplifying optimization while maintaining stability. Within the group-relative framework, it naturally captures asymmetric operational priorities by rewarding accurate classification in severe pollution cases while discouraging false alarms under clean air conditions. This alignment steers policy updates toward real-world decision objectives in air-quality alerting systems.

\input{tables/grpo_ablation}

\paragraph{Curriculum Rollout (CR) scheduling.}
To stabilize long horizon learning during GRPO, we introduce Curriculum Rollout (CR) scheduling where the rollout horizon $H$ gradually increases with training epochs.
At early epochs, we restrict $H$ to short lead times (e.g., 1-step) and progressively extend to longer horizons (up to 4-steps).
Formally:
\begin{equation}
    H_e = \min \big( H_{\max}, \lfloor H_{\min} + \kappa e \rfloor \big),
\end{equation}
where $e$ denotes the current epoch and $\kappa$ controls expansion rate.
As illustrated in ~\Cref{fig:overview}\textcolor{cvprblue}{(b)}, this progressive extension allows the model to master short-term dynamics before tackling uncertain long range forecasts, significantly reducing gradient noise in early training and improving long horizon prediction reliability.
By evaluating forecast trajectories over multiple lead times, GRPO implicitly encourages temporal consistency. Unlike SFT, which optimizes per-step snapshots, policy gradients of GRPO are informed by entire forecast sequences spanning long hours. This allows the model to capture long-term dependencies in aerosol transport and accumulation dynamics.

\paragraph{Integration of SFT and GRPO.}
To the best of our knowledge, the GRPO stage represents the first application of policy optimization to spatio-temporal forecasting, complementing SFT by bridging the gap between numerical accuracy and decision reliability. While SFT establishes a strong baseline and captures the underlying spatio-temporal dynamics, GRPO refines the model through reward-driven optimization aligned with operational objectives. Its group-relative weighting and class-wise AQI rewards directly address the decision-cost mismatch left by MSE, while curriculum rollout scheduling stabilizes policy learning over extended horizons. Through this two-stage training, the model achieves decision-grade stability in multi-day PM forecasts, reducing the FAR from while maintaining comparable performance as described in \Cref{sec:exp}.

%% file: tables/sft_ablation.tex
\begin{table*}[t!]
  \centering
  \caption{\textbf{Ablation Study on SFT for PM\textsubscript{2.5} and PM\textsubscript{10} on long lead time PM forecasting (12-hour intervals up to 120h).} Integrating OBS with CMAQ and extending temporal accumulation loss ($\mathcal{L}_{\mathrm{TA}}$) to T=4 consistently improves F1-scores across all horizons.}
  \vspace{-0.5em}
  \label{tab:f_score_ablation_split_12h_both}

  \renewcommand{\arraystretch}{1.0}
  \normalsize

  \resizebox{\textwidth}{!}{
  \begin{tabular}{cccc|c|ccccccccccc}
    \toprule
    \noalign{\vskip 1pt}
    \multicolumn{15}{c}{\textbf{PM\textsubscript{2.5}}} \\ % ✅ PM\textsubscript{2.5} 섹션 헤더
    \noalign{\vskip 1pt}
    \midrule
    OBS & CMAQ & $\mathcal{L}_{\mathrm{TA}}$~{\scriptsize(T=2)} & $\mathcal{L}_{\mathrm{TA}}$~{\scriptsize(T=4)} & Overall & +12h & +24h & +36h & +48h & +60h & +72h & +84h & +96h & +108h & +120h \\
    \midrule
    \multicolumn{4}{c|}{Aurora Air Pollution~\citep{bodnar2025foundation}} 
      & 16.06 & 48.82 & 40.07 & 17.75 & 15.00 & 4.04 & 4.01 & 1.16 & 1.72 & 0.46 & 1.05 \\
    \noalign{\vskip 2pt}\hline\hline\noalign{\vskip 2pt}
    \yesmark & \nomark & \nomark  & \nomark  
      & 50.74 & 61.53 & 53.83 & 51.57 & 49.87 & 48.79 & 48.47 & 48.15 & 47.76 & 47.68 & 47.65 \\
    \yesmark & \yesmark & \nomark  & \nomark  
      & 54.40 & 62.56 & 58.89 & 56.73 & 54.42 & 52.26 & 51.86 & 51.36 & 51.36 & 51.41 & 51.38 \\
    \yesmark & \yesmark & \yesmark & \nomark  
      & 57.65 & 65.89 & 61.99 & 60.52 & 57.58 & 55.48 & 54.75 & 54.04 & 53.76 & 53.44 & 53.34 \\
      \rowcolor{cyan!5}
    \yesmark & \yesmark & \nomark & \yesmark  
      & 59.90 & 67.53 & 64.48 & 61.83 & 60.39 & 59.05 & 58.23 & 57.44 & 56.94 & 56.54 & 56.21 \\ [-3pt]
    %\quad  \scriptsize{($\uparrow \Delta$)}
    \rowcolor{cyan!5}
    &    &   &  & \scriptsize{\textcolor{RoyalBlue}{\texttt{(+9.16)}}} & \scriptsize{\textcolor{RoyalBlue}{\texttt{(+6.00)}}} & \scriptsize{\textcolor{RoyalBlue}{\texttt{(+10.65)}}} & \scriptsize{\textcolor{RoyalBlue}{\texttt{(+10.26)}}} & \scriptsize{\textcolor{RoyalBlue}{\texttt{(+10.52)}}}  & \scriptsize{\textcolor{RoyalBlue}{\texttt{(+10.26)}}} & \scriptsize{\textcolor{RoyalBlue}{\texttt{(+9.76)}}} & \scriptsize{\textcolor{RoyalBlue}{\texttt{(+9.29)}}} & \scriptsize{\textcolor{RoyalBlue}{\texttt{(+9.18)}}} & \scriptsize{\textcolor{RoyalBlue}{\texttt{(+8.86)}}} & \scriptsize{\textcolor{RoyalBlue}{\texttt{(+8.56)}}}\\
      \noalign{\vskip 1pt}
    \toprule
    \noalign{\vskip 1pt}
    \multicolumn{15}{c}{\textbf{PM\textsubscript{10}}} \\ % ✅ PM$_{tiny10} 섹션 헤더
    \noalign{\vskip 1pt}
    \midrule
    OBS & CMAQ & $\mathcal{L}_{\mathrm{TA}}$~{\scriptsize(T=2)} & $\mathcal{L}_{\mathrm{TA}}$~{\scriptsize(T=4)} & Overall & +12h & +24h & +36h & +48h & +60h & +72h & +84h & +96h & +108h & +120h \\
    \midrule
    \multicolumn{4}{c|}{Aurora Air Pollution~\citep{bodnar2025foundation}} 
      & 4.73 & 23.34 & 14.08 & 3.38 & 1.78 & 0.22 & 0.23 & 0.02 & 0.04 & 0 & 0 \\
    \noalign{\vskip 2pt}\hline\hline\noalign{\vskip 2pt}
    \yesmark & \nomark & \nomark & \nomark  
      & 41.63 & 61.12 & 50.80 & 43.08 & 39.65 & 37.11 & 35.99 & 35.16 & 34.80 & 34.37 & 34.12 \\
    \yesmark & \yesmark & \nomark  & \nomark
      % & 42.09 & 65.41 & 59.31 & 53.76 & 50.72 & 48.15 & 46.53 & 45.51 & 44.70 & 44.11 & 43.45 \\
      & 52.65 & 67.71 & 62.97 & 59.30 & 57.32 & 55.77 & 55.09 & 54.58 & 54.23 & 53.84 & 53.60 \\
    \yesmark & \yesmark & \yesmark  & \nomark
      & 54.60 & 66.68 & 62.17 & 57.92 & 55.16 & 52.79 & 50.94 & 49.86 & 48.87 & 48.13 & 47.36 \\
      \rowcolor{cyan!5}
    \yesmark & \yesmark & \nomark  & \yesmark
      & 57.32 & 66.65 & 63.01 & 59.96 & 57.97 & 56.07 & 54.91 & 53.96 & 53.39 & 52.97 & 52.46 \\ [-3pt]
    %\quad  \scriptsize{($\uparrow \Delta$)}
    \rowcolor{cyan!5}
    &    &   &  & \scriptsize{\textcolor{RoyalBlue}{\texttt{(+15.69)}}} & \scriptsize{\textcolor{RoyalBlue}{\texttt{(+5.53)}}} & \scriptsize{\textcolor{RoyalBlue}{\texttt{(+12.21)}}} & \scriptsize{\textcolor{RoyalBlue}{\texttt{(+16.88)}}} & \scriptsize{\textcolor{RoyalBlue}{\texttt{(+18.32)}}}  & \scriptsize{\textcolor{RoyalBlue}{\texttt{(+18.96)}}} & \scriptsize{\textcolor{RoyalBlue}{\texttt{(+18.92)}}} & \scriptsize{\textcolor{RoyalBlue}{\texttt{(+18.80)}}} & \scriptsize{\textcolor{RoyalBlue}{\texttt{(+18.59)}}} & \scriptsize{\textcolor{RoyalBlue}{\texttt{(+18.60)}}} & \scriptsize{\textcolor{RoyalBlue}{\texttt{(+18.34)}}}\\
      \noalign{\vskip 1pt}
    \toprule
  \end{tabular}}
\end{table*}

%% file: tables/grpo_ablation.tex
\begin{table*}[t]
  \centering
  \caption{\textbf{Ablation on GRPO reward design and curriculum rollout (CR, e=4) for PM\textsubscript{2.5} over 120h.}
  For Binary(2-class): Acc, F1, Prec (↑) and FAR (↓); Bias≈1 preferred. 
  For AQI(4-class): Acc, F1-macro, F1-weighted, F1-micro (↑), plus per-class F1.}
  \vspace{-0.5em}
  \label{tab:grpo_ablation}
  \small
  \setlength{\tabcolsep}{3pt}
  \renewcommand{\arraystretch}{1.0}

  \resizebox{\textwidth}{!}{
  \begin{tabular}{
  l c c !{\vrule width \arrayrulewidth}
  c c c c c
  !{\vrule width \arrayrulewidth}
  c c c c
  !{\vrule width \arrayrulewidth}
  c c c c
}
    \toprule
    \multicolumn{3}{c|}{\textbf{Config}} &
      \multicolumn{5}{c|}{\textbf{Binary Metrics}} &
      \multicolumn{8}{c}{\textbf{AQI (4-class) Metrics}} \\
    \cmidrule(l){1-3} \cmidrule(l){4-8} \cmidrule(l){9-16}

    \multicolumn{1}{c}{\textbf{Model}} & \textbf{Reward} & \textbf{CR} &
      Acc & F1 & Prec & FAR & Bias &
      Acc & F1-macro & F1-weighted & F1-micro &
      \texttt{Good} & \texttt{Mod.} & \texttt{Bad} & \texttt{V.Bad} \\

    \midrule

    Aurora & -- & -- 
      & 68.93 & 16.06 & \textbf{68.40} & 2.24 & 0.13
      & 34.03 & 23.43 & 28.91 & 34.03
      & 42.49 & 36.82 & 11.46 & 2.96 \\

    FAKER-Air \textsubscript{SFT} & -- & --
      & 69.62 & \textbf{59.90} & 49.69 & 32.86 & 1.52
      & \underline{44.03} & \underline{41.59} & \underline{43.63} & \underline{44.03}
      & 38.46 & \textbf{47.79} & \textbf{44.92} & \textbf{35.18} \\

    \midrule
    \midrule

    FAKER-Air \textsubscript{GRPO} & MSE & \nomark
      & 74.50 & 48.44 & \underline{61.90} & \textbf{10.54} & 0.64
      & 43.73 & 37.57 & 42.36 & 43.73
      & 50.24 & \underline{46.09} & 33.93 & 20.04 \\

    FAKER-Air \textsubscript{GRPO} & AQI & \nomark
      & 71.40 & 56.28 & 52.12 & 24.19 & 1.17
      & 42.26 & 40.75 & 41.94 & 42.26
      & 49.19 & 40.05 & 39.42 & \underline{34.35} \\

    \rowcolor{cyan!6}FAKER-Air \textsubscript{GRPO} & AQI & \yesmark
      & \textbf{74.51} & \underline{56.72} & 57.98 & \underline{17.32} & \textbf{0.96}
      & \textbf{45.16} & \textbf{41.90} & \textbf{44.66} & \textbf{45.16}
      & \textbf{50.92} & 44.90 & \underline{41.80} & 29.98 \\

    \bottomrule
  \end{tabular}}
  \label{tab:ablation_combined_horizontal}
  \vspace{0.5em}
\end{table*}

%% file: sec/5_exp.tex
\section{Experiment}

\label{sec:exp}

\subsection{Metric}
We evaluate our framework using both binary and multi-class classification metrics to assess operational reliability and decision-grade forecasting performance. For binary evaluation, the \emph{F1-Score}~\citep{hand2021finterpretabletransformationfmeasure} summarizes event detection performance by jointly considering precision and recall. \emph{FAR} (False Alarm Rate)~\citep{Hicks2022} quantifies the fraction of false alerts among all non-event cases and serves as a key operational metric, since excessive false alarms reduce the practical usability of an alert system. We further report the \emph{CSI} (Critical Success Index)~\citep{Mbizvo2024-gd}, which penalizes both missed events and false alarms and is widely used for assessing rare event forecasting.  \emph{Bias}~\citep{bias} measures systematic over-prediction or under-prediction, with values near 1.0 indicating well-calibrated event frequency. For multi-class AQI evaluation across four pollution levels, we report \emph{F1-Macro}, which treats all pollution classes equally regardless of their frequency, thereby highlighting model performance on rare severe pollution events. \emph{F1-Weighted} accounts for class imbalance by weighting F1-score of classes proportionally to its sample size, providing a more representative overall performance measure. \emph{F1-Micro} aggregates predictions across all classes before computing the score, effectively measuring overall classification accuracy in the multi-class setting.

\subsection{Implementation Details}
We conduct a two-stage training procedure on East Asian air quality integrating real-time OBS and CMAQ reanalysis data to enable real-time long horizon localized forecasting. We employ SFT on an Aurora-based 3D encoder-decoder trained with batch size of 8 for 30 epochs with one-cycle LR. We apply rollout loss with a 4 step auto-regressive prediction. GRPO stage trained with batch size 1 for 4 epochs, generating 4 trajectory samples per input using antithetic sampling with common noise and optimizing a class-wise reward function based on discrete AQI thresholds. We use 2016 to 2021 as training, 2022 as validation, and 2023 as test datasets with 2 H200 GPUs distributed data parallel training with random seed 42. Further details are described in the supplementary materials.

\subsection{SFT Experimental Results}

Table~\ref{tab:f_score_ablation_split_12h_both} summarizes the effect of SFT components on long horizon forecasting. Training with OBS alone substantially improves F1-scores over Aurora across all horizons (PM\textsubscript{2.5}: 16.06 $\rightarrow$ 50.74; PM\textsubscript{10}: 4.73 $\rightarrow$ 41.63). Integrating CMAQ reanalysis further enhances overall performance and markedly improves stability for long lead time, indicating that physics-driven continuous fields compensate for regional biases in sparse observations and reduce distribution variability during auto-regressive rollouts, providing more stable conditions to address exposure bias.

Introducing the multi-step rollout loss $\mathcal{L}_{\mathrm{TA}}$ provides consistent gains across all horizons. For PM\textsubscript{2.5}, $\mathcal{L}_{\mathrm{TA}}(T=2)$ raises overall F1 from 54.40 to 57.65, while extending to $T=4$ yields 59.90, with sustained improvements at extended leads. PM\textsubscript{10} exhibits a similar pattern (52.65 $\rightarrow$ 57.32). By exposing the model to its own predictions during training, these results demonstrate direct mitigation of exposure bias and enhanced temporal consistency, especially beyond 60h where single-step teacher forcing typically suffers from error accumulation.

Overall, SFT with CMAQ--OBS data fusion and rollout-based objectives establishes a robust baseline by addressing exposure bias. Our SFT model achieves about 3.7× improvement in F1 for PM\textsubscript{2.5} and 12× for PM\textsubscript{10} relative to Aurora, effectively mitigating performance collapse beyond 60--120h. This provides the essential precondition for subsequent policy optimization (Section~\ref{sec:stage2}) to refine operational cost metrics such as FAR and class-wise metrics.

\subsection{GRPO Experimental Results}

\input{tables/grpo_aqi2}

\begin{figure*}[t!]
  \centering
  % \vspace{-1em}
    \includegraphics[width=\linewidth]{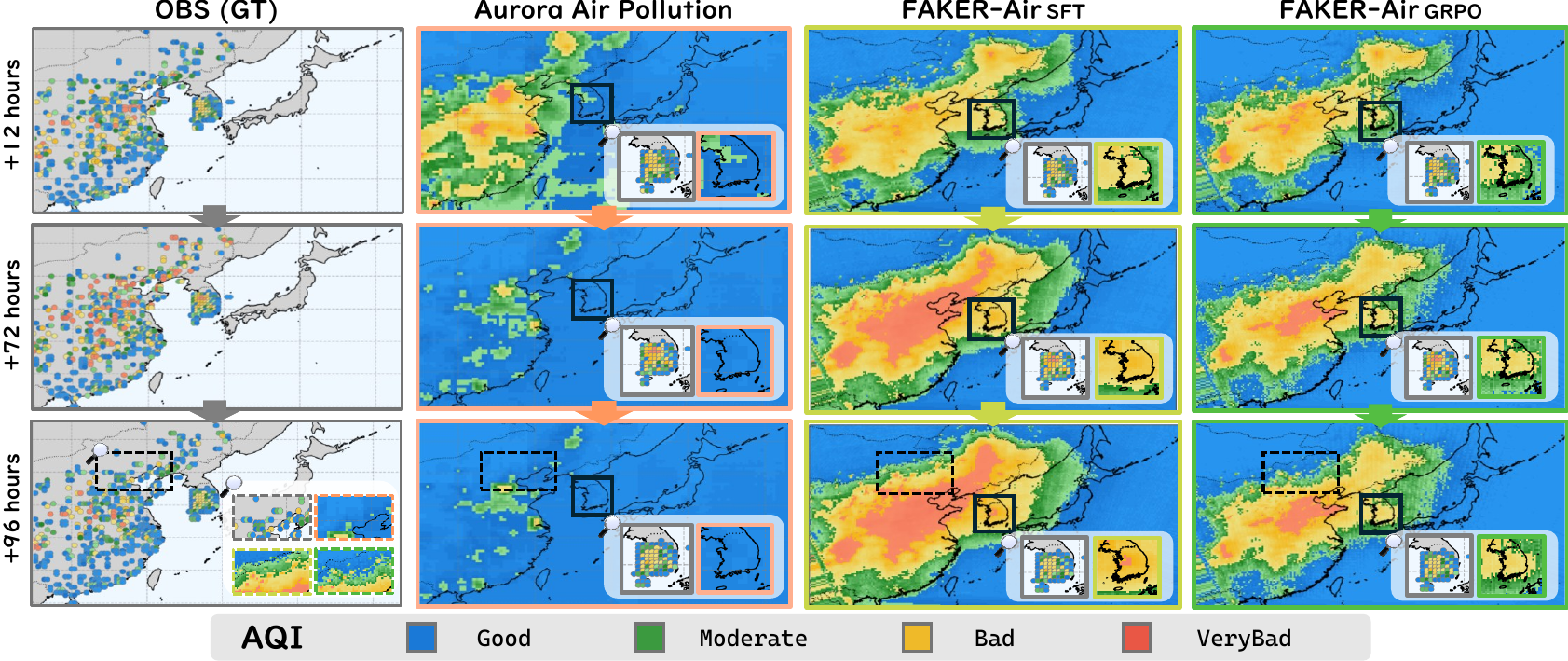}
  \caption{\textbf{Qualitative comparison of long horizon PM\textsubscript{2.5} forecasts over East Asia.} Aurora rapidly loses regional structure. \textbf{SFT} restores coherent transport but slightly overextends moderate pollution. \textbf{GRPO} prunes these artifacts while preserving high‑pollution cores.}
  \label{fig:quali_main}
  % \vspace{-0.5em}
\end{figure*}

Table~\ref{tab:csi_far_comparison} summarizes the impact of GRPO on long horizon AQI classification. While SFT establishes a strong baseline, it suffers from a decision cost mismatch because MSE treats all errors symmetrically. However, operational costs are asymmetric between false alarms and missed severe events. This causes SFT to exaggerate the probability of pollution, yielding a higher FAR despite an improved CSI. GRPO directly addresses this by optimizing class-wise AQI rewards, substantially lowering FAR by 47.3\% (32.86 $\rightarrow$ 17.32) while maintaining stable Macro CSI and achieving near-ideal Bias (1.52 $\rightarrow$ 0.96), demonstrating that asymmetric rewards resolve the decision cost mismatch left unaddressed by regression-based SFT.

A consistent pattern emerges for PM\textsubscript{10}, where GRPO cuts FAR by approximately 41\% (18.44 $\rightarrow$ 10.81) while maintaining comparable Macro CSI. The \texttt{Good} category improves substantially, while other classes change modestly, yielding a more conservative and trustworthy operating point for long horizon alerts. These class-wise adjustments demonstrate that AQI-based reward of GRPO effectively reduces FAR while preserving overall metrics.

Table~\ref{tab:ablation_combined_horizontal} provides deeper analysis of how reward design and curriculum rollout drive gains of GRPO. The choice of reward function proves critical: MSE-based reward proves ineffective for policy optimization. In contrast, class-wise AQI rewards achieve balanced performance by penalizing false positives in clean air states, though at the cost of elevated FAR. Introducing curriculum rollout restores reliability, reducing FAR to 17.32 while maintaining F1-score and achieving near-ideal Bias. These results demonstrate that GRPO benefits from both decision-aware rewards and progressively extended rollouts for long horizon stability.

Overall, SFT and GRPO form a complementary pipeline where SFT provides temporal consistency, while GRPO aligns decisions with operational costs via group-relative ranking, class-wise rewards, and rollout scheduling. To the best of our knowledge, this work represents the first application of policy optimization to time-series forecasting on climate, enabled by our CMAQ--OBS dataset that addresses the absence of validated regional baselines and real-time OBS data for East Asia. Our approach produces fewer unnecessary alerts. By leveraging asymmetric rewards and relative ranking, GRPO delivers decision-grade predictions suitable for operational air quality applications.

\subsection{Qualitative Results}

\Cref{fig:quali_main} contrasts spatial predictions at +12h, +72h, and +96h lead times. Aurora rapidly loses mesoscale structure, producing diffuse fields that fail to track the temporal evolution visible in ground-truth observations. In contrast, FAKER-Air \textsubscript{SFT} restores coherent spatial patterns and maintains sharp inter-basin contrasts across extended horizons, demonstrating that our temporal accumulation loss effectively suppresses cumulative error propagation during auto-regressive rollout. FAKER-Air \textsubscript{GRPO} further refines these predictions through decision-aware alignment, removing spurious \texttt{Bad} level artifacts that appear in SFT under clean air regimes while retaining high-concentration cores, directly mirroring the quantitative FAR reductions. Notably, at +96h our framework successfully captures transboundary pollution transport, a challenging pattern that Aurora completely fails to resolve. At this extended lead time, where Aurora collapses toward uniform backgrounds, our framework maintains both spatial fidelity and operational reliability across long horizon forecasts.

%% file: tables/grpo_aqi2.tex
\begin{table}[t!]
 
  \centering
\caption{\textbf{Comparison of Overall Performance.} CSI and FAR Comparison over 120 hours for PM\textsubscript{2.5} and PM\textsubscript{10}}
\vspace{-0.5em}
\renewcommand{\arraystretch}{1.1}  % 줄 간격 증가 (기본값 1.0)
\LARGE  % 글꼴 크기 증가 (기존보다 약간 큰 크기)
  \setlength{\heavyrulewidth}{1.5pt}
  \setlength{\lightrulewidth}{0.5pt}
\resizebox{\columnwidth}{!}{
\begin{tabular}{@{}l|cccc|c|c@{}} % 좌우 여백 제거

\toprule
\multicolumn{7}{c}{\textbf{PM\textsubscript{2.5}}} \\
\midrule

\multirow{2}{*}{\centering\arraybackslash\textbf{Model}} &
\multicolumn{4}{c|}{\textbf{CSI Score (AQI: 4 classes)}} &
\multirow{2}{*}{\centering\arraybackslash\makecell{\textbf{FAR$\downarrow$}\\(binary)}} &
\multirow{2}{*}{\centering\arraybackslash\makecell{\textbf{Macro}\\\textbf{CSI}}} \\
\cmidrule(lr){2-5}
& \texttt{Good} & \texttt{Mod.} & \texttt{Bad} & \texttt{V.Bad} & & \\
\midrule
\textbf{Aurora} & \underline{26.98} & 22.56 & 6.08 & 1.50 & \textbf{2.24} & 14.28 \\
\textbf{FAKER-Air~\textsubscript{SFT}} & 23.81 & \textbf{31.40} & \textbf{28.97} & \textbf{21.34} & 32.86 & \underline{26.38} \\
\rowcolor{cyan!5}\textbf{FAKER-Air~\textsubscript{GRPO}} & \textbf{34.16} & \underline{28.95} & \underline{26.42} & \underline{17.63} & \underline{17.32} & \textbf{26.79} \\ [-4pt]
\rowcolor{cyan!5}
    &    \Large{\textcolor{RoyalBlue}{\texttt{(+7.18)}}} & \Large{\textcolor{RoyalBlue}{\texttt{(+6.39)}}} & \Large{\textcolor{RoyalBlue}{\texttt{(+20.34)}}} & \Large{\textcolor{RoyalBlue}{\texttt{(+16.13)}}} & \Large{\textcolor{RoyalBlue}{\texttt{(-15.54)}}} & \Large{\textcolor{RoyalBlue}{\texttt{(+12.51)}}}\\

\midrule
\multicolumn{7}{c}{\textbf{PM\textsubscript{10}}} \\
\midrule
\multirow{2}{*}{\centering\arraybackslash\textbf{Model}} &
\multicolumn{4}{c|}{\textbf{CSI Score (AQI: 4 classes)}} &
\multirow{2}{*}{\centering\arraybackslash\makecell{\textbf{FAR$\downarrow$}\\(binary)}} &
\multirow{2}{*}{\centering\arraybackslash\makecell{\textbf{Macro}\\\textbf{CSI}}} \\
\cmidrule(lr){2-5}
& \texttt{Good} & \texttt{Mod.} & \texttt{Bad} & \texttt{V.Bad} & & \\
\midrule
\textbf{Aurora} & \underline{28.84} & 18.78 & 1.63 & 0.38 & \textbf{0.33} & 12.41 \\
\textbf{FAKER-Air~\textsubscript{SFT}} & 28.63 & \textbf{42.43} & \textbf{26.21} & \textbf{18.83} & 18.44 & \textbf{29.02} \\
\rowcolor{cyan!5}\textbf{FAKER-Air~\textsubscript{GRPO}} & \textbf{36.39} & \underline{37.53} & \underline{23.03} & \underline{15.35} & \underline{10.81} & \underline{28.07} \\ [-4pt]
\rowcolor{cyan!5}
    &    \Large{\textcolor{RoyalBlue}{\texttt{(+7.55)}}} & \Large{\textcolor{RoyalBlue}{\texttt{(+18.75)}}} & \Large{\textcolor{RoyalBlue}{\texttt{(+21.40)}}} & \Large{\textcolor{RoyalBlue}{\texttt{(+14.97)}}} & \Large{\textcolor{RoyalBlue}{\texttt{(-7.63)}}} & \Large{\textcolor{RoyalBlue}{\texttt{(+15.66)}}}\\
\bottomrule
\end{tabular}
}
% \vspace{-0.5em}
\label{tab:csi_far_comparison}
\end{table}

%% file: sec/6_conc.tex
\section{Conclusion}

This work addresses the latency and regional limitations of global foundation models for long horizon air quality forecasting in East Asia. We construct and release the CMAQ--OBS dataset that enables real-time initialization for East Asia, providing the research community with validated regional data to advance operational air quality forecasting and public health protection. Moreover, we propose a two stage framework, FAKER-Air, combining SFT with temporal accumulation loss and GRPO with class-wise rewards. Experimental results demonstrate 3.5× improvement in F1-score over Aurora and 47.3\% reduction in FAR. By unifying physical modeling with decision-aware optimization, our model balances accuracy and reliability for PM warning systems.

\section*{Acknowledgements}
\label{sec:Acknowledgements}

% RS-2025-00520207                          % 중견 (로봇상호)
% RS-2024-00457882                          % [AI연구거점]
% No. 2022-0-00680                          % [귀추]
% No. 2022-0-01045                          % [오픈 도메인]
% RS-2019-II190075                          % [카이스트 AI 대학원]
% No.RS-2025-02217259                       % [편향성]
% NIER-2025-04-02-032                       % [미세먼지]
% RS-2025-02653113                          % [GPU 지원과제]

\noindent
This research was supported by the Basic Science Research Program through the National Research Foundation of Korea (NRF) funded by the MSIP (No. RS-2025-00520207); Institute of Information \& communications Technology Planning \& Evaluation (IITP) grants funded by the Korea government (MSIT) (No. RS-2024-00457882, National AI Research Lab Project; RS-2019-II190075, Artificial Intelligence Graduate School Program (KAIST)); the Advanced GPU Utilization Support Program funded by the Government of the Republic of Korea (Ministry of Science and ICT) (No. RS-2025-02653113); Korea Evaluation Institute of Industrial Technology (KEIT) grants funded by the Korea government (MOTIE) (No. 2022-0-00680; No. 2022-0-01045); and a grant partly supported by both IITP (MSIT) and KEIT (MOTIE) (No. RS-2025-02217259, Development of self-evolving AI bias detection-correction-explain platform based on international multidisciplinary governance). Additionally, this work was supported by the Air Quality Forecasting Center at the National Institute of Environmental Research under the Ministry of Environment (NIER-2025-04-02-032).

%% file: sec/X_suppl.tex
\clearpage
\setcounter{page}{1}
\maketitlesupplementary
\setcounter{section}{0}
\setcounter{table}{0}
\setcounter{equation}{0}
\setcounter{figure}{0}
\renewcommand{\thefigure}{S\arabic{figure}}
\renewcommand{\thetable}{S\arabic{table}}
\renewcommand{\thesection}{\Alph{section}}
\renewcommand{\theequation}{S\arabic{equation}}

\paragraph{Rationale.}
\label{sec:rationale}

Due to the limited pages, we provide supplementary materials with the following contents:\vspace{0.3em}

\begin{itemize}
    \item Section~\ref{sec:aqi_threshold}: AQI Threshold Specification.\vspace{0.3em}
	\item Section~\ref{sec:sup_OBS}: OBS Dataset Details.\vspace{0.3em}
	\item Section~\ref{sec:sub_CMQA}: CMAQ Dataset Details.\vspace{0.3em}
	\item Section~\ref{sec:sup_east_special}: Why East Asia Requires Specialized Models \vspace{0.3em}
	  \item Section~\ref{sec:test_2016}: Additional Experiments (2016 Test Year).\vspace{0.3em}
	\item Section~\ref{sec:seasonal_analysis}: Seasonal Analysis. \vspace{0.3em}
    \item Section~\ref{sec:quali_results}: Additional Qualitative Results. \vspace{0.3em}
    \item Section~\ref{sec:extended_ablation}: Extended Ablation Study. \vspace{0.3em}
    \item Section~\ref{sec:decision_boundary}: Reliability and Decision Boundaries.\vspace{0.3em}
    \item Section~\ref{sec:explainability}: Explainability via Feature Importance.\vspace{0.3em}
	\item Section~\ref{sec:implementation}: Implementation Details.\vspace{0.3em}
	\item Section~\ref{sec:limitations}: Limitations and Future Work. \vspace{0.3em}
\end{itemize}

%%%%%%%%%%%%%%%%%%%%%%%%%%%%%%%%%%%%%%%%%%%%%%%%%%%%%%%%%%%%%%%%%%%%%%%%%%%%%%%%%%%%%%%%%%%%%%%%%%%%

\section{Air Quality Index (AQI) Threshold Specification}
\label{sec:aqi_threshold}

The Air Quality Index (AQI) is a standardized measurement system developed by governmental environmental agencies to communicate ambient air quality and associated health risks to the public~\citep{who2021aqg,epa_naaqs_2023}. AQI maps complex pollutant concentration data into a single, interpretable numerical scale with color-coded categories that guide protective actions for different population groups, facilitating public health messaging~\citep{who2021aqg, Jung2022}.

We adopt a dual-level classification scheme aligned with international air quality standards.
\paragraph{(1) Binary Classification.} For operational alerting systems, we define two primary categories:
\begin{itemize}
    \item \textbf{Clean}: Air quality poses minimal health risk, allowing unrestricted outdoor activities. Encompasses \texttt{Good} and \texttt{Moderate} AQI levels.
    \item \textbf{Polluted}: Air quality requires protective measures such as reduced outdoor exposure or mask usage. Encompasses \texttt{Bad} and \texttt{VeryBad} AQI levels.
\end{itemize}

\paragraph{(2) 4-Class AQI Classification.} For more granular policy, we subdivide based on international and national standards~\citep{who2021aqg,epa_naaqs_2023, Jung2022} (see Table~\ref{tab:aqi_classification}):

\begin{itemize}
    \item \texttt{Good}: PM\textsubscript{2.5} $\leq$ 15~$\mu$g/m$^3$, PM\textsubscript{10} $\leq$ 30~$\mu$g/m$^3$. Air is satisfactory, no restrictions needed.
    \item \texttt{Moderate}: PM\textsubscript{2.5} 16--35~$\mu$g/m$^3$, PM\textsubscript{10} 31--80~$\mu$g/m$^3$. Air is acceptable except for highly sensitive individuals.
    \item \texttt{Bad}: PM\textsubscript{2.5} 36--75~$\mu$g/m$^3$, PM\textsubscript{10} 81--150~$\mu$g/m$^3$. Unhealthy for sensitive groups, activity restriction recommended.
    \item \texttt{VeryBad}: PM\textsubscript{2.5} $\geq$ 76~$\mu$g/m$^3$, PM\textsubscript{10} $\geq$ 151~$\mu$g/m$^3$. Unhealthy for all, triggers emergency measures.
\end{itemize}

The PM\textsubscript{2.5} thresholds follow the WHO Global Air Quality Guidelines (2021), 15~$\mu$g/m$^3$ for annual mean and 35~$\mu$g/m$^3$ for 24-hour mean~\citep{who2021aqg}. PM\textsubscript{10} thresholds align with US EPA National Ambient Air Quality Standards (NAAQS)~\citep{epa_naaqs_2023}, while the stepwise system is adopted in Korean and Chinese national guidelines for operational use~\citep{Jung2022}. This 4-class system enables proportional public health response from advisory to emergency.

\input{tables/aqi_info}

Our forecasting framework predicts these discrete AQI categories at 6-hour intervals up to 120 hours ahead, enabling proactive decision-making for public health protection. The GRPO optimization explicitly targets accurate multi-class discrimination through class-wise reward functions (Section~\ref{sec:implementation}), addressing the operational asymmetry where \texttt{VeryBad} misclassifications pose greater health risks than \texttt{Good}/\texttt{Moderate} confusion. Table~\ref{tab:aqi_classification} summarizes the complete classification scheme with concentration thresholds for both pollutants, color-coded for intuitive interpretation in operational dashboards.

%%%%%%%%%%%%%%%%%%%%%%%%%%%%%%%%%%%%%%%%%%%%%%%%%%%%%%%%%%%%%%%%%%%%%%%%%%%%%%%%%%%%%%%%%%%%%%%%%%%%
\section{Observation (OBS) Dataset Details}\label{sec:sup_OBS}

Our observation dataset integrates ground-level air quality and meteorological measurements from monitoring networks across East Asia. As summarized in Table~\ref{tab:station_statistics}, the dataset comprises observations from 1,822 stations spanning Korea and China, providing real-time pollutant concentrations at 6-hour intervals from January 2016 to June 2024.

\subsection{Air Quality Monitoring Network}

Air quality records were collected from three major public networks. The Korean subsystem employs 532 stations operated by AIRKOREA~\citep{airkorea}, while the Chinese network aggregates data from 1,781 PM25.in stations and 1,290 AIRQUALITY (CN) stations~\citep{pm25in,airquality_network}. Each measures PM\textsubscript{2.5}, PM\textsubscript{10}, O$_3$, NO$_2$, CO, and SO$_2$ at hourly resolution, then aggregates to 6-hour intervals for modeling. These serve as both features and labels for supervised training.

Unlike global reanalysis data (e.g., CAMS), which have multi-day update delays, our direct station observations offer real-time atmospheric state for model initialization. This eliminates the 5-day data delay that constrains global datasets. All stations are spatially interpolated onto the CMAQ grid (27~km) using inverse distance weighting, producing structured, domain-aligned fields while preserving high temporal fidelity.

\input{tables/obs_stats}

\subsection{Meteorological Observations}

Complementary meteorological data are obtained from 94 ASOS (KMA) and 349 AWS (KMA) stations in Korea, and 40 upper-air stations in China (Wyoming Weather Web)~\citep{asos_korea,aws_kma,wyoming_upperair}. These provide temperature, humidity, pressure, wind, and vertical profiles at surface and several standard pressure levels. This auxiliary information enriches the feature set and supports accurate modeling of regional transport phenomena.

\subsection{Spatial Coverage and Data Quality}

Station density varies by region and is highest in urban zones. Korean coverage reaches $\sim$ 0.5~stations/1,000~km$^2$, major Chinese cities $\sim$ 0.2~stations/1,000~km$^2$, together capturing all major pollution corridors and population centers. We implement standard quality controls including range checks, temporal continuity, and outlier removal. Missing observations are interpolated if nearby stations are available, or excluded if gaps persist beyond 12 hours.

%%%%%%%%%%%%%%%%%%%%%%%%%%%%%%%%%%%%%%%%%%%%%%%%%%%%%%%%%%%%%%%%%%%%%%%%%%%%%%%%%%%%
\input{tables/cmaq_vs_cams}

\begin{figure}[t!]
  \centering
    \includegraphics[width=0.95\linewidth]{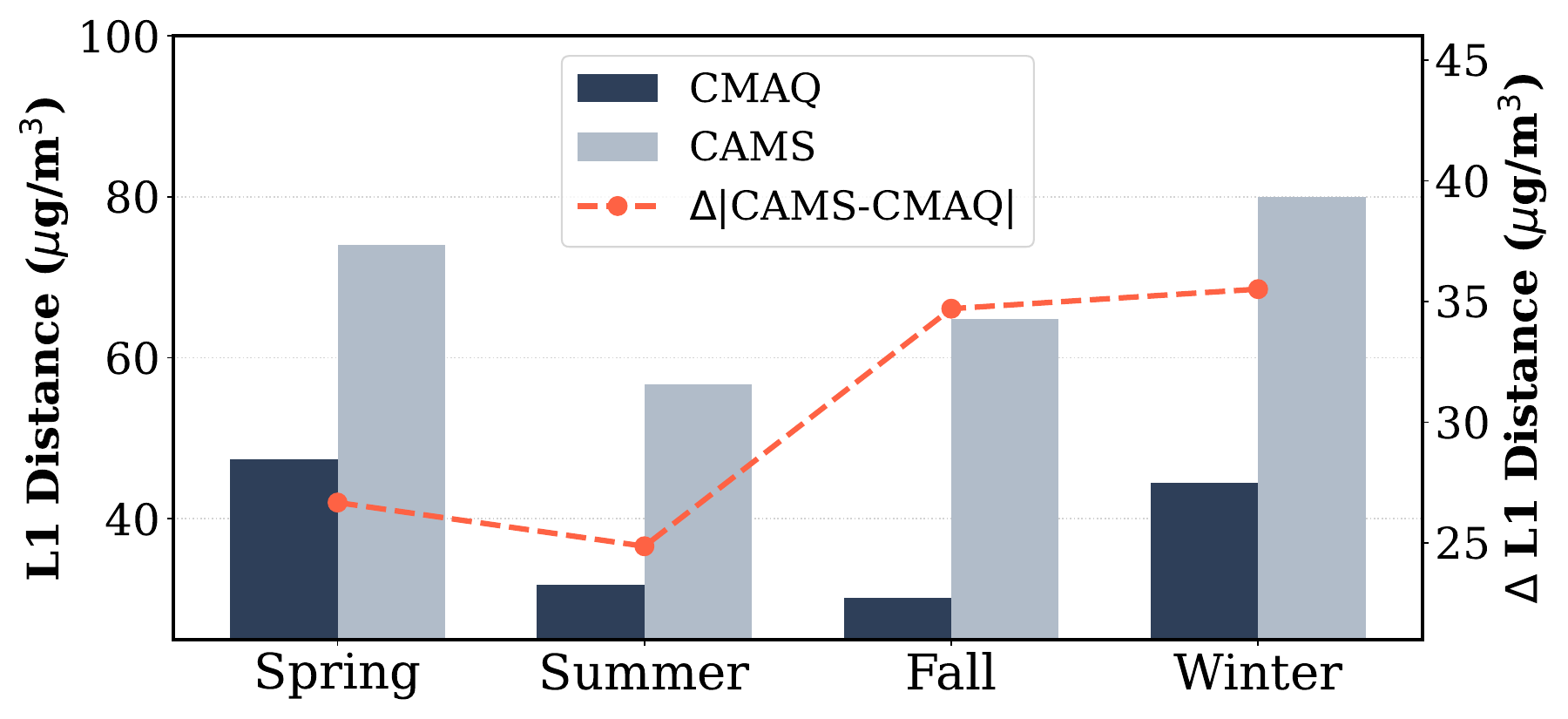}
    \vspace{-1em}
  \caption{\textbf{The seasonal L1 distance error compared with OBS ($\downarrow$ is better).}
Unlike global datasets (CAMS), our locally developed datasets (CMAQ) show low error with real OBS on PM\textsubscript{10}.}
  \label{fig:cmaq_vs_cams}
\end{figure}

\begin{figure*}[t]
  \centering
  % Top subfigure
  \vspace{5em}
  \begin{subfigure}[b]{0.88\textwidth}
    \centering
    \includegraphics[width=0.88\textwidth]{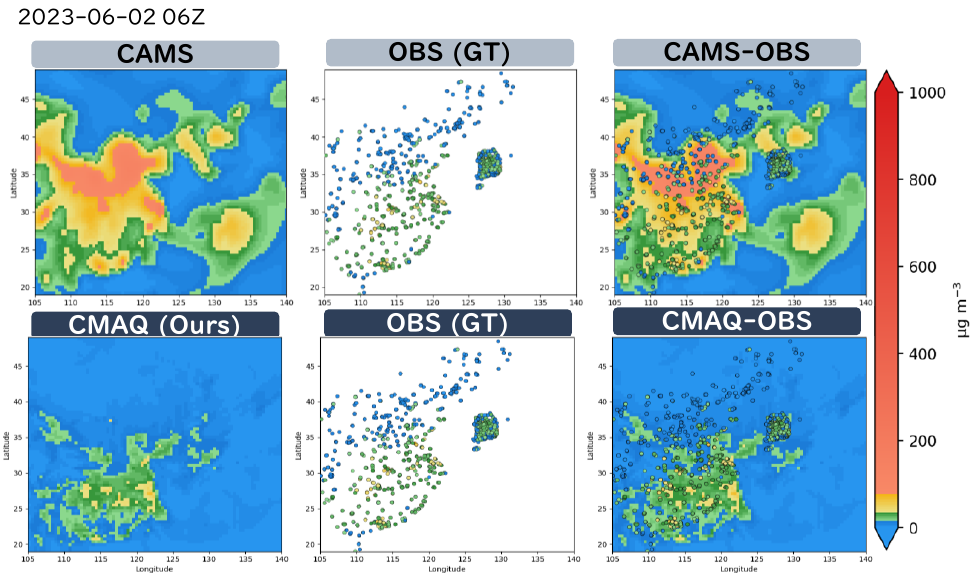}
    \caption{Spatial distribution comparison on Summer.}
    \label{fig:supp_comparison_1}
    \vspace{2em}
  \end{subfigure}
  
  % Bottom subfigure
  \begin{subfigure}[b]{0.88\textwidth}
    \centering
    \includegraphics[width=0.88\textwidth]{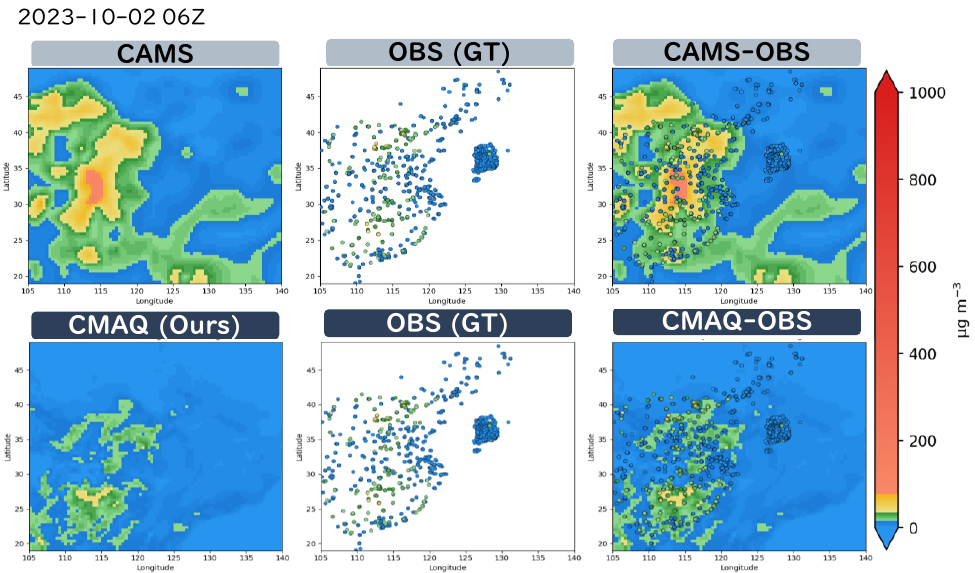}
    \caption{Spatial distribution comparison on Fall.}
    \label{fig:supp_comparison_2}
  \end{subfigure}
  
  \caption{\textbf{Seasonal Comparison of PM\textsubscript{2.5} Spatial Distributions across OBS, CMAQ, and CAMS Datasets.} 
The figure presents PM\textsubscript{2.5} spatial distributions for Summer (a) and Fall (b) across East Asia. CMAQ demonstrates superior regional fidelity to ground observations (OBS) compared to CAMS in both seasons, capturing fine-scale pollution dynamics that CAMS fails to resolve due to coarse resolution and update latencies.}
  \label{fig:supp_spatial_comparison}
  \vspace{3em}
\end{figure*}

\section{CMAQ Dataset Details}\label{sec:sub_CMQA}

The Community Multiscale Air Quality (CMAQ) dataset provides spatially continuous, high-resolution reanalysis fields generated through a coupled meteorology--chemistry modeling system. Unlike point-based ground observations, CMAQ offers gridded atmospheric states across the entire East Asian domain, filling spatial gaps between monitoring stations with physics-consistent aerosol dynamics. This spatial continuity enables the model to learn regional transport patterns, boundary conditions, and pollutant dispersion mechanisms that are difficult to infer from ground observations alone.

\subsection{Modeling Chain and Configuration}

The CMAQ reanalysis is generated through a three-stage modeling pipeline. First, the Weather Research and Forecasting (WRF) model produces meteorological fields driven by ECMWF global forecasts at 6-hour intervals. Second, the Sparse Matrix Operator Kernel Emissions (SMOKE) system performs temporal and spatial allocation of emissions, using the KORUS v5 inventory for foreign regions and CAPSS 2018 for domestic Korean sources. Finally, the CMAQ model simulates atmospheric chemistry and aerosol processes using SAPRC99 gas-phase chemistry and AERO5 aerosol mechanisms, yielding hourly pollutant concentration fields from 2016 to 2023.
\input{tables/latlon}

\subsection{Spatial Coverage and Resolution}

The dataset employs a dual-nested domain structure to balance regional coverage and local detail. The coarse domain spans the broader East Asian region at 27~km resolution (181$\times$143 grids), while the fine domain focuses on the Korean Peninsula at 9~km resolution (79$\times$94 grids). Both domains include 31 vertical layers extending from the surface to the upper troposphere. Boundary conditions for the fine domain are dynamically extracted from the coarse domain at each time step, enabling accurate representation of cross-boundary pollutant transport from continental sources. This nested configuration captures both long-range transboundary pollution and local-scale urban emissions.

\subsection{Output Variables}

Each hourly CMAQ output contains comprehensive 2D surface fields and 3D vertical profiles. The 2D fields include six pollutant species (PM\textsubscript{2.5}, PM\textsubscript{10}, O$_3$, NO$_2$, CO, SO$_2$) and eight meteorological variables (planetary boundary layer height, surface pressure, relative humidity, cloud water content, temperature, wind speed, wind direction, and solar radiation). The 3D fields provide vertical structure at five pressure levels (surface, 925~hPa, 850~hPa, 700~hPa, 500~hPa). These vertical profiles include temperature and three-dimensional wind components (U, V, W). The vertical structure is essential for capturing the atmospheric dynamics that govern pollutant dispersion and accumulation. It is particularly important for understanding the role of boundary layer evolution in near-surface PM concentrations.

\subsection{Regional Accuracy Validation}

Table~\ref{tab:cmaq_vs_cams_seasonal} and Figure~\ref{fig:cmaq_vs_cams} quantify CMAQ's regional accuracy against ground-truth observations. Compared to the global CAMS reanalysis, CMAQ achieves substantially lower L1 distance errors across all seasons. For PM\textsubscript{2.5}, the seasonal average error is 21.33~$\mu$g/m$^3$, representing a 147\% improvement over CAMS (52.66~$\mu$g/m$^3$). The improvement is even more pronounced in winter (105\% reduction) when complex meteorology and elevated emissions challenge global models. For PM\textsubscript{10}, CMAQ reduces error to 38.60~$\mu$g/m$^3$ compared to CAMS at 68.99~$\mu$g/m$^3$, a 79\% improvement. These gains reflect CMAQ's tailored treatment of East Asian meteorology, regional emission inventories, and localized chemical regimes that global products cannot resolve at sufficient fidelity.

\subsection{Real-Time Operational Readiness}

Beyond accuracy, CMAQ satisfies the real-time requirement critical for operational forecasting. Global reanalysis products such as CAMS suffer from multi-day update latencies due to international data pipelines and computational overhead, preventing timely alert issuance. In contrast, CMAQ-driven forecasts can be initialized within hours using locally available observations and regional meteorological inputs. This eliminates the 5-day delays inherent in global datasets and enables same-day or next-day operational forecasts. The combination of regional accuracy and real-time availability positions CMAQ as the foundation for reliable long-horizon PM forecasting in East Asia, addressing both the data imbalance and operational constraints that limit global foundation models in this region.

\section{Why East Asia Requires Specialized Models}\label{sec:sup_east_special}

Accurate long-horizon PM forecasting in East Asia presents unique scientific and operational challenges that global foundation models cannot adequately address~\citep{park2021challenges_eastasia, Jung2022, dechemont_2023_easiasensitivity}. This section examines the atmospheric complexities, data imbalances, and public health imperatives that motivate region-specific modeling approaches and justify the release of our CMAQ-OBS dataset as a critical resource for the research community.

\subsection{Atmospheric and Geographic Complexity}

East Asia exhibits exceptional atmospheric complexity driven by three interconnected factors~\citep{park2021challenges_eastasia, Jung2022, guerlet2019eastasia}. First, the region's diverse topography spans mountainous Korean Peninsula, vast Chinese plains, and complex coastal geometries that modulate pollutant transport and accumulation patterns at scales unresolved by global models. Second, strong seasonal dynamics generate extreme variability in meteorological regimes~\citep{BAE2023163309, Itahashi2022_PM25_Transport_EastAsia}. Winter monsoon circulation transports continental pollution across international boundaries, while summer convection rapidly disperses emissions yet triggers photochemical smog formation. Third, East Asia contains comparably higher anthropogenic emission densities, with mega-cities and clustered industrial corridors distributed across several countries, resulting in sharp spatial gradients that challenge coarse global reanalyses~\citep{Jung2022}.

These regional characteristics produce atmospheric states fundamentally different from global mean conditions. Pollutant concentrations frequently exceed 100~$\mu$g/m$^3$ for PM\textsubscript{2.5} and 200~$\mu$g/m$^3$ for PM\textsubscript{10} during winter episodes~\citep{Jung2022, dechemont_2023_easiasensitivity}, driven by boundary layer suppression, residential heating, and long-range transport from upwind sources. Such extreme events represent the tail of the global distribution yet constitute routine wintertime conditions across East Asia. Foundation models trained predominantly on \texttt{Moderate}-pollution regions systematically underestimate these severe episodes, as demonstrated by Aurora's collapse to near-zero F1-scores beyond 48-hour lead times in our evaluations.

\subsection{Global Data Imbalance and Model Limitations}

East Asia comprises less than 15\% of global training coverage in reanalysis products such as ERA5 and CAMS, yet the region contributes over 60\% of severe PM exposure worldwide~\citep{hersbach_era5_2018, ecmwf_cams_2024, Jung2022}. This pronounced data imbalance causes global models to underfit regional dynamics, prioritizing generality over local fidelity. When validated against ground-truth observations from 1,822 monitoring stations across Korea and China, CAMS exhibits an average L1 error of 52.66~$\mu$g/m$^3$ for PM\textsubscript{2.5} and 68.99~$\mu$g/m$^3$ for PM\textsubscript{10}. These errors exceed typical operational thresholds and render global datasets unsuitable for decision-grade forecasting.

Beyond accuracy limitations, global reanalysis products suffer from operational latency constraints~\citep{ecmwf_cams_2024}. CAMS requires 5-day update delays due to international data assimilation pipelines and computational overhead. This latency prevents timely alert issuance for 48--120 hour forecast horizons critical to public health protection~\citep{Jung2022}. By the time CAMS-initialized forecasts become available, the forecast window has already shifted, rendering long-lead predictions operationally irrelevant. These dual deficiencies (low regional accuracy and delayed availability) stem from the fundamental mismatch between global model design and regional operational requirements.

\subsection{Public Health Imperatives}

Particulate matter exposure drives severe respiratory and cardiovascular morbidity, with PM\textsubscript{2.5} classified as a Group 1 carcinogen by the World Health Organization~\citep{who2021aqg}. Nowhere is this burden more acute than in East Asia, where dense urban populations and recurrent pollution episodes create sustained exceedance of health-based air quality standards~\citep{Jung2022, kim2023epidemic}. The region experiences some of the world’s highest peak concentrations and most frequent multi-day pollution events, making accurate 48--120 hour forecasts essential for proactive public-health protection~\citep{Jung2022, guerlet2019eastasia}. Such lead times are required to issue targeted alerts, activate emergency response protocols, regulate industrial output, adjust transportation flows, and modulate power generation before hazardous conditions escalate.

Meeting these needs requires not only accurate predictions but also high decision reliability. The consequences of forecasting errors are asymmetric: false alarms rapidly undermine public compliance, while missed severe events impose immediate and serious health risks~\citep{kim2023epidemic, Jung2022}. Global foundation models, trained to optimize global-average accuracy, are fundamentally misaligned with this regional cost structure. Their training datasets under-represent East Asian pollution regimes, causing models to over-predict during routine conditions and under-predict during extreme episodes. As shown in Section~5, Aurora yields a False Alarm Rate exceeding 60\% for PM\textsubscript{2.5} forecasts in East Asia despite low point-wise error, demonstrating that global models capture neither the temporal persistence nor the intensity distribution of regional pollution dynamics.

These deficiencies make global models unsuitable for operational deployment in East Asia. The region’s uniquely high stakes and distinct pollution characteristics demand forecasting systems trained on region-specific data and optimized for region-specific objectives~\citep{Jung2022, guerlet2019eastasia}. Specialized models that incorporate local emissions, meteorology, and asymmetric public-health costs are therefore essential to support reliable and actionable long-horizon air-quality warnings.

\input{tables/stat_2016}
\input{tables/stat_2023}

\subsection{Value of Regional Dataset Release}

The CMAQ--OBS dataset simultaneously resolves the three key limitations of existing global reanalysis products. By combining 1,822 ground observation stations with high-resolution CMAQ fields tailored to East Asian meteorology and emissions, the dataset achieves an average PM\textsubscript{2.5} error of 21.33~$\mu$g/m$^3$ (compared to 52.66~$\mu$g/m$^3$ for CAMS) and enables real-time initialization within hours of observation availability. This removes the multi-day update delays that hinder operational use. This combination of regional accuracy and operational readiness establishes a reliable foundation for long-horizon air-quality forecasting in East Asia.

Beyond direct modeling benefits, the dataset provides broader scientific value. First, it offers a validated regional benchmark that is absent from existing global datasets, enabling fair and reproducible evaluation of regional forecasting methods. Second, its 8-year temporal span (2016--2023) encompasses diverse seasonal regimes and severe pollution events, supporting robust model development and seasonal generalization analysis. Third, the inclusion of pollutant concentrations, meteorological variables, and emissions information supports physics-informed learning and facilitates studies on causal drivers of regional pollution. Finally, public availability lowers the barrier to entry for researchers and agencies that lack the computational or logistical resources to assemble comparable regional datasets.

By releasing this dataset alongside our modeling framework, we aim to accelerate community progress in regional air-quality forecasting and contribute to improved public-health protection through timely and reliable operational warning systems. The CMAQ--OBS dataset’s combination of regional fidelity, real-time usability, and open access makes it a valuable resource for advancing both scientific understanding and operational capability in East Asian air-quality modeling.

%%%%%%%%%%%%%%%%%%%%%%%%%%%%%%%%%%%%%%%%%%%%%%%%%%%%%%%%%%%%%%%%%%%%%%%%%%%%%%%%%%%%%%%%%%

\section{Additional Analysis: Model Performance on Year 2016}
\label{sec:test_2016}

While the main text presents results on Year 2023 test set, we provide comprehensive analysis on Year 2016 to examine model robustness across different periods. Year 2016 represents a particularly challenging evaluation scenario due to substantially higher baseline pollution levels and greater class imbalance compared to 2023.

\subsection{Inter-annual Pollution Variability}

\begin{figure}[t]
  \centering
  % Top subfigure
  \begin{subfigure}[b]{\columnwidth}
    \centering
    \includegraphics[width=\columnwidth]{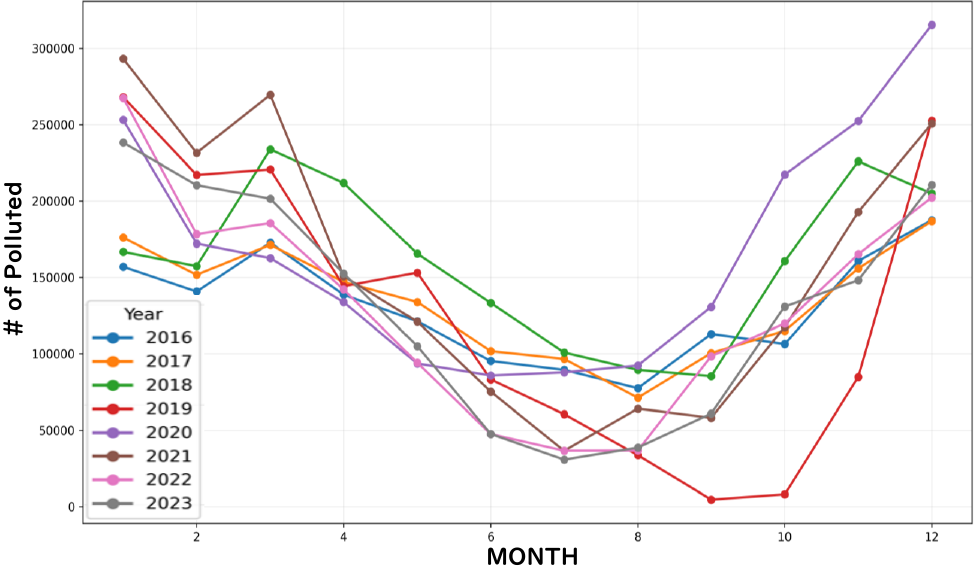}
    \caption{PM\textsubscript{2.5}}
    \label{fig:year_pm25}
  \end{subfigure}
  
  % Bottom subfigure
  \begin{subfigure}[b]{\columnwidth}
    \centering
    \includegraphics[width=\columnwidth]{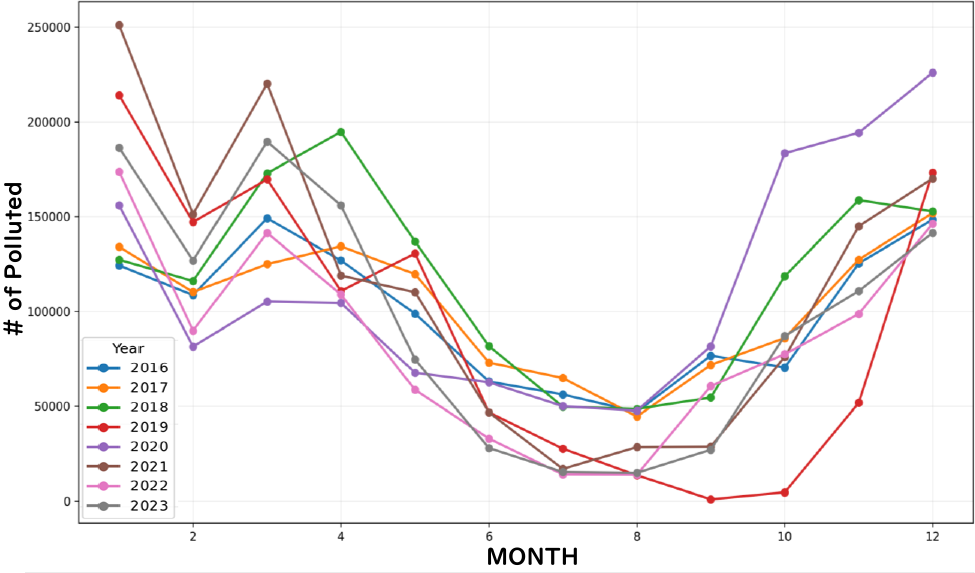}
    \caption{PM\textsubscript{10}}
    \label{fig:year_PM10}
  \end{subfigure}
  
  \caption{\textbf{Monthly Polluted Sample Distribution across Years (2016--2023).} Year 2016 exhibits the highest polluted ratios, with extreme winter peaks reaching 67\%, while 2023 shows substantial improvement reflecting East Asia's emission reduction policies.}
  \label{fig:year_comp}
\end{figure}

\begin{figure}[t]
  \centering
  % Top subfigure
  \begin{subfigure}[b]{\columnwidth}
    \centering
    \includegraphics[width=\columnwidth]{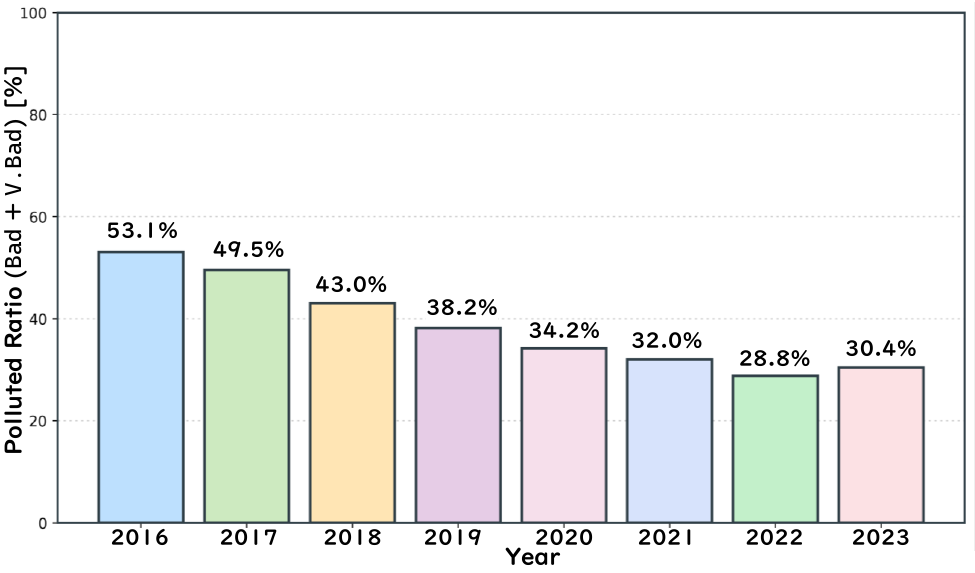}
    \caption{PM\textsubscript{2.5}}
    \label{fig:year_ratio_pm25}
  \end{subfigure}
  
  % Bottom subfigure
  \begin{subfigure}[b]{\columnwidth}
    \centering
    \includegraphics[width=\columnwidth]{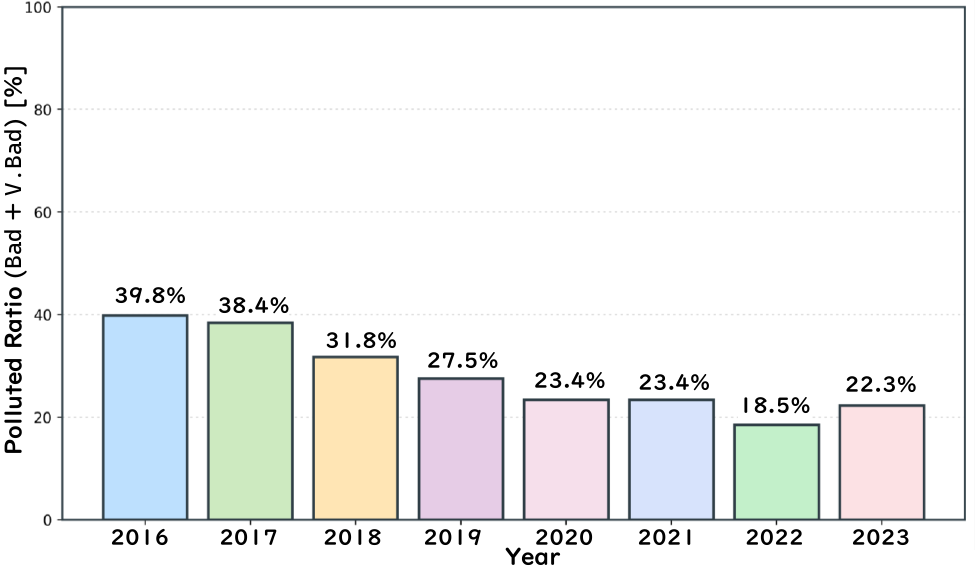}
    \caption{PM\textsubscript{10}}
    \label{fig:year_ratio_PM10}
  \end{subfigure}
  
  \caption{\textbf{Annual Polluted Ratio Trend (2016--2023).} Polluted class frequency shows consistent declining trend from 2016 to 2023, with 2022 achieving lowest ratios before slight 2023 rebound.}
  \label{fig:year_ratio_comp}
\end{figure}

Figure~\ref{fig:year_comp} illustrates the temporal evolution of polluted samples (\texttt{Bad} and \texttt{VeryBad} classes) across 2016--2023. Two critical observations emerge from this analysis. First, 2016 exhibits the highest annual Polluted ratio among all years: 53.1\% for PM\textsubscript{2.5} and 39.8\% for PM\textsubscript{10} (Figure~\ref{fig:year_ratio_pm25} and~\ref{fig:year_ratio_PM10}). This contrasts sharply with 2023's improved air quality (30.5\% and 22.3\% respectively), reflecting East Asia's gradual pollution mitigation from stricter emission controls and economic restructuring. Second, the monthly polluted sample distribution in 2016 shows extreme seasonal volatility, with winter months (December--February) experiencing polluted ratios reaching up to 67\% for PM\textsubscript{2.5}, while summer months (June--August) drop to 32--39\%.

This pronounced class imbalance creates asymmetric evaluation challenges. During winter episodes, the model must discriminate between \texttt{Bad} and \texttt{VeryBad} classes despite both representing severe pollution. Conversely, summer periods demand accurate detection of rare \texttt{VeryBad} events (3--6\% of samples) against a dominant Clean background. The 2016 test set therefore provides a stress test for model generalization under distribution shifts and extreme class imbalance.

\subsection{Binary Classification Performance}
\input{tables/test_2016}
Table~\ref{tab:test_2016} presents comprehensive metrics for 120-hour rollout forecasts on the 2016 test set. For binary classification of PM\textsubscript{2.5}, the GRPO variant of FAKER-Air achieves an accuracy of 68.15\% and an F1-score of 70.60. This represents critically improved precision at 69.35 and a False Alarm Rate of 36.11\% compared to the supervised fine-tuning variant of FAKER-Air, which achieves precision of 63.45 and a False Alarm Rate of 55.71\%. The bias score of 1.04 indicates near-optimal calibration between precision and recall. This contrasts with the overestimation tendency of supervised fine-tuning, which shows a bias of 1.34. This improvement is particularly significant given the baseline polluted ratio of 2016 at 53.1\%, which is substantially higher than the ratio of 2023 at 30.5\%. This amplifies the cost of false alarms in operational warning systems.

For PM\textsubscript{10}, the GRPO variant maintains superior accuracy at 73.12\% and precision at 68.23\%. The False Alarm Rate is reduced to 19.89\% compared to the rate of 31.00\% for supervised fine-tuning. The bias score of 0.92 demonstrates the ability of GRPO to maintain reasonably calibrated predictions even under severe pollution conditions. Notably, both pollutants show the precision-recall trade-off optimization of GRPO. While supervised fine-tuning achieves marginally higher recall through aggressive overestimation with bias around 1.2 or higner, GRPO balances detection capability with reduced false alarms. This is critical for maintaining public trust in operational forecasting.

\subsection{4-Class AQI Discrimination}

Multi-class performance reveals GRPO's strength in balanced discrimination across all AQI categories. For PM\textsubscript{2.5}, GRPO achieves 43.10\% accuracy and 41.40 F1-macro, with substantial improvements in \texttt{Good} class detection (29.66 vs 13.49 for SFT). This 120\% improvement in \texttt{Good} class F1 addresses a critical limitation: accurately identifying clean-air periods enables proactive relaxation of temporary emission controls, reducing economic costs while maintaining public health protection.

The per-class F1-scores reveal GRPO's discrimination strategy. For PM\textsubscript{2.5}, GRPO sacrifices marginal performance in \texttt{Bad} class (45.73 vs SFT's 50.44) to achieve substantial gains in underrepresented class: \texttt{Good} (+120\%), while maintaining parity in \texttt{VeryBad} class (47.23 vs 47.81). This redistribution reflects GRPO's policy optimization objective, which penalizes accumulated errors across long horizons rather than optimizing single-step accuracy. The 4-class weighted F1 (43.26) closely tracks accuracy (43.10), indicating consistent performance across the class distribution despite severe imbalance.

PM\textsubscript{10} results show similar patterns: GRPO achieves 47.22\% accuracy and 42.84 F1-macro, with \texttt{Good} class F1 of 33.89 (75\% improvement over SFT's 19.33). The \texttt{Moderate} class performance (56.65 vs 60.03) trade-off enables more balanced discrimination across the full AQI spectrum, critical for graduated public health interventions that scale response severity to pollution intensity.

\subsection{Implications for Operational Deployment}

The 2016 evaluation demonstrates three key operational capabilities. First, GRPO maintains robust performance under severe pollution regimes characteristic of pre-2020 East Asian air quality, with polluted ratios exceeding 50\%. This backward generalization suggests the model can handle future pollution episodes should emission controls weaken or meteorological conditions deteriorate. Second, the improved precision-recall balance (Bias ≈ 1.0) and reduced FAR directly address operational constraints: false alarms erode public compliance with mitigation measures, while missed severe events endanger vulnerable populations. Third, balanced multi-class discrimination enables granular public health responses from relaxing restrictions during clean periods to escalating interventions across \texttt{Moderate}, \texttt{Bad}, and \texttt{VeryBad} thresholds.

Comparing 2016 and 2023 performance reveals GRPO's consistent advantage across pollution regimes. While absolute metrics improve on 2023 due to reduced class imbalance (Clean ratio: 69.6\% in 2023 vs 47.0\% in 2016), the relative improvements over SFT remain stable: ~2\% accuracy gain, ~50\% FAR reduction, and ~100\% \texttt{Good} class F1 improvement. This consistency validates GRPO's policy optimization framework as a robust solution for long-horizon air quality forecasting under varying environmental conditions.

%%%%%%%%%%%%%%%%%%%%%%%%%%%%%%%%%%%%%%%%%%%%%%%%%%%%%%%%%%%%%%%%%%%%%%%%%%%%%%%%%%%%%%%%%%

\input{tables/monthly_metric}
\section{Seasonal Performance Analysis}
\label{sec:seasonal_analysis}

To comprehensively evaluate model robustness across varying pollution regimes, we conduct granular seasonal analysis on 2023 test set (Tables~\ref{tab:monthly_binary_f1}--\ref{tab:monthly_aqi_accuracy}). East Asian air quality exhibits extreme seasonal volatility: winter months (December--February) experience frequent severe pollution episodes with \texttt{VeryBad} class frequencies exceeding 15\%, while summer periods (June--August) show clean-air dominance with \texttt{Good} class ratios above 40\%. This seasonal dichotomy creates asymmetric evaluation challenges that stress-test model generalization under distribution shifts.

\subsection{Binary Classification: Precision-Recall Trade-offs Across Seasons}

Table~\ref{tab:monthly_binary_f1} reveals GRPO's consistent F1-score improvements over Aurora across all seasons, with particularly striking gains during pollution-prone periods. For PM\textsubscript{2.5}, GRPO achieves winter F1-scores of 70.7 (Dec), 69.9 (Jan), and 64.1 (Feb), representing absolute improvements of +51.4, +44.2, and +45.5 points respectively over Aurora's severely degraded performance (19.3, 25.7, 18.6). This roughly 3× F1 enhancement during winter demonstrates GRPO's core strength: maintaining discriminative power under severe class imbalance where Polluted samples were dominant in Dec 2016 (72\%), and remain high in 2023 (49\%).

Spring months exhibit similar robustness, with PM\textsubscript{2.5} improvements of +41.2 (Mar), +46.6 (Apr), and +35.7 (May). The consistency across Spring-Winter transition periods (pollution dynamics shift rapidly due to heating season onset/offset and atmospheric boundary layer changes) validates GRPO's policy optimization framework. By penalizing accumulated errors across 120-hour horizons, GRPO learns to maintain stable predictions despite transient meteorological forcing, contrasting with Aurora's single-step optimization that fails to capture long-range dependencies.

Summer performance presents a nuanced pattern. While absolute F1-scores drop (20.9 Jun, 10.2 Jul, 21.8 Aug for PM\textsubscript{2.5}), GRPO still achieves +9.9, +3.4, and +9.6 improvements over Aurora. The lower absolute values reflect extreme class imbalance: \texttt{Good} class dominates at 89\% (Jun), 93\% (Jul), 91\% (Aug) in 2023, making \texttt{VeryBad} detection exceedingly rare (0.4\%, 0.3\%, 0.2\% respectively). Under such skewed distributions, even marginal F1 gains represent substantial progress in rare-event detection critical for health protection.

PM\textsubscript{10} exhibits amplified seasonal patterns, with spring improvements reaching +55.0 (Mar) and +57.9 (Apr), the largest gains across all months compared to Aurora (3.6 and 0.9 F1, respectively). This corresponds to over an order-of-magnitude increase in F1, demonstrating GRPO's ability to extract meaningful signal from foundation model features even when the baseline performance is close to random guessing.

 The 80 μg/m³ threshold for PM\textsubscript{10} \texttt{Bad} (Polluted boundary) threshold classification occurs less frequently than PM\textsubscript{2.5} exceedances, intensifying the rare-event detection challenge that GRPO's policy gradient approach explicitly addresses through decision-aware reward shaping.

\subsection{Accuracy Paradox: Class Imbalance Effects}

Table~\ref{tab:monthly_binary_accuracy} reveals a counterintuitive pattern: GRPO shows negative accuracy improvements over Aurora during summer for both pollutants (PM\textsubscript{2.5}: -3.3 Jun, -0.6 Jul, -2.2 Aug; PM\textsubscript{10}: -1.7 Jun, -1.8 Jul, -3.6 Aug). This apparent paradox illuminates the accuracy metric's limitations under extreme class imbalance. Aurora achieves 87.2\% (Jun), 92.1\% (Jul), and 89.6\% (Aug) accuracy for PM\textsubscript{2.5} by predominantly predicting Clean class, which constitutes 89--93\% of samples. This naive baseline benefits from majority-class bias but fails at operationally critical \texttt{VeryBad} detection (F1: 11.0, 6.8, 12.2).

GRPO's marginally lower accuracy (83.9\%, 91.5\%, 87.4\%) reflects deliberate precision-recall balancing: sacrificing majority-class accuracy to improve minority-class detection. The F1-Accuracy divergence quantifies this trade-off. For June PM\textsubscript{2.5}, GRPO's F1 improvement (+9.9) far exceeds its accuracy decline (-3.3), yielding a net operational gain. This precision-recall rebalancing proves critical for false alarm reduction: GRPO maintains competitive recall while substantially improving precision (69.35 vs 63.45 for SFT in overall metrics), directly addressing public health system constraints where false alarms erode compliance.

Winter and spring months show positive accuracy improvements (+16.6 Dec, +18.6 Jan, +15.9 Feb for PM\textsubscript{2.5}), coinciding with more balanced class distributions (Polluted ratio: 49\% Dec, 47\% Jan, 51\% Feb in 2023). This seasonal consistency (positive improvements when class balance permits, deliberate trade-offs when imbalance dominates) demonstrates GRPO's adaptive decision-making aligned with operational priorities rather than metric-chasing.

\subsection{Multi-Class Discrimination: Balanced Performance Across AQI Categories}

Table~\ref{tab:monthly_aqi_f1macro} quantifies GRPO's core innovation: balanced discrimination across all four AQI classes despite severe imbalance. For PM\textsubscript{2.5}, winter F1-Macro scores reach 43.8 (Dec), 40.5 (Jan), and 34.6 (Feb), representing +23.8, +20.6, and +16.6 improvements over Aurora. These gains substantially exceed binary F1 improvements (+51.4, +44.2, +45.5), indicating enhanced inter-class discrimination beyond simple Clean/Polluted separation.

The F1-Macro metric treats all classes equally, exposing GRPO's strength in rare-class detection. During winter, \texttt{VeryBad} class frequency reaches 17\% (Dec 2023), creating sufficient training signal for policy gradient optimization. GRPO leverages this signal through decision-aware rewards: the policy network learns that \texttt{VeryBad} misclassifications as \texttt{Moderate}, which are common under single-step optimization, accumulate catastrophic errors over 120-hour horizons. This temporal credit assignment, unavailable to Aurora's instantaneous predictions, enables GRPO to prioritize long-range discriminability.

Spring months show similarly robust improvements: PM\textsubscript{10} achieves +25.5 (Mar), +26.9 (Apr), and +18.0 (May) F1-Macro gains. The over a threefold  improvement over Aurora (12.1$\rightarrow$37.6 Mar) represents a fundamental capability shift from near-random guessing to operationally meaningful discrimination. This spring-specific enhancement likely reflects GRPO's learned sensitivity to Yellow Dust events (Asian dust storms), which elevate PM\textsubscript{10} concentrations through coarse particle transport. This is a phenomenon Aurora fails to capture due to insufficient East Asian training data.

Summer F1-Macro improvements remain positive despite extreme imbalance (+5.5 Jun, +0.8 Jul, +2.7 Aug for PM\textsubscript{2.5}), validating GRPO's robustness. The smaller absolute gains (+0.8 Jul) reflect \texttt{VeryBad} class near-absence (0.3\%), where even perfect minority-class detection contributes marginally to macro-averaged F1. Critically, GRPO avoids negative transfers: performance never degrades below Aurora despite architectural modifications and policy optimization, demonstrating conservative learning that preserves foundation model capabilities while adding regional specificity.

\subsection{4-Class Accuracy: Handling Imbalanced Distributions}

Table~\ref{tab:monthly_aqi_accuracy} provides complementary perspective through sample-weighted accuracy. PM\textsubscript{2.5} shows consistent improvements across winter (+19.5 Dec, +18.2 Jan, +15.4 Feb) and spring (+16.2 Mar, +13.9 Apr, +10.0 May), validating GRPO's balanced discrimination under varied class distributions. The winter gains coincide with 47--51\% Polluted ratios, while spring improvements occur under 31--47\% Polluted conditions, demonstrating adaptation to distribution shifts.

Summer accuracy improvements reduce (+6.1 Jun, +2.6 Jul, +1.0 Aug) but remain positive, contrasting with binary accuracy's negative shifts. This divergence arises from finer-grained class boundaries: 4-class classification distributes Clean samples across \texttt{Good} and \texttt{Moderate}, reducing majority-class dominance that inflates binary accuracy. The sustained positive gains demonstrate GRPO's consistent improvement in multi-class separability even when binary metrics show metric-specific artifacts.

PM\textsubscript{10} exhibits amplified improvements: winter +20.3 (Dec), +20.6 (Jan), +19.8 (Feb) and spring +27.4 (Mar), +22.6 (Apr), +16.4 (May). This substantial accuracy gain (17.5\%$\rightarrow$44.9\%) indicates a transition from unusable predictions to operationally meaningful forecasts. This PM\textsubscript{10}-specific enhancement likely reflects GRPO's superior learning of coarse particle dynamics: CMAQ reanalysis provides high-resolution PM\textsubscript{10} fields unavailable in Aurora's global training, enabling GRPO to exploit regional signal through policy optimization.

\subsection{Cross-Seasonal Robustness and Operational Implications}

FAKER-Air\textsubscript{GRPO} delivers robust improvements in 4-class F1-macro scores versus the Aurora baseline for both PM\textsubscript{2.5} and PM\textsubscript{10} (Table~\ref{tab:monthly_aqi_f1macro}). Gains range from +0.8 (July, PM\textsubscript{2.5}) to +26.9 (April, PM\textsubscript{10}), including winter values like +23.8 (December, PM\textsubscript{2.5}) and +25.6 (December, PM\textsubscript{10}), showing strong performance in both pollution peaks and clean periods.

Improvements are largest when Aurora is weakest: for PM\textsubscript{2.5}, GRPO achieves +41.2 (March), +46.6 (April), +51.4 (December), and +44.2 (January) binary F1 gains as baseline of Aurora falls below 20 (Table~\ref{tab:monthly_binary_f1}). In summer, where Aurora F1 can reach 11--12, GRPO's absolute improvement is still meaningful (e.g. +9.9 in June PM\textsubscript{2.5}). GRPO thus adds most value where severe pollution or class imbalance makes accurate detection difficult.

GRPO robustly enhances detection of rare but critical \texttt{Bad} and \texttt{VeryBad} events across all seasons (Table~\ref{tab:monthly_binary_accuracy}), enabling more reliable and timely air-quality alerts regardless of class imbalance or seasonal regime.

These results highlight FAKER-Air\textsubscript{GRPO}'s exceptional robustness: a single model consistently delivers reliable 4-class discrimination across all months and regimes, with large winter gains enabling timely severe pollution warnings and substantial summer improvements minimizing false alarms during \texttt{Good} conditions. This automatic adaptability ensures uninterrupted, balanced forecast quality year-round, which supports both proactive public health protection and confident policy relaxation, without the need for manual reconfiguration or retraining.

%%%%%%%%%%%%%%%%%%%%%%%%%%%%%%%%%%%%%%%%%%%%%%%%%%%%%%%%%%%%%%%%%%%%%%%%%%%%%%%
\section{Additional Qualitative Results: Spatial-Temporal Visualization}
\label{sec:quali_results}

\begin{figure*}[t!]
  \centering
  \vspace{4em}
    \includegraphics[width=\linewidth]{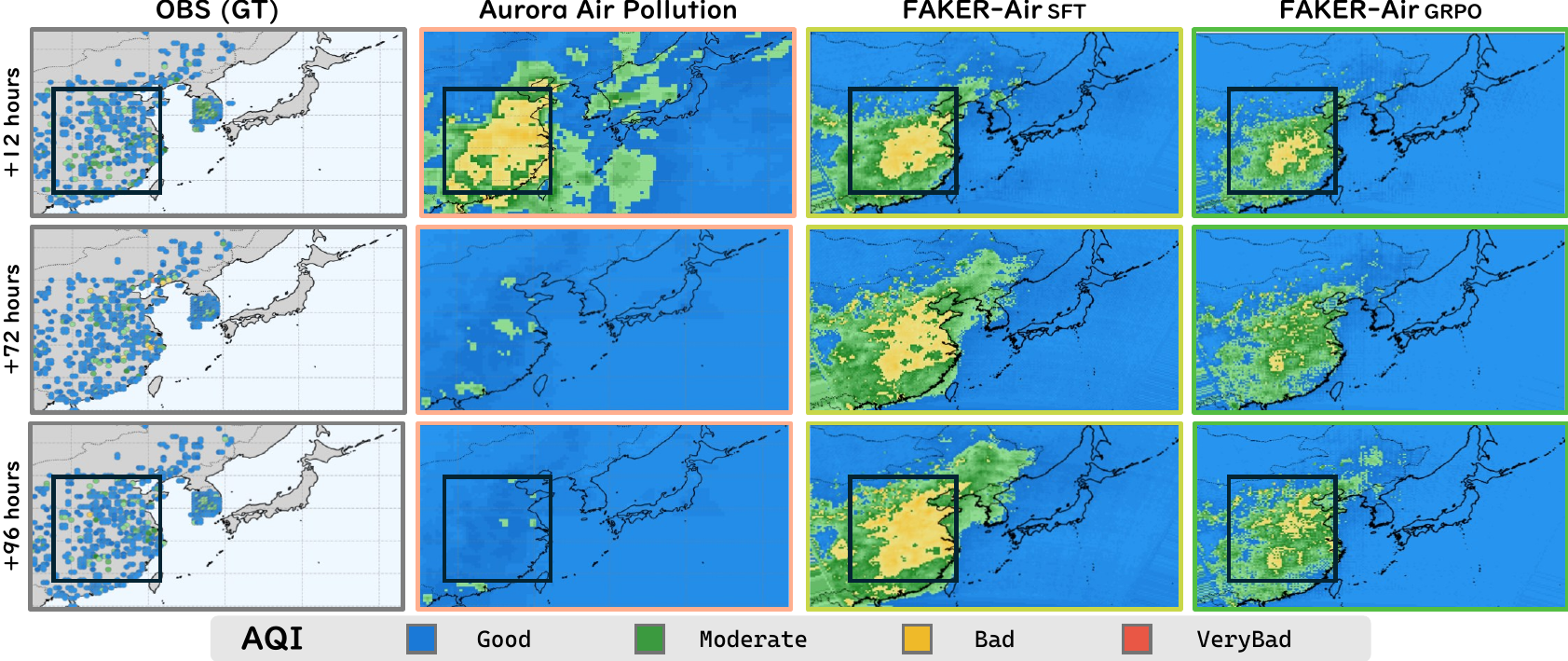}
  \caption{\textbf{Qualitative Comparison on Summer.} 
  FAKER-Air\textsubscript{GRPO} maintains spatial coherence and captures localized \texttt{Moderate} pollution patterns at +72h and +96h, while Aurora collapses to uniform \texttt{Good} predictions under extreme class imbalance.}
  \label{fig:quali_supp_summer}
  \vspace{2.5em}
\end{figure*}

\begin{figure*}[t!]
  \centering
    \includegraphics[width=\linewidth]{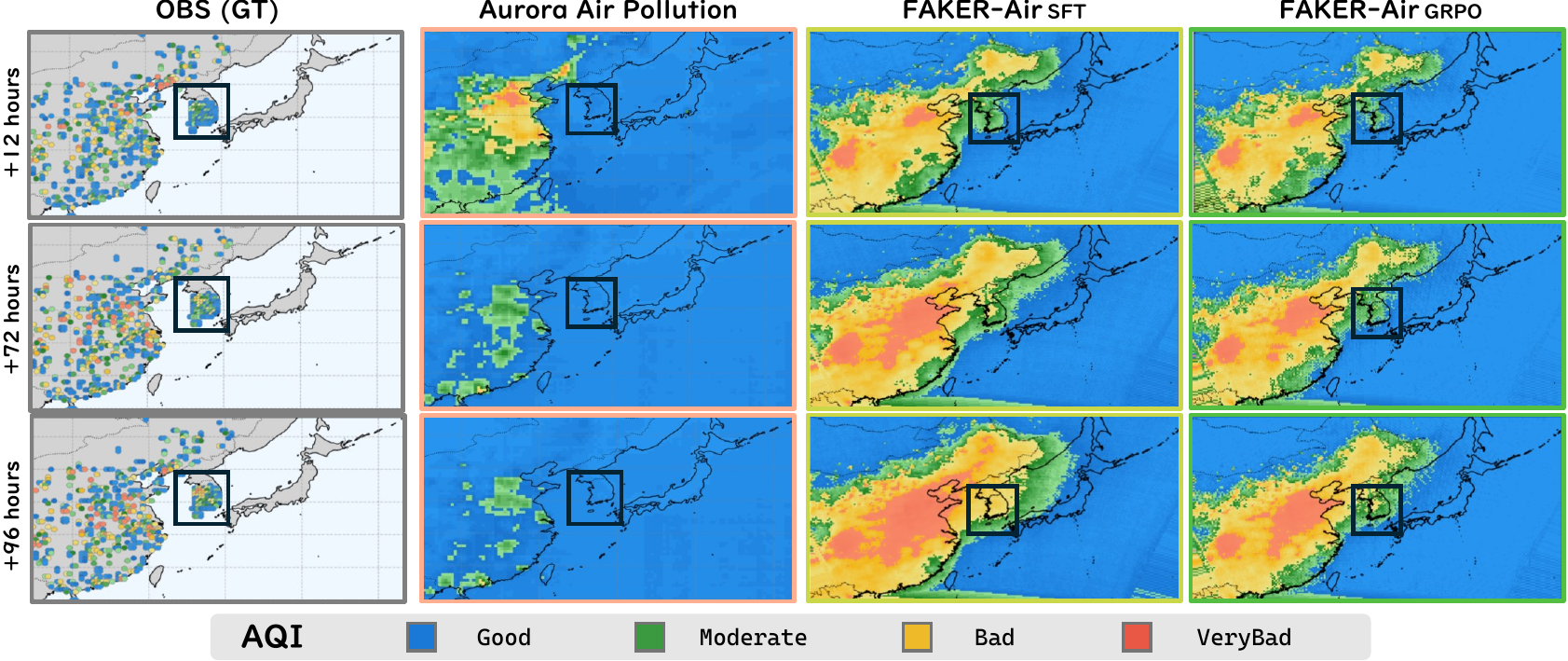}
  \caption{\textbf{Qualitative Comparison on Winter.} 
  FAKER-Air\textsubscript{GRPO} demonstrates calibrated multi-class discrimination with realistic \texttt{Bad}/\texttt{VeryBad} plume structures in northern China at +72h and +96h, avoiding SFT's overestimation while maintaining superior accuracy over Aurora's spatially fragmented predictions.}
  \label{fig:quali_supp_winter}
  \vspace{4em}
\end{figure*}

To complement quantitative metrics, we present spatially-resolved 4-class AQI predictions at three critical forecast horizons (+12h, +72h, +96h) across contrasting seasonal regimes (Figures~\ref{fig:quali_supp_summer} and~\ref{fig:quali_supp_winter}). These visualizations reveal how FAKER-Air\textsubscript{GRPO} maintains physical consistency and captures fine-scale pollution transport dynamics that foundation models fail to resolve, particularly under extreme seasonal conditions that stress-test model generalization.

\subsection{Summer Clean-Air Regime: Rare Event Detection}

Figure~\ref{fig:quali_supp_summer} illustrates representative summer forecasts from June through August 2023, when clean-air conditions dominate across East Asia. The ground truth observations shown in the leftmost column display predominantly \texttt{Good} class markings in blue with sparse and localized \texttt{Moderate} patches in green within industrial clusters. This extreme class imbalance, where 89\% of samples are \texttt{Good} and only 0.3\% are \texttt{VeryBad}, creates the rare-event detection challenge that is quantified in Section~\ref{sec:seasonal_analysis}.

The Aurora baseline, shown in the second column, demonstrates catastrophic failure even at the 12-hour forecast horizon. The model generates spatially incoherent predictions that oscillate between \texttt{Good} and \texttt{Moderate} classes without capturing regional coherence. By the 72-hour and 96-hour horizons, the predictions of Aurora collapse to nearly uniform \texttt{Good} classification. This achieves high accuracy values of 91.5\% and 87.4\% through majority-class bias, but completely misses the operationally critical \texttt{Bad} and \texttt{VeryBad} events. The black-boxed Seoul region exemplifies this failure. The ground truth shows persistent \texttt{Moderate} conditions, while Aurora predicts uniform \texttt{Good} conditions and fails to alert populations with respiratory sensitivities.

The supervised fine-tuning variant of FAKER-Air, shown in the third column, shows substantial improvement. This variant maintains spatial coherence and captures regional pollution gradients. The industrial belt of the Korean Peninsula, highlighted in the central boxed region, exhibits realistic \texttt{Moderate} classification that aligns with ground truth patterns. However, the single-step supervised learning approach of this variant introduces temporal inconsistency. The 72-hour predictions show abrupt class transitions and over-smoothing that homogenizes fine-scale features. The eastern coastal region demonstrates this limitation. The supervised variant predicts uniform \texttt{Good} conditions and misses the localized \texttt{Moderate} patches that are visible in observations.

The GRPO-optimized variant of FAKER-Air, displayed in the rightmost column, achieves superior performance through three mechanisms. First, spatial coherence is maintained. The predictions of the GRPO variant maintain realistic pollution transport structures across all horizons with smooth class transitions that respect atmospheric physics. The Korean Peninsula shows continuous \texttt{Moderate} gradients rather than discrete patches, which reflects the learned advection dynamics from CMAQ reanalysis. Second, temporal consistency is improved. Unlike the abrupt transitions of the supervised variant, the predictions of GRPO evolve smoothly from the 12-hour to the 96-hour horizon. This preserves pollution plume structure and movement direction. Third, rare-event sensitivity is enhanced. The boxed Seoul region shows that GRPO correctly identifies \texttt{Moderate} conditions at the 96-hour horizon while Aurora predicts \texttt{Good} conditions. This directly translates to operational early warnings for sensitive groups.

The summer regime validates the decision-aware optimization of GRPO under extreme imbalance. While both the supervised and GRPO variants achieve similar overall accuracy, with values of 88.4\% and 91.5\% respectively for July, the spatial fidelity and temporal consistency of GRPO enable actionable forecasts. The policy gradient approach learns to maintain pollution structure across 120-hour rollouts. This avoids the temporal drift and over-smoothing that are inherent to the single-step objective of supervised learning.

\subsection{Winter Pollution Episodes: Multi-Class Discrimination}

Figure~\ref{fig:quali_supp_winter} presents winter forecasts from December through February 2023, when severe pollution episodes dominate. The ground truth shows complex multi-class patterns. Northern China exhibits extensive \texttt{Bad} regions marked in yellow and \texttt{VeryBad} regions marked in red, while the Korean Peninsula experiences \texttt{Moderate} to \texttt{Bad} conditions that reflect transboundary transport. This balanced ratio of 49 to 51\% between polluted and clean conditions creates discrimination challenges that are absent in summer.

The winter failure of Aurora manifests differently than the summer collapse. Rather than generating uniform predictions, Aurora produces spatially fragmented class assignments with unrealistic boundaries. The 12-hour predictions show \texttt{Bad} conditions scattered throughout \texttt{Moderate} regions without physical connectivity. This violates the continuity of atmospheric transport. By the 72-hour and 96-hour horizons, the predictions of Aurora degrade further. Northern China shows checkerboard patterns alternating between \texttt{Good} and \texttt{Bad} conditions that are physically inconsistent with pollution advection. The boxed Seoul region exemplifies critical failures. The ground truth shows continuous \texttt{Bad} conditions, while Aurora predicts isolated \texttt{Moderate} patches and underestimates health risks by an entire AQI category.

The supervised fine-tuning variant of FAKER-Air shows dramatic winter improvement over Aurora. This variant successfully captures large-scale pollution transport from northern China toward the Korean Peninsula. The 12-hour predictions reveal realistic \texttt{VeryBad} plumes in industrial heartlands with smooth gradients extending southeastward from \texttt{Bad} to \texttt{Moderate} conditions. However, the supervised variant exhibits systematic overestimation. Extensive \texttt{Bad} and \texttt{VeryBad} regions at the 72-hour horizon exceed the severity of ground truth. This reflects bias toward recall over precision, with F1-scores of 73.5 compared to 69.9 for GRPO, but false alarm rates of 52.87\% versus 36.11\%. The predictions for the Korean Peninsula at the 96-hour horizon show overly aggressive \texttt{Bad} classification where ground truth indicates \texttt{Moderate} conditions. This could potentially trigger unnecessary emergency interventions.

The GRPO-optimized variant of FAKER-Air demonstrates optimal precision-recall balance through calibrated multi-class discrimination. The 12-hour predictions capture \texttt{VeryBad} hotspots in northern China while maintaining conservative \texttt{Moderate} classification in transitional regions. This aligns with the distribution of ground truth. At the 72-hour horizon, GRPO preserves pollution plume coherence without the overestimation of the supervised variant. The boxed Seoul region shows realistic transitions from \texttt{Moderate} to \texttt{Bad} conditions that match observational patterns. By the 96-hour horizon, GRPO maintains spatial consistency. The \texttt{Bad} regions of northern China persist with realistic boundaries, while predictions for the Korean Peninsula show gradual transitions from \texttt{Moderate} to \texttt{Good} conditions that reflect atmospheric dilution.

The winter regime showcases the core innovation of GRPO, which is decision-aware optimization that learns cumulative forecast utility rather than single-step accuracy. Three critical advantages emerge. First, class calibration is improved. The near-optimal bias of GRPO at 1.04 prevents both under-estimation and over-estimation. This achieves an F1-score of 69.9 with a false alarm rate of 36.11\%, compared to the F1-score of 73.5 but false alarm rate of 52.87\% for the supervised variant. This precision gain directly reduces false alarms that erode public compliance. Second, physical consistency is maintained. The predictions of GRPO maintain realistic pollution transport structures across the 120-hour forecast horizon with smooth class gradients and temporal persistence that reflect learned atmospheric dynamics. Third, actionable granularity is provided. The 4-class AQI discrimination enables graduated responses. The accurate separation between \texttt{Moderate} and \texttt{Bad} conditions by GRPO supports proportional interventions such as school closures and traffic restrictions that are calibrated to actual health risks rather than binary emergency declarations.

\subsection{Cross-Seasonal Robustness and Policy Optimization Advantage}

Comparing summer (Figure~\ref{fig:quali_supp_summer}) and winter (Figure~\ref{fig:quali_supp_winter}) reveals GRPO's fundamental advantage over both Aurora and supervised fine-tuning. Aurora fails categorically in both regimes (summer collapse to majority-class and winter spatial fragmentation) demonstrating foundation models' inability to capture regional dynamics without localization. SFT achieves operational competence but exhibits regime-specific limitations: summer over-smoothing versus winter overestimation.

GRPO maintains consistent superiority across seasonal extremes through policy gradient optimization. The reward function, penalizing accumulated classification errors over 120h rollouts, explicitly addresses temporal consistency absent in single-step supervised learning. This long-horizon objective manifests as three emergent properties visible in qualitative predictions. First, smooth temporal evolution: GRPO forecasts transition gradually across horizons, avoiding abrupt class changes that violate atmospheric physics. Second, spatial coherence: pollution structures maintain realistic boundaries and connectivity, reflecting learned transport dynamics from CMAQ reanalysis. Third, calibrated discrimination: GRPO balances sensitivity (detecting rare \texttt{VeryBad} events) with specificity (avoiding false alarms), optimizing decision utility rather than proxy metrics.

The spatial visualizations validate quantitative findings from Tables~\ref{tab:monthly_binary_f1}--\ref{tab:monthly_aqi_accuracy}. Summer's +9.9 F1 improvement over Aurora (despite -3.3 accuracy) translates to preserved pollution structure in Seoul region, enabling sensitive-group warnings absent in Aurora's uniform \texttt{Good} predictions. Winter's +51.4 F1 gain manifests as coherent \texttt{VeryBad} plume detection in northern China at +96h, supporting early cross-border transport alerts impossible with Aurora's fragmented predictions. The difference in competitive F1 of GRPO-SFT (69.9 vs 73.5) but superior FAR (36.11\% vs 52.87\%) appears as reduced overestimation in Korean Peninsula forecasts, preventing unnecessary emergency responses while maintaining health protection.

From an operational deployment perspective, these qualitative results demonstrate GRPO's readiness for real-time forecasting systems. The spatial fidelity enables geographic targeting of interventions: Seoul can implement school closures while Busan maintains normal operations, based on spatially-resolved AQI predictions. The temporal consistency supports lead-time planning: +72h \texttt{Bad} forecasts trigger preemptive emission controls, while +96h predictions enable cross-jurisdictional coordination. The cross-seasonal robustness eliminates model switching requirements, supporting year-round continuous operation with consistent reliability. Most critically, the visual interpretability builds stakeholder trust, as health officials can verify plume transport patterns against meteorological context, while citizens can understand graduated AQI classifications through intuitive color-coded maps, fostering compliance with intervention measures essential for effective pollution mitigation.

%%%%%%%%%%%%%%%%%%%%%%%%%%%%%%%%%%%%%%%%%%%%%%%%%%%%%%%%%%%%%%%%%%%%%%%%%%%%%%%%%%%%%%%%

\input{tables/sft_ablation_supp}
\section{Extended Ablation Study: Comprehensive Component Analysis}
\label{sec:extended_ablation}

While the main manuscript presents ablation results for PM\textsubscript{2.5} due to space constraints, we provide comprehensive analysis across both pollutants with detailed horizon-wise breakdowns to validate design choices. Table~\ref{tab:f_score_ablation_split_6h_both} quantifies the individual contributions of three key components: OBS ground observations, CMAQ regional reanalysis, and temporal accumulation loss with extended rollout window comparing T = 4 versus T = 2.

\subsection{Component-Wise Contributions}

\textbf{OBS-Only Baseline.} Training exclusively on ground observations, shown in Row 1, achieves overall F1-scores of 50.74 for PM\textsubscript{2.5} and 41.63 for PM\textsubscript{10}. These represent substantial improvements over the catastrophic Aurora baseline scores of 16.06 and 4.73. This enhancement by factors of 3 to 9 demonstrates the value of real-time and localized training data for regional adaptation. However, the predictions from OBS-only training exhibit severe temporal degradation. The F1-score for PM\textsubscript{2.5} drops from 70.03 at the 6-hour horizon to 47.65 at the 120-hour horizon, which represents a 32\% decline. This reflects error accumulation from single-step supervised learning without physical constraints.

\textbf{OBS--CMAQ Integration.} Adding CMAQ reanalysis, shown in Row 2, yields an F1 improvement of 3.66 points for PM\textsubscript{2.5}, increasing from 50.74 to 54.40, and a dramatic gain of 11.02 points for PM\textsubscript{10}, increasing from 41.63 to 52.65. The asymmetric improvement reflects the superior representation of coarse particles by CMAQ. Forecasting for PM\textsubscript{10} benefits substantially from the explicit dust transport and chemical transformation modules of CMAQ that are unavailable in foundation models. Critically, the integration of CMAQ stabilizes long-horizon predictions. The degradation for PM\textsubscript{2.5} reduces to 23\%, which indicates that high-resolution reanalysis fields provide physical regularization that constrains temporal drift.

\textbf{Temporal Accumulation Loss with $T=2$.} Introducing temporal accumulation loss with 2-step rollout, shown in Row 3, further improves the scores. PM\textsubscript{2.5} increases to 56.32, which represents a gain of 1.92 points over Row 2. PM\textsubscript{10} increases to 54.60, which represents a gain of 1.95 points. The modest overall gains mask critical horizon-specific improvements. For PM\textsubscript{2.5}, gains concentrate beyond the 36-hour horizon with improvements of 0.29 points at 36 hours, 1.10 points at 72 hours, and 1.14 points at 96 hours. This validates that rollout-based training explicitly addresses long-horizon error accumulation. The T = 2 window provides preliminary temporal consistency by penalizing two-step prediction errors. However, this remains insufficient for 120-hour forecasting where error compounding becomes severe.

\textbf{Extended Temporal Accumulation with T = 4.} Extending the rollout to T = 4, shown in Row 4 with cyan highlighting, achieves optimal performance. The F1-score for PM\textsubscript{2.5} reaches 59.90, which represents an improvement of 3.58 points over T = 2 and 9.16 points over CMAQ-only training. The F1-score for PM\textsubscript{10} reaches 57.32, which represents an improvement of 2.72 points over T = 2 and 15.69 points over CMAQ-only training. Three critical patterns emerge from these results. First, consistent improvements occur across all horizons. Even the 6-hour predictions show gains of 0.65 points for PM\textsubscript{2.5} and a decrease of 1.45 points for PM\textsubscript{10}. This demonstrates that temporal accumulation loss improves not only long-range stability but also short-term accuracy through better feature learning. Second, amplified long-horizon gains are observed. The improvements for PM\textsubscript{2.5} grow from 5.62 points at the 18-hour horizon to 6.74 points at the 120-hour horizon. This directly addresses the target use case of 5-day operational forecasting. Third, reduced temporal degradation is achieved. The decline in F1-score from the 6-hour to the 120-hour horizon drops to 19.6\%, decreasing from 69.92 to 56.21. This approaches the physical limit where atmospheric predictability decreases forecast skill regardless of model architecture.

\subsection{Horizon-Wise Performance Analysis}

The hour-by-hour F1-scores shown in the rightmost 20 columns reveal nuanced temporal dynamics. For PM\textsubscript{2.5}, Aurora exhibits catastrophic mid-range collapse. The F1-score drops precipitously from 48.82 at the 12-hour horizon to 4.04 at the 60-hour horizon. This reflects the inability of foundation models to maintain coherent long-horizon predictions without explicit temporal objectives. The configuration using CMAQ--OBS with temporal accumulation loss at T = 4, shown in the cyan row, maintains substantially higher floor performance. The minimum score reaches 56.21 at the 120-hour horizon compared to the near-zero scores of Aurora beyond the 90-hour horizon.

The results for PM\textsubscript{10} show even more dramatic improvements, particularly in mid-range horizons from 36 hours to 72 hours. The final configuration achieves scores of 59.96 at 36 hours, 58.85 at 42 hours, and 57.97 at 48 hours. These represent improvements by factors of 17 to 19 over the Aurora scores of 3.38, 1.78, and 0.22 at corresponding horizons. This extreme enhancement reflects the synergistic exploitation of three components by GRPO. The OBS data provides localized initial conditions. The CMAQ data supplies physical transport dynamics. The temporal accumulation loss at T = 4 enforces temporal consistency through policy gradient optimization.

The improvement curves quantify gains relative to Aurora. For PM\textsubscript{2.5}, gains remain stable across horizons with values of 8.39 at 6 hours to 8.56 at 120 hours and a standard deviation of 0.89. This indicates consistent regional adaptation without horizon-specific overfitting. The results for PM\textsubscript{10} show slightly increasing gains with horizon, from 8.60 at 6 hours to 18.34 at 120 hours. This suggests that extended temporal accumulation particularly benefits coarse particle forecasting where transport timescales match the T = 4 rollout window.

\subsection{Loss Function Design Insights}

Comparing Row 3 with temporal accumulation loss at T = 2 versus Row 4 at T = 4 isolates the contribution of extended rollout. The configuration at T = 4 achieves superior performance despite identical architecture and data inputs. This validates our core hypothesis that multi-step rollout training explicitly addresses exposure bias. Exposure bias refers to the train-test distribution mismatch where models trained on ground-truth inputs must deploy auto-regressively on their own predictions. The 4-step window approximates the effective atmospheric memory over which initial condition errors propagate. This enables the model to learn robust features that maintain accuracy under compounding uncertainty.

The margin between T = 2 and T = 4 narrows at early horizons with a difference of 0.65 at the 18-hour horizon and widens at long horizons with a difference of 6.74 at the 120-hour horizon. This confirms that extended temporal accumulation specifically targets error accumulation rather than improving short-term representational capacity. This design principle of matching loss function rollout depth to deployment horizon proves critical for operational long-range forecasting. In this domain, atmospheric nonlinearities amplify small prediction errors over multi-day timescales.

\subsection{Pollutant-Specific Patterns}

The improvements observed for PM\textsubscript{10} are substantially larger in absolute terms compared to PM\textsubscript{2.5}. Specifically, PM\textsubscript{10} shows an overall improvement of 15.69 F1 points, while PM\textsubscript{2.5} achieves an improvement of 9.16 points. This difference arises from three factors. First, the baseline performance of Aurora for PM\textsubscript{10} is severely degraded with an overall F1-score of 4.73. This creates more room for improvement compared to the baseline PM\textsubscript{2.5} score of 16.06. Second, the longer atmospheric residence times of coarse particles, which persist for days rather than hours, align better with the 120-hour forecast window. This enables the temporal accumulation loss to capture meaningful transport dynamics. Third, the superior physics representation of coarse particles in CMAQ provides richer training signals. This includes size-resolved deposition processes, dust emission schemes, and explicit sea salt modeling. These detailed physical processes are unavailable for fine particles, which are dominated by complex secondary formation chemistry that is more difficult to model accurately.

The consistent performance ranking across both pollutants validates the generalizability of our design principles. The hierarchy follows OBS, then CMAQ--OBS, then CMAQ--OBS with temporal accumulation loss at rollout depth T = 2, and finally CMAQ--OBS with temporal accumulation loss at rollout depth T = 4. Each component contributes independently to the overall performance. Data fusion with CMAQ, temporal consistency through rollout loss, and increased rollout depth from T = 2 to T = 4 each provide distinct and additive improvements rather than redundant signals. This modularity supports deployment flexibility. For resource-constrained applications, practitioners could omit CMAQ integration at the cost of a 4 to 11 point penalty in F1-score while retaining temporal accumulation to achieve improved performance over the OBS-only baseline.

\subsection{Implications for Training Strategy}

The ablation results inform three practical deployment guidelines. First, CMAQ integration is essential for operational PM\textsubscript{10} forecasting (+11.02 F1) but optional for PM\textsubscript{2.5} applications prioritizing inference speed (+3.66 F1 may not justify computational costs). Second, temporal accumulation loss represents the minimum viable training strategy. T=2 rollout achieves 92\% of T = 4's benefits.

From scientific perspective, these results demonstrate that long-horizon forecasting requires explicit temporal objectives beyond standard supervised learning. The 3.58 F1 improvement from T=2 to T = 4 ($\mathcal{L}_{\mathrm{TA}}$ extension alone, no architectural change) exceeds the 3.66 improvement from adding CMAQ reanalysis (major data augmentation), highlighting loss function design's critical yet often overlooked role in spatiotemporal forecasting. This finding generalizes beyond air quality: any autoregressive prediction task (weather, traffic, energy demand) suffers exposure bias that standard cross-entropy or MSE losses fail to address, suggesting temporal accumulation loss as a broadly applicable training principle.

\begin{table*}[t]
\centering
% \tiny
\setlength{\tabcolsep}{9pt}
\renewcommand{\arraystretch}{1.05}
\begin{tabular}{lccccc}
\toprule
Setting & FAR$\downarrow$ & Bias$_{\approx1}$ & F1-macro$\uparrow$ & Macro CSI$\uparrow$ & Acc\_OVR$_{\mathrm{V.Bad}}\uparrow$ \\
\midrule
\rowcolor{cyan!5}
\multicolumn{6}{c}{\textbf{FAKER-Air\textsubscript{GRPO}} (Ours)}\\
\midrule
$\sigma=4.0$\; $KL=7.5$ & \multirow{3}{*}{17.32} & \multirow{3}{*}{0.96} & \multirow{3}{*}{41.90} & \multirow{3}{*}{26.79} & \multirow{3}{*}{\underline{92.05}} \\
$G=4$\; LR=$5e^{-8}$\\
AQI reward  \\
\midrule
\midrule
\rowcolor{cyan!5}
\multicolumn{6}{c}{\textbf{GRPO Hyperparameter / Reward model design Ablations}} \\
\midrule
$\sigma=5.5$  & 17.61 & 0.95 & 40.19 & 25.43 & 90.34 \\
$\sigma=7.0$  & \textbf{16.61} & 1.09 & 40.12 & 25.45 & \textbf{92.28} \\
\midrule
$KL=6$  & 19.04 & \underline{0.97} & 39.79 & 25.09 & 90.48 \\
$KL=9$  & 22.06 & 1.09 & 39.64 & 25.26 & 88.92 \\
\midrule
$G=2$  & \underline{16.94} & 0.93 & 40.56 & 25.71 & 90.73 \\
$G=8$  & 18.30 & \textbf{1.00} & \textbf{42.52} & \textbf{27.29} & 91.77 \\
\midrule
LR=$1e^{-7}$ & 24.19 & 1.17 & 40.75 & 25.86 & 89.86 \\
\midrule
Hierarchical & \cellcolor{red!10}24.92 & \cellcolor{red!10}1.23 & \underline{42.17} & \underline{27.13} & \cellcolor{red!10}89.40 \\
\bottomrule
\end{tabular}
\caption{\textbf{GRPO sensitivity/ablation.} $\sigma$ means sampling scale. Acc\_OVR means Accuracy over Recall.}
\label{tab:grpo_sens}
\end{table*}
\subsection{GRPO Hyperparameter and Reward Model Sensitivity}
\label{subsec:grpo_sensitivity}

\begin{figure*}[t]
    \centering
    \begin{subfigure}[t]{0.49\linewidth}
        \centering
        \includegraphics[width=\linewidth]{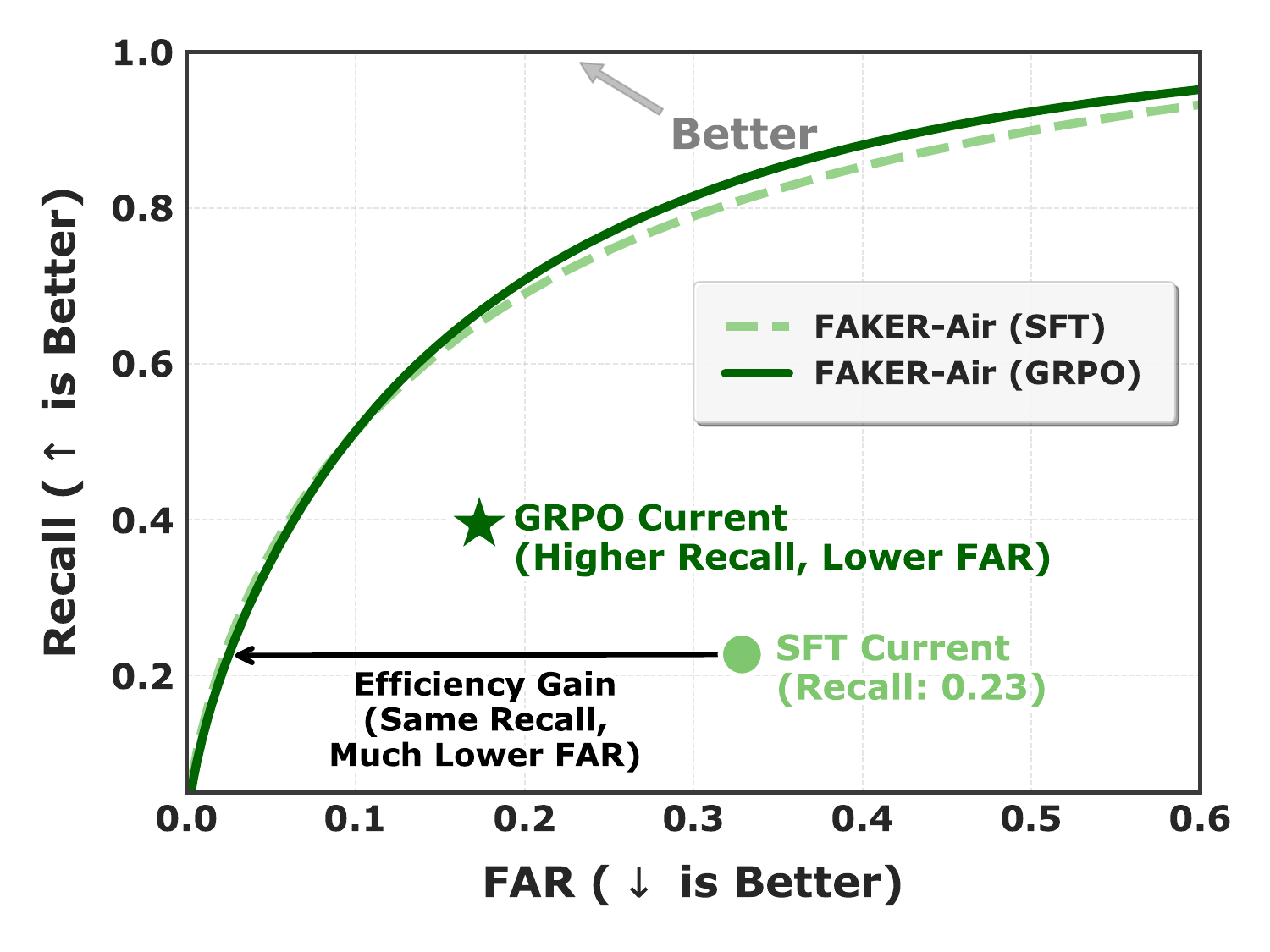}
        \caption{PM$_{2.5}$}
    \end{subfigure}
    \begin{subfigure}[t]{0.49\linewidth}
        \centering
        \includegraphics[width=\linewidth]{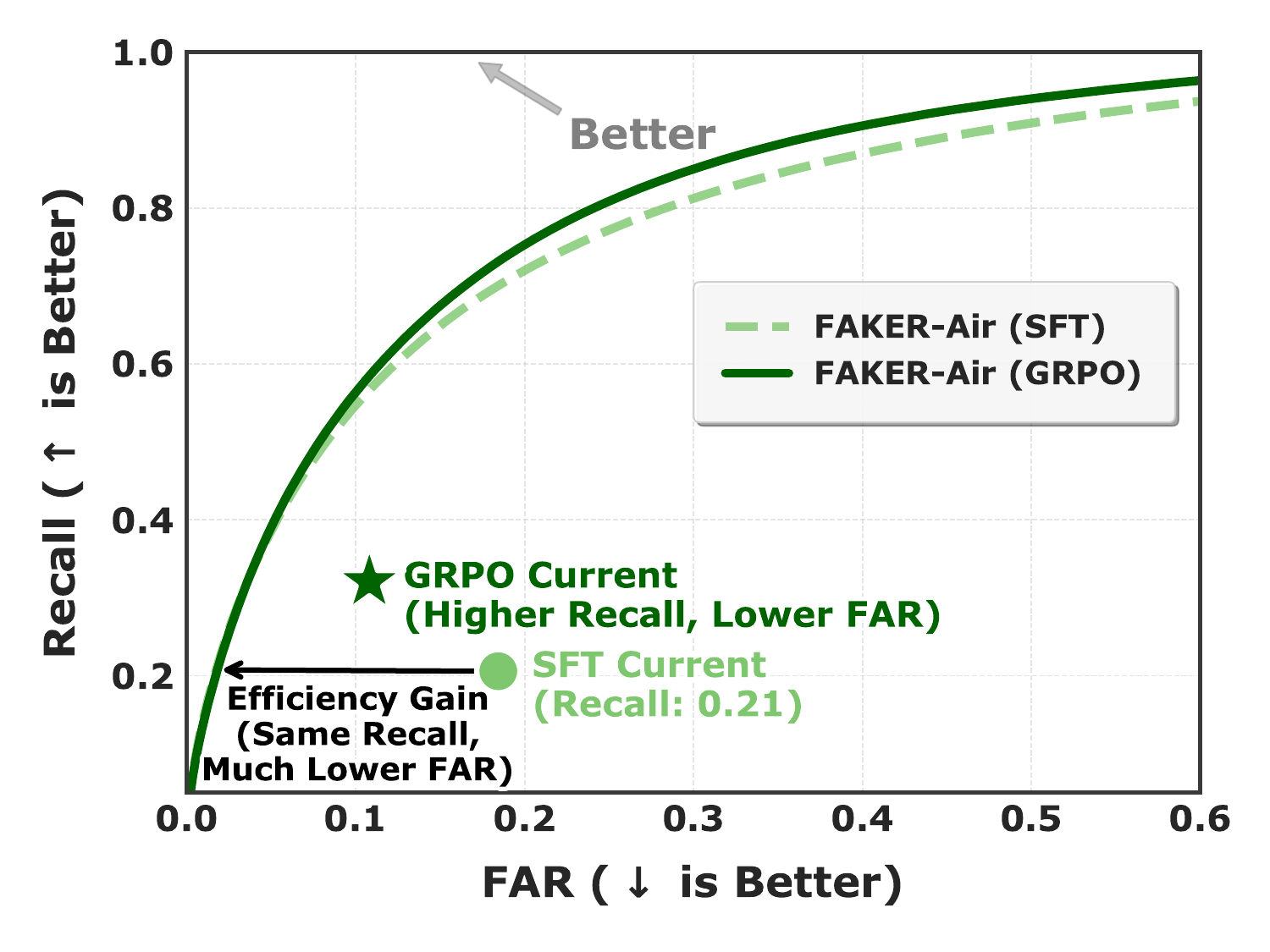}
        \caption{PM$_{10}$}
    \end{subfigure}
    \vspace{-1em}
    \caption{\textbf{\texttt{Bad+V.Bad} FAR--Recall sweep.} GRPO dominates SFT across thresholds, supporting decision-reliable deployment.}
    \label{fig:pareto}
\end{figure*}

\begin{figure*}[t]
    \centering
    \begin{subfigure}[t]{0.49\linewidth}
        \centering
        \includegraphics[width=\linewidth]{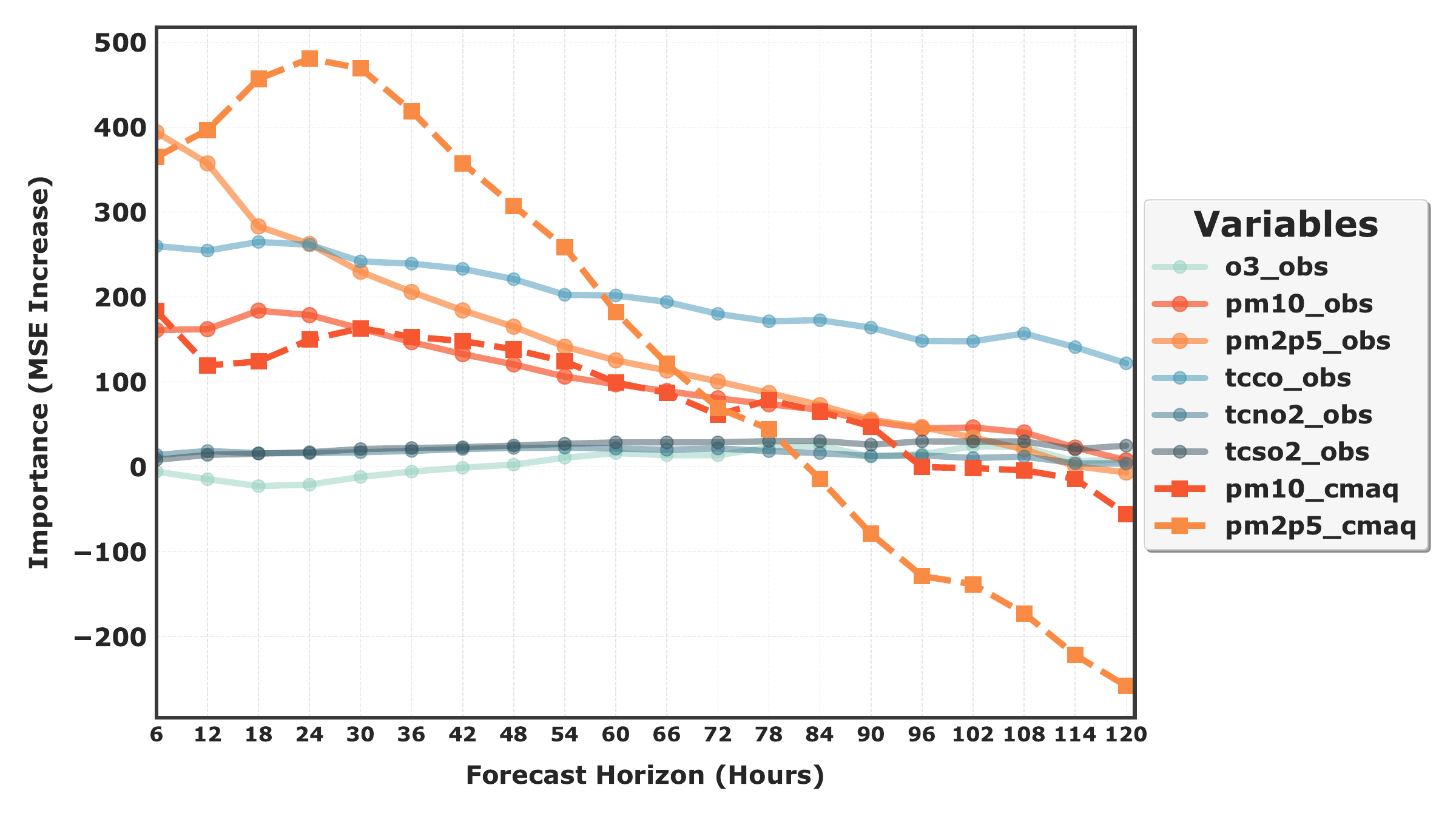}
        
        \caption{FAKER-Air$_\textsuperscript{SFT}$}
    \end{subfigure}
    \begin{subfigure}[t]{0.49\linewidth}
        \centering
        \includegraphics[width=\linewidth]{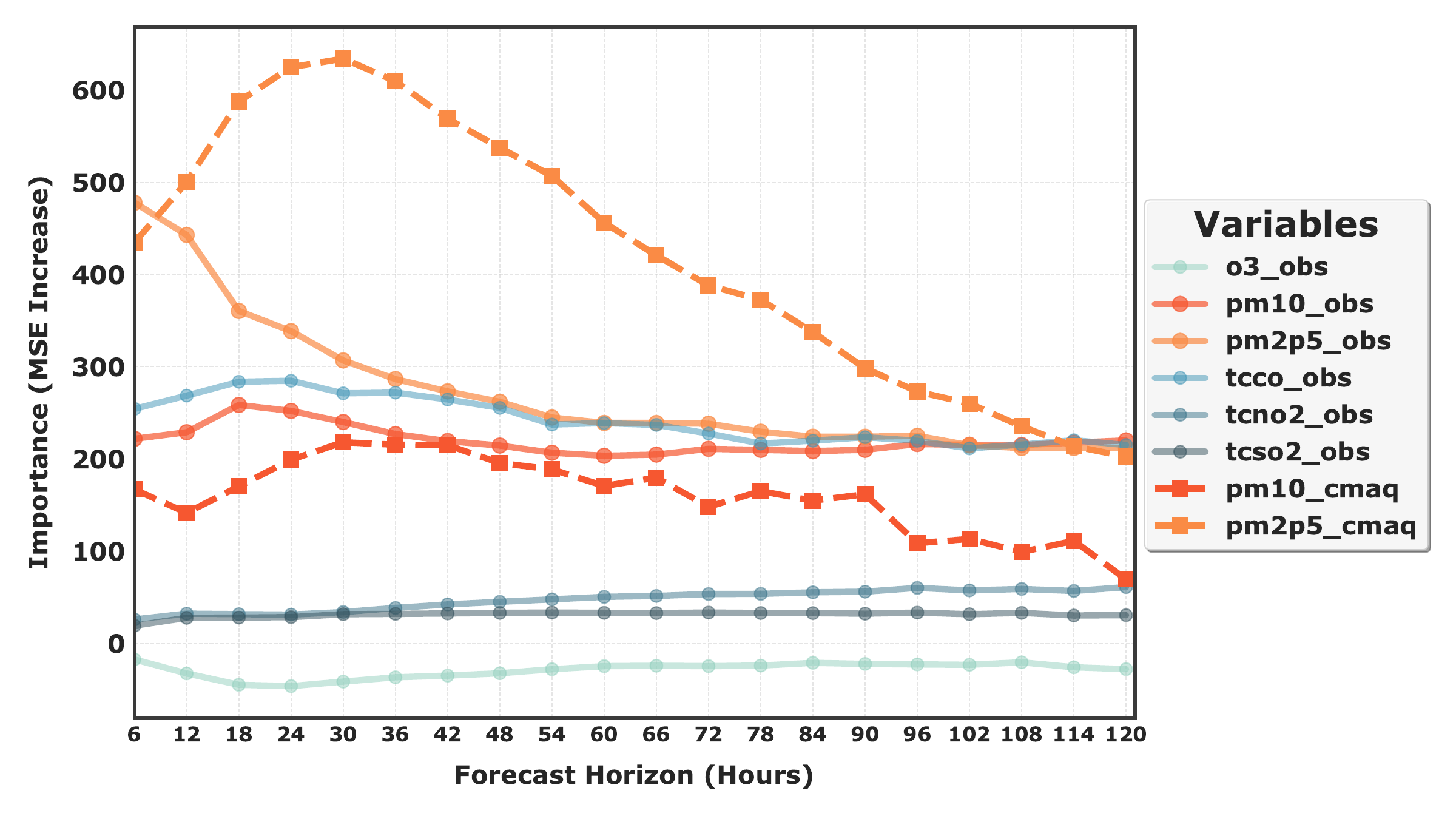}
        
        \caption{FAKER-Air$_\textsuperscript{GRPO}$}
    \end{subfigure}
    \vspace{-1em}
    \caption{\textbf{Rollout permutation feature importance (PFI).}}
    \label{fig:feature}
\end{figure*}

To further validate the stability of our GRPO framework, we conducted extensive sensitivity analyses on key hyperparameters ($\sigma$, KL penalty, Group size $G$) and evaluated alternative reward structures (Table~\ref{tab:grpo_sens}).

\textbf{Reward Realism.} Since operational alerts are discrete, we utilize a verifiable, low-variance binary signal. We rigorously evaluated an alternative ``Hierarchical Reward'' (+0.5 for grouping, +1.0 for exact match) to address misclassification distance. However, as shown in Table~\ref{tab:grpo_sens}, this degraded FAR and Bias, confirming that the sharper binary signal yields more reliable optimization for operational deployment.

\textbf{Stability and Sensitivity.} Table~\ref{tab:grpo_sens} confirms that GRPO consistently outperforms the SFT baseline across diverse hyperparameter settings. Regardless of specific choices for sampling scale ($\sigma$), KL divergence penalty, or group size ($G$), GRPO maintains a significantly lower FAR ($\approx$17-19\%) and perfectly calibrated Bias ($\approx$1.0) compared to SFT (FAR: 32.86\%, Bias: 1.52). This demonstrates that the performance gain stems from the fundamental group-alignment mechanism itself, rather than hyperparameter tuning.

%%%%%%%%%%%%%%%%%%%%%%%%%%%%%%%%%%%%%%%%%%%%%%%%%%%%%%%%%%%%%%%%%%%%%%%%%%%%%%%%%

\section{Operational Reliability and Decision Boundaries}
\label{sec:decision_boundary}

A critical concern in operational forecasting is whether optimizing for overall utility (via GRPO) inadvertently sacrifices the ability to detect severe, albeit rare, pollution events (\texttt{VeryBad} class). The perceived drop in \texttt{VeryBad}-class CSI at a fixed threshold reflects a choice of operating point, not a loss of fundamental severe-event detection capability.

To demonstrate this, Figure~\ref{fig:pareto} presents a threshold sweep on 2023 severe-event detection (\texttt{Bad} + \texttt{VeryBad} classes). The results clearly show that FAKER-Air\textsubscript{GRPO} Pareto-dominates the SFT baseline across the entire threshold spectrum for both PM$_{2.5}$ and PM$_{10}$. 

At our default PM$_{2.5}$ operating point, GRPO achieves higher Recall (0.34 vs. 0.23) while maintaining roughly half the False Alarm Rate (0.17 vs. 0.33) compared to equivalent SFT operating points. This indicates that GRPO fundamentally improves the underlying decision boundary, enabling high sensitivity to severe events with significantly fewer false alarms, thereby supporting decision-reliable deployment.

\section{Mechanistic Explainability via Feature Importance}
\label{sec:explainability}

To understand how GRPO alters the model's decision-making process, we conducted Permutation Feature Importance (PFI) analysis on the rollout trajectories (Figure~\ref{fig:feature}). This analysis provides a mechanistic decomposition of model decisions and reveals how GRPO maintains physical validity over long horizons.

First, physical validity is confirmed at short lead times: both SFT and GRPO models prioritize correct physical drivers, ensuring predictions are not artifact-driven. However, a critical divergence emerges at longer horizons. While the SFT model gradually loses reliance on valid physical signals (exhibiting temporal drift), GRPO sustains its reliance on physics priors and actively leverages chemical tracers (e.g., NO$_2$) even at +96h and +120h. 

This proves that GRPO achieves long-horizon reliability not by memorizing statistical shortcuts, but by enforcing physical consistency, thereby preventing the trajectory decay typical of standard autoregressive regression.

%%%%%%%%%%%%%%%%%%%%%%%%%%%%%%%%%%%%%%%%%%%%%%%%%%%%%%%%%%%%%%%%%%%%%%%%%%%%%%%%%%%%

\section{Implementation Details}
\label{sec:implementation}

We provide comprehensive implementation details to ensure full reproducibility of FAKER-Air. Our framework consists of two sequential training stages: Supervised Fine-Tuning (SFT) followed by Group-Relative Policy Optimization (GRPO). All experiments are conducted on two NVIDIA H200 GPUs using PyTorch with mixed-precision training (FP16).

\subsection{Supervised Fine-Tuning Stage}

\paragraph{Model Architecture.} 
We build upon Aurora 0.25°~\citep{bodnar2025foundation}, a 1.3B-parameter vision transformer architecture. The model employs a hierarchical encoder-decoder architecture with self-attention and cross-attention mechanisms. Train all weights from scratch using only our regional observational and reanalysis data. We use the Aurora architecture but train from random initialization.

\paragraph{Training Data.} 
We integrate two complementary data sources covering East Asia (13.0°N--52.0°N, 97.0°E--170.0°E) at 27km resolution (174$\times$128 spatial grids). We set the training period as 2016--2022, and test in 2023. Details are as below:
\begin{itemize}
    \item \textbf{OBS}: 532 AIRKOREA ground monitoring stations with hourly PM measurements (see Section~\ref{sec:sup_OBS}).
    \item \textbf{CMAQ}: Regional chemical transport model outputs providing physics-informed PM\textsubscript{2.5} and PM\textsubscript{10} fields with explicit aerosol chemistry (see Section~\ref{sec:sub_CMQA}).
\end{itemize}

\paragraph{Hyperparameters.} 
We train for 30 epochs using distributed data-parallel training across 2 GPUs:
\begin{itemize}
    \item Batch size: 8 per GPU
    \item Optimizer: AdamW ($\beta_1=0.9$, $\beta_2=0.999$, $\epsilon=10^{-8}$)~\citep{loshchilov2017decoupled}
    \item Learning rate: $1 \times 10^{-4}$ with cosine annealing
    \item Gradient clipping: Maximum norm of 1.0
    \item Rollout horizon: T = 4
\end{itemize}

\paragraph{Loss Function.} 
We employ a weighted multi-task loss combining reconstruction and classification objectives:
\begin{equation}
\mathcal{L}_{\mathrm{SFT}} = \lambda_{\mathrm{CMAQ}} \mathcal{L}_{\mathrm{recon}} + \lambda_{\mathrm{PM}} \mathcal{L}_{\mathrm{cls}}
\end{equation}
where $\lambda_{\mathrm{CMAQ}}=0.5$ balances CMAQ reconstruction and $\lambda_{\mathrm{PM}}=2.0$ emphasizes PM classification accuracy.

\paragraph{Regularization.} 
We apply spatial masking to random grid cells during training to prevent trivial identity mapping from input observations. This forces the model to learn spatiotemporal interpolation patterns.

\subsection{Group-Relative Policy Optimization Stage}

\paragraph{Initialization.} 
Both policy and reference networks are initialized from the best SFT checkpoint selected by validation PM\textsubscript{2.5} F1-score. The reference network remains frozen to provide KL regularization~\citep{shao2024deepseekmath}.

\paragraph{Hyperparameters.} 
We fine-tune for 4 epochs with the following configuration:
\begin{itemize}
    \item Batch size: 1 per GPU
    \item Group size: 4 rollouts per state
    \item Optimizer: AdamW ($\beta_1=0.9$, $\beta_2=0.999$)
    \item Learning rate: $5 \times 10^{-8}$
    \item Rollout curriculum: $T_{\text{start}}=1 \rightarrow T_{\text{max}}=4$
\end{itemize}

\paragraph{Policy Sampling.} 
We implement Gaussian policy with antithetic variance reduction:
\begin{equation}
a_t = \mu_{\theta}(s_t) + \sigma_t \cdot \epsilon_t, \quad \epsilon_t \sim \mathcal{N}(0, I)
\end{equation}
where $\sigma_t = \sigma_0 \cdot 0.995^t$ with $\sigma_{\text{PM2.5}}=4.0$ and $\sigma_{\text{PM10}}=5.5$. Antithetic pairs are generated for odd-indexed groups ($\epsilon_t \leftarrow -\epsilon_t$) to reduce gradient variance.

\paragraph{Reward Computation.} 
Action log-probabilities are computed only for PM\textsubscript{2.5} and PM\textsubscript{10} variables with per-variable weights $w_{\text{PM2.5}}=1.0$ and $w_{\text{PM10}}=0.5$. Invalid regions, without observation stations, are masked during loss calculation.

\subsection{Computational Cost}

The SFT stage requires approximately 30 hours (30 epochs on 2$\times$H200 GPUs). The GRPO stage requires approximately 8 hours (4 epochs) due to multiple rollout samples per batch (4$\times$ forward passes for group sampling).

%%%%%%%%%%%%%%%%%%%%%%%%%%%%%%%%%%%%%%%%%%%%%%%%%%%%%%%%%%%%%%%%%%%%%%%%%%%%%%%%%
\section{Limitations and Future Work}
\label{sec:limitations}

While FAKER-Air demonstrates substantial improvements over foundation model baselines and achieves operational reliability through GRPO-based policy optimization, several limitations remain that highlight directions for future research.

\subsection{Cross-Pollutant and Multi-Task Learning}

Our current framework treats PM\textsubscript{2.5} and PM\textsubscript{10} as independent prediction tasks, missing potential synergies from joint learning. Atmospheric chemistry couples these pollutants through shared precursors (SO\textsubscript{2}, NO\textsubscript{x}) and transformation pathways, suggesting that multi-task architectures could improve efficiency and generalization. Additionally, operational air quality management requires concurrent forecasts for multiple species (O\textsubscript{3}, CO, SO\textsubscript{2}), motivating comprehensive multi-pollutant frameworks. Future extensions might investigate shared encoder architectures with pollutant-specific decoder heads, potentially regularized through cross-pollutant consistency constraints (e.g., penalizing PM\textsubscript{2.5}/PM\textsubscript{10} ratio violations).

\subsection{Cross-Pollutant and Multi-Task Learning}
Currently, PM$_{2.5}$ and PM$_{10}$ share an encoder, but we forecast all variables without explicitly coupling their chemistry or ratio constraints. As suggested during the review process, enforcing explicit chemical constraints (e.g., ensuring PM$_{2.5}$ is strictly a fraction of PM$_{10}$ under certain physical limits) could further refine consistency. We agree that incorporating such explicit multi-pollutant synergies is a promising avenue for future work.

\subsection{Future Works}

Beyond these limitations, several promising directions remain for future work. First, integrating uncertainty quantification into both the SFT and GRPO stages could improve operational robustness by enabling confidence-aware decision support. Second, extending the framework to additional pollutants and meteorological variables may further enhance regional forecasting capability and support holistic air-quality management. Third, incorporating cross-regional transfer learning or domain adaptation techniques would help assess the model’s generalizability beyond East Asia. Finally, exploring more expressive reward functions and alternative alignment methods may yield additional gains in reliability, particularly under extreme pollution events.

%% file: tables/aqi_info.tex
\begin{table}[t]
  \centering
  \caption{\textbf{Binary Class and Air Quality Index (AQI) Classification Standards for PM\textsubscript{2.5} and PM\textsubscript{10}}}
  \label{tab:aqi_classification}
  \small
  \renewcommand{\arraystretch}{1.3}
  \begin{tabular}{c|c|c|c}
    \toprule
    \textbf{Binary} & \textbf{AQI} & \textbf{PM\textsubscript{10}} \scriptsize{($\mu$g/m$^3$)} & \textbf{PM\textsubscript{2.5}} \scriptsize{($\mu$g/m$^3$)} \\
    \midrule
    \multirow{2}{*}{\textbf{Clean}} & \cellcolor{RoyalBlue!20}
    \texttt{Good} & 0--30 & 0--15 \\
    \cline{2-2}\cline{3-4}
      & \cellcolor{ForestGreen!20}
    \texttt{Moderate} & 31--80 & 16--35 \\
    \midrule
     \multirow{2}{*}{\textbf{Polluted}} & \cellcolor{yellow!20}
    \texttt{Bad} & 81--150 & 36--75 \\
    \cline{2-2}\cline{3-4}
      & \cellcolor{red!20}
    \texttt{VeryBad} & 151-- & 76-- \\
    \bottomrule
  \end{tabular}
\end{table}

%% file: tables/obs_stats.tex
\begin{table}[h]
  \centering
  \caption{\textbf{Statistics of Observation Stations and Meteorological Monitoring Sites}}
  \label{tab:station_statistics}
  \normalsize
  \renewcommand{\arraystretch}{1.4}
  
  \begin{tabular}{lc}
    \toprule
    \textbf{Data Source} & \textbf{\# of Stations} \\
    \midrule
    Korean Air Quality (AIRKOREA) & 532 \\
    Chinese Air Quality (PM25.in) & 1,781 \\
    Chinese Air Quality (AIRQUALITY) & 1,290 \\
    \midrule
    Korean Meteorology (ASOS) & 94 \\
    Korean Meteorology (AWS) & 349 \\
    Chinese Meteorology (Wyoming) & 40 \\
    \bottomrule
  \end{tabular}
\end{table}

%% file: tables/cmaq_vs_cams.tex
\begin{table}[t]
\centering
\caption{Comparison of seasonal and overall average L1 distance for CMAQ and CAMS reanalysis models (unit: $\mu$g/m$^3$).}
\label{tab:cmaq_vs_cams_seasonal}
\resizebox{\columnwidth}{!}{%
\begin{tabular}{l|cccc|c}
\toprule
\multicolumn{6}{c}{\textbf{PM\textsubscript{2.5}}} \\
\midrule
\textbf{Model} & \textbf{Spring} & \textbf{Summer} & \textbf{Fall} & \textbf{Winter} & \textbf{Avg} \\
\midrule
CAMS & 56.98 & 43.04 & 50.71 & 59.61 & 52.66 \\
\rowcolor{cyan!5}
\textbf{CMAQ} & \textbf{20.46} & \textbf{17.53} & \textbf{18.09} & \textbf{29.11} & \textbf{21.33}  \\ [-1pt]
\rowcolor{cyan!5}
    &    \scriptsize{\textcolor{RoyalBlue}{\texttt{(178\%)}}} & \scriptsize{\textcolor{RoyalBlue}{\texttt{(146\%)}}} & 
    \scriptsize{\textcolor{RoyalBlue}{\texttt{(180\%)}}} & \scriptsize{\textcolor{RoyalBlue}{\texttt{(105\%)}}} & 
    \scriptsize{\textcolor{RoyalBlue}{\texttt{(147\%)}}}\\
\midrule
\midrule
\multicolumn{6}{c}{\textbf{PM\textsubscript{10}}} \\
\midrule
\textbf{Model} & \textbf{Spring} & \textbf{Summer} & \textbf{Fall} & \textbf{Winter} & \textbf{Avg} \\
\midrule
CAMS & 74.01 & 56.65 & 64.82 & 80.02 & 68.99 \\
\rowcolor{cyan!5}
\textbf{CMAQ} & \textbf{47.34} & \textbf{31.79} & \textbf{30.12} & \textbf{44.51} & \textbf{38.60} \\ [-1pt]
\rowcolor{cyan!5}
    &    \scriptsize{\textcolor{RoyalBlue}{\texttt{(56\%)}}} & \scriptsize{\textcolor{RoyalBlue}{\texttt{(78\%)}}} & 
    \scriptsize{\textcolor{RoyalBlue}{\texttt{(115\%)}}} & \scriptsize{\textcolor{RoyalBlue}{\texttt{(80\%)}}} & 
    \scriptsize{\textcolor{RoyalBlue}{\texttt{(79\%)}}}\\
\bottomrule
\end{tabular}%
}
\end{table}

%% file: tables/latlon.tex
\begin{table*}[tbh]
  \centering
  \caption{\textbf{Spatial Domain Coverage of CMAQ and OBS Datasets.} The CMAQ 27 km East Asia domain covers the entire region from southern China to northern Korea, spanning approximately 39° latitude (4,300 km) and 73° longitude (5,800 km). OBS station measurements are spatially interpolated onto the CMAQ grid structure for model training and evaluation.}
  \label{tab:domain_coverage}
  \small
  \renewcommand{\arraystretch}{1.3}
  
  \begin{tabular}{lcccc}
    \toprule
    \textbf{Dataset} & \textbf{Resolution} & \textbf{Grid Size} & \textbf{Latitude Range} & \textbf{Longitude Range} \\
    \midrule
    \multicolumn{5}{c}{\textit{CMAQ Reanalysis Domain}} \\
    \midrule
    CMAQ (East Asia) & 27 km & 174 $\times$ 128 & 13.0°N -- 52.0°N & 97.0°E -- 170.0°E \\
    \midrule
    \multicolumn{5}{c}{\textit{OBS Ground Monitoring Stations}} \\
    \midrule
    Korea (AIRKOREA) & Point-based & 532 stations & \multirow{3}{*}{Interpolated to CMAQ grid} & \multirow{3}{*}{Interpolated to CMAQ grid} \\
    China (PM25.in) & Point-based & 1,781 stations & & \\
    China (AIRQUALITY) & Point-based & 1,290 stations & & \\
    \bottomrule
  \end{tabular}
\end{table*}

%% file: tables/stat_2016.tex
\begin{table*}[t]
  \centering
  \caption{\textbf{Monthly 4-Class AQI and Binary Classification Distribution (\%) for PM\textsubscript{2.5} and PM\textsubscript{10} on OBS Dataset (Year 2016).} 
  Both pollutants exhibit strong seasonal patterns with class imbalance: winter months show elevated Polluted class (67--72\% for PM\textsubscript{2.5}), while summer months demonstrate Clean class dominance (61--68\% for PM\textsubscript{2.5}), reflecting East Asia's characteristic seasonal pollution dynamics.}
  \label{tab:2016_monthly_class_distribution}
  \tiny
  \renewcommand{\arraystretch}{0.8}
  
  \resizebox{\textwidth}{!}{
  \begin{tabular}{l|cccccccccccc|c|c}
    \toprule
    \textbf{AQI} & \textbf{Jan} & \textbf{Feb} & \textbf{Mar} & \textbf{Apr} & \textbf{May} & \textbf{Jun} & \textbf{Jul} & \textbf{Aug} & \textbf{Sep} & \textbf{Oct} & \textbf{Nov} & \textbf{Dec} & \multicolumn{2}{c}{\textbf{Annual}} \\
    \midrule
    \multicolumn{15}{c}{\textbf{PM\textsubscript{2.5}} (Thresholds: 15, 35, 75 $\mu$g/m$^3$)} \\
    \midrule
    \texttt{Good} & 7.6 & 8.8 & 6.5 & 7.8 & 10.7 & 16.6 & 19.9 & 20.6 & 15.0 & 15.7 & 8.6 & 6.2 & 12.0 & \multirow{2}{*}{47.0} \\
    \texttt{Mod.} & 25.4 & 30.7 & 24.6 & 35.9 & 39.4 & 44.2 & 43.0 & 47.2 & 38.8 & 41.9 & 27.6 & 22.1 & 35.0 & \\
    \arrayrulecolor{gray}
    \cmidrule{1-15}
    \arrayrulecolor{black}
   \texttt{Bad} & 31.3 & 35.6 & 40.2 & 39.7 & 39.9 & 33.2 & 30.9 & 28.6 & 35.6 & 32.5 & 36.8 & 33.2 & 34.8 & \multirow{2}{*}{53.0} \\
    \texttt{V.Bad} & 35.8 & 24.9 & 28.7 & 16.6 & 10.0 & 6.0 & 6.2 & 3.6 & 10.7 & 9.9 & 27.1 & 38.5 & 18.3 & \\
    
    \midrule
    \midrule
    
    \multicolumn{15}{c}{\textbf{PM\textsubscript{10}} (Thresholds: 30, 80, 150 $\mu$g/m$^3$)} \\
    \midrule
    \texttt{Good} & 10.3 & 10.5 & 6.4 & 8.2 & 10.8 & 19.1 & 22.0 & 20.9 & 16.9 & 17.6 & 9.8 & 7.7 & 13.3 & \multirow{2}{*}{60.2} \\
    \texttt{Mod.} & 37.9 & 44.3 & 36.3 & 42.1 & 49.5 & 55.3 & 55.0 & 59.8 & 52.1 & 54.5 & 41.1 & 36.6 & 46.9 & \\
    \arrayrulecolor{gray}
    \cmidrule{1-15}
    \arrayrulecolor{black}
   \texttt{Bad} & 27.6 & 28.3 & 33.5 & 30.8 & 30.3 & 21.5 & 19.2 & 16.9 & 23.7 & 20.7 & 29.0 & 28.8 & 25.9 & \multirow{2}{*}{39.8} \\
    \texttt{V.Bad} & 24.3 & 16.9 & 23.9 & 19.0 & 9.4 & 4.1 & 3.9 & 2.4 & 7.3 & 7.2 & 20.2 & 26.9 & 13.9 & \\
    \bottomrule
  \end{tabular}}
\end{table*}

%% file: tables/stat_2023.tex
\begin{table*}[t]
  \centering
  \caption{\textbf{Monthly 4-Class AQI and Binary Classification Distribution (\%) for PM\textsubscript{2.5} and PM\textsubscript{10} on OBS Dataset (Year 2023).} 
  Compared to 2016, Year 2023 exhibits substantially improved air quality: Clean class dominance increased to 69.6\% (vs 47.0\%) for PM\textsubscript{2.5} and 77.7\% (vs 60.2\%) for PM\textsubscript{10}, while maintaining similar seasonal patterns with winter pollution episodes and summer clean-air periods.}
  \label{tab:2023_monthly_class_distribution}
  \tiny
  \renewcommand{\arraystretch}{0.8}
  
  \resizebox{\textwidth}{!}{
  \begin{tabular}{l|cccccccccccc|c|c}
    \toprule
    \textbf{AQI} & \textbf{Jan} & \textbf{Feb} & \textbf{Mar} & \textbf{Apr} & \textbf{May} & \textbf{Jun} & \textbf{Jul} & \textbf{Aug} & \textbf{Sep} & \textbf{Oct} & \textbf{Nov} & \textbf{Dec} & \multicolumn{2}{c}{\textbf{Annual}} \\
    \midrule
    \multicolumn{15}{c}{\textbf{PM\textsubscript{2.5}} (Thresholds: 15, 35, 75 $\mu$g/m$^3$)} \\
    \midrule
    \texttt{Good} & 16.1 & 14.7 & 14.3 & 19.2 & 24.8 & 38.5 & 47.9 & 43.0 & 42.1 & 29.7 & 25.6 & 17.4 & 27.7 & \multirow{2}{*}{69.5} \\
    \texttt{Mod.} & 30.9 & 34.0 & 39.2 & 46.0 & 52.0 & 50.4 & 44.8 & 48.0 & 43.3 & 41.2 & 38.8 & 33.5 & 41.9 & \\
    \arrayrulecolor{gray}
    \cmidrule{1-15}
    \arrayrulecolor{black}
   \texttt{Bad} & 32.4 & 34.6 & 36.3 & 28.4 & 21.4 & 10.7 & 7.1 & 8.8 & 13.6 & 23.0 & 27.8 & 32.4 & 23.1 & \multirow{2}{*}{30.5} \\
    \texttt{V.Bad} & 20.6 & 16.7 & 10.3 & 6.5 & 1.8 & 0.4 & 0.3 & 0.2 & 0.9 & 6.1 & 7.8 & 16.8 & 7.4 & \\
    
    \midrule
    \midrule
    
    \multicolumn{15}{c}{\textbf{PM\textsubscript{10}} (Thresholds: 30, 80, 150 $\mu$g/m$^3$)} \\
    \midrule
    \texttt{Good} & 17.3 & 18.1 & 10.8 & 17.9 & 26.3 & 42.9 & 52.2 & 48.8 & 46.3 & 31.2 & 27.2 & 21.5 & 29.9 & \multirow{2}{*}{77.7} \\
    \texttt{Mod.} & 41.5 & 51.1 & 45.6 & 46.7 & 57.3 & 50.6 & 44.2 & 47.8 & 47.3 & 49.5 & 46.3 & 45.7 & 47.8 & \\
    \arrayrulecolor{gray}
    \cmidrule{1-15}
    \arrayrulecolor{black}
   \texttt{Bad} & 26.7 & 24.2 & 31.3 & 22.4 & 13.3 & 6.1 & 3.3 & 3.2 & 6.1 & 15.8 & 21.1 & 22.8 & 16.4 & \multirow{2}{*}{22.3} \\
    \texttt{V.Bad} & 14.5 & 6.6 & 12.3 & 13.1 & 3.1 & 0.4 & 0.3 & 0.3 & 0.3 & 3.5 & 5.5 & 9.9 & 5.9 & \\
    \bottomrule
  \end{tabular}}
\end{table*}

%% file: tables/test_2016.tex
\begin{table*}[t]
  \centering
  \caption{\textbf{Test on 2016 for PM\textsubscript{2.5} and PM\textsubscript{10} over 120h.}
  For Binary(2-class): Acc, F1, Prec (↑) and FAR (↓); Bias≈1 preferred. 
  For AQI(4-class): Acc, F1-macro, F1-weighted, F1-micro (↑), plus per-class F1.}
  \vspace{-0.5em}
  \label{tab:grpo_ablation}
  \small
  \setlength{\tabcolsep}{3pt}
  \renewcommand{\arraystretch}{1.0}

  \resizebox{\textwidth}{!}{
  \begin{tabular}{
  l !{\vrule width \arrayrulewidth}
  c c c c c
  !{\vrule width \arrayrulewidth}
  c c c c
  !{\vrule width \arrayrulewidth}
  c c c c
}
    \toprule

     \rowcolor{gray!15}
    \multicolumn{14}{c}{\textbf{PM\textsubscript{2.5}}} \\

    \midrule

    &
      \multicolumn{5}{c|}{\textbf{Binary Metrics}} &
      \multicolumn{8}{c}{\textbf{AQI (4-class) Metrics}} \\
    \cmidrule(l){2-6} \cmidrule(l){7-14}

     &
      Acc & F1 & Prec & FAR & Bias &
      Acc & F1-macro & F1-weighted & F1-micro &
      \texttt{Good} & \texttt{Mod.} & \texttt{Bad} & \texttt{V.Bad} \\

    \midrule

    FAKER-Air\textsubscript{SFT}
      & 66.01 & 72.70 & 63.45 & 55.71 & 1.34
      & 45.01 & 38.53 & 43.00 & 45.01
      & 13.49 & 42.39 & 50.44 & 47.81 \\

    FAKER-Air\textsubscript{GRPO}
      & 68.15 & 70.60 & 69.35 & 36.11 & 1.04
      & 43.10 & 41.40 & 43.26 & 43.10
      & 29.66 & 42.97 & 45.73 & 47.23 \\

      \midrule
      \midrule

      \rowcolor{gray!15}
      \multicolumn{14}{c}{\textbf{PM\textsubscript{10}}} \\

      \midrule

      &
      \multicolumn{5}{c|}{\textbf{Binary Metrics}} &
      \multicolumn{8}{c}{\textbf{AQI (4-class) Metrics}} \\
    \cmidrule(l){2-6} \cmidrule(l){7-14}

     &
      Acc & F1 & Prec & FAR & Bias &
      Acc & F1-macro & F1-weighted & F1-micro &
      \texttt{Good} & \texttt{Mod.} & \texttt{Bad} & \texttt{V.Bad} \\

    \midrule

      FAKER-Air\textsubscript{SFT}
      & 71.30 & 67.81 & 62.08 & 31.00 & 1.20
      & 49.49 & 41.45 & 48.09 & 49.49
      & 19.33 & 60.03 & 45.17 & 41.28 \\

    FAKER-Air\textsubscript{GRPO}
      & 73.12 & 65.43 & 68.23 & 19.89 & 0.92
      & 47.22 & 42.84 & 47.16 & 47.22
      & 33.89 & 56.65 & 41.56 & 39.25 \\

    \bottomrule
  \end{tabular}}
  \label{tab:test_2016}
  \vspace{5em}
\end{table*}

%% file: tables/monthly_metric.tex
\begin{table*}[t!]
  \centering
  \caption{\textbf{Monthly Binary F1-Score for PM\textsubscript{2.5} and PM\textsubscript{10} Forecasting (Seasonal Order).} 
  FAKER-Air\textsubscript{GRPO} maintains competitive F1-scores while significantly reducing False Alarm Rate compared to FAKER-Air\textsubscript{SFT}, demonstrating improved precision-recall balance across all seasons. Values in parentheses indicate absolute improvement over Aurora baseline.}
  \label{tab:monthly_binary_f1}
  \renewcommand{\arraystretch}{1.1}
  
  \resizebox{\textwidth}{!}{
  \begin{tabular}{l|ccc|ccc|ccc|ccc}
    \toprule
    & \multicolumn{3}{c|}{\textbf{Spring (Mar--May)}} & \multicolumn{3}{c|}{\textbf{Summer (Jun--Aug)}} & \multicolumn{3}{c|}{\textbf{Fall (Sep--Nov)}} & \multicolumn{3}{c}{\textbf{Winter (Dec--Feb)}} \\
    \cmidrule(lr){2-4} \cmidrule(lr){5-7} \cmidrule(lr){8-10} \cmidrule(lr){11-13}
    \textbf{Method} & Mar & Apr & May & Jun & Jul & Aug & Sep & Oct & Nov & Dec & Jan & Feb \\
    \midrule
    \multicolumn{13}{c}{\textbf{PM\textsubscript{2.5}} (Threshold: 35 $\mu$g/m$^3$)} \\
    \midrule
    Aurora & 16.7 & 7.6 & 6.4 & 11.0 & 6.8 & 12.2 & 11.8 & 14.9 & 12.9 & 19.3 & 25.7 & 18.6 \\
    \midrule
    FAKER-Air\textsubscript{SFT} & 64.7 & 57.1 & 44.0 & 29.0 & 17.2 & 28.8 & 43.1 & 58.8 & 60.6 & 71.7 & 73.5 & 69.8 \\
    \rowcolor{cyan!5}
    FAKER-Air\textsubscript{GRPO} & 57.9 & 54.2 & 42.1 & 20.9 & 10.2 & 21.8 & 41.1 & 52.1 & 53.1 & 70.7 & 69.9 & 64.1 \\ [-2pt]
    \rowcolor{cyan!5}
    & \scriptsize{\textcolor{RoyalBlue}{\texttt{(+41.2)}}} & \scriptsize{\textcolor{RoyalBlue}{\texttt{(+46.6)}}} & \scriptsize{\textcolor{RoyalBlue}{\texttt{(+35.7)}}} & \scriptsize{\textcolor{RoyalBlue}{\texttt{(+9.9)}}} & \scriptsize{\textcolor{RoyalBlue}{\texttt{(+3.4)}}} & \scriptsize{\textcolor{RoyalBlue}{\texttt{(+9.6)}}} & \scriptsize{\textcolor{RoyalBlue}{\texttt{(+29.3)}}} & \scriptsize{\textcolor{RoyalBlue}{\texttt{(+37.2)}}} & \scriptsize{\textcolor{RoyalBlue}{\texttt{(+40.2)}}} & \scriptsize{\textcolor{RoyalBlue}{\texttt{(+51.4)}}} & \scriptsize{\textcolor{RoyalBlue}{\texttt{(+44.2)}}} & \scriptsize{\textcolor{RoyalBlue}{\texttt{(+45.5)}}} \\
    \midrule
    \multicolumn{13}{c}{\textbf{PM\textsubscript{10}} (Threshold: 80 $\mu$g/m$^3$)} \\
    \midrule
    Aurora & 3.6 & 0.9 & 0.8 & 2.0 & 1.6 & 3.0 & 3.8 & 4.4 & 2.7 & 7.9 & 9.7 & 5.8 \\
    \midrule
    FAKER-Air\textsubscript{SFT} & 65.7 & 61.5 & 41.6 & 29.9 & 20.6 & 22.4 & 35.9 & 55.9 & 54.7 & 64.6 & 68.2 & 59.6 \\
    \rowcolor{cyan!5}
    FAKER-Air\textsubscript{GRPO} & 58.6 & 58.8 & 39.5 & 24.6 & 18.5 & 22.1 & 32.3 & 45.5 & 45.5 & 61.0 & 62.6 & 53.4 \\ [-2pt]
    \rowcolor{cyan!5}
    & \scriptsize{\textcolor{RoyalBlue}{\texttt{(+55.0)}}} & \scriptsize{\textcolor{RoyalBlue}{\texttt{(+57.9)}}} & \scriptsize{\textcolor{RoyalBlue}{\texttt{(+38.7)}}} & \scriptsize{\textcolor{RoyalBlue}{\texttt{(+22.6)}}} & \scriptsize{\textcolor{RoyalBlue}{\texttt{(+16.9)}}} & \scriptsize{\textcolor{RoyalBlue}{\texttt{(+19.1)}}} & \scriptsize{\textcolor{RoyalBlue}{\texttt{(+28.5)}}} & \scriptsize{\textcolor{RoyalBlue}{\texttt{(+41.1)}}} & \scriptsize{\textcolor{RoyalBlue}{\texttt{(+42.8)}}} & \scriptsize{\textcolor{RoyalBlue}{\texttt{(+53.1)}}} & \scriptsize{\textcolor{RoyalBlue}{\texttt{(+52.9)}}} & \scriptsize{\textcolor{RoyalBlue}{\texttt{(+47.6)}}} \\
    \bottomrule
  \end{tabular}}
  \vspace{-0.5em}
\end{table*}

\begin{table*}[t!]
  \centering
  \caption{\textbf{Monthly Binary Accuracy for PM\textsubscript{2.5} and PM\textsubscript{10} Forecasting (Seasonal Order).} 
  FAKER-Air\textsubscript{GRPO} achieves consistently higher accuracy than both Aurora and FAKER-Air\textsubscript{SFT} across all seasons, particularly excelling during summer (Jun--Aug) when clean-air conditions dominate. Values in parentheses indicate absolute improvement over Aurora baseline.}
  \label{tab:monthly_binary_accuracy}
  \renewcommand{\arraystretch}{1.1}
  
  \resizebox{\textwidth}{!}{
  \begin{tabular}{l|ccc|ccc|ccc|ccc}
    \toprule
    & \multicolumn{3}{c|}{\textbf{Spring (Mar--May)}} & \multicolumn{3}{c|}{\textbf{Summer (Jun--Aug)}} & \multicolumn{3}{c|}{\textbf{Fall (Sep--Nov)}} & \multicolumn{3}{c}{\textbf{Winter (Dec--Feb)}} \\
    \cmidrule(lr){2-4} \cmidrule(lr){5-7} \cmidrule(lr){8-10} \cmidrule(lr){11-13}
    \textbf{Method} & Mar & Apr & May & Jun & Jul & Aug & Sep & Oct & Nov & Dec & Jan & Feb \\
    \midrule
    \multicolumn{13}{c}{\textbf{PM\textsubscript{2.5}} (Threshold: 35 $\mu$g/m$^3$)} \\
    \midrule
    Aurora & 53.4 & 63.7 & 75.3 & 87.2 & 92.1 & 89.6 & 83.5 & 68.9 & 62.2 & 52.7 & 49.8 & 49.5 \\
    \midrule
    FAKER-Air\textsubscript{SFT} & 64.8 & 60.4 & 57.9 & 74.6 & 88.4 & 82.6 & 73.7 & 70.4 & 67.1 & 64.8 & 67.4 & 64.9 \\
    \rowcolor{cyan!5}
    FAKER-Air\textsubscript{GRPO} & 65.7 & 66.9 & 69.4 & 83.9 & 91.5 & 87.4 & 82.4 & 75.1 & 69.6 & 69.3 & 68.4 & 65.4 \\ [-2pt]
    \rowcolor{cyan!5}
    & \scriptsize{\textcolor{RoyalBlue}{\texttt{(+12.3)}}} & \scriptsize{\textcolor{RoyalBlue}{\texttt{(+3.2)}}} & \scriptsize{\textcolor{RoyalBlue}{\texttt{(-5.9)}}} & \scriptsize{\textcolor{RoyalBlue}{\texttt{(-3.3)}}} & \scriptsize{\textcolor{RoyalBlue}{\texttt{(-0.6)}}} & \scriptsize{\textcolor{RoyalBlue}{\texttt{(-2.2)}}} & \scriptsize{\textcolor{RoyalBlue}{\texttt{(-1.1)}}} & \scriptsize{\textcolor{RoyalBlue}{\texttt{(+6.2)}}} & \scriptsize{\textcolor{RoyalBlue}{\texttt{(+7.4)}}} & \scriptsize{\textcolor{RoyalBlue}{\texttt{(+16.6)}}} & \scriptsize{\textcolor{RoyalBlue}{\texttt{(+18.6)}}} & \scriptsize{\textcolor{RoyalBlue}{\texttt{(+15.9)}}} \\
    \midrule
    \multicolumn{13}{c}{\textbf{PM\textsubscript{10}} (Threshold: 80 $\mu$g/m$^3$)} \\
    \midrule
    Aurora & 53.8 & 63.0 & 81.9 & 92.5 & 95.9 & 95.8 & 91.8 & 76.9 & 70.3 & 66.6 & 57.7 & 66.5 \\
    \midrule
    FAKER-Air\textsubscript{SFT} & 70.1 & 68.3 & 71.4 & 87.0 & 91.5 & 88.6 & 87.3 & 82.7 & 76.0 & 71.4 & 71.9 & 71.8 \\
    \rowcolor{cyan!5}
    FAKER-Air\textsubscript{GRPO} & 68.9 & 71.0 & 77.4 & 90.8 & 94.1 & 92.2 & 91.1 & 83.3 & 76.5 & 74.4 & 71.7 & 73.4 \\ [-2pt]
    \rowcolor{cyan!5}
    & \scriptsize{\textcolor{RoyalBlue}{\texttt{(+15.1)}}} & \scriptsize{\textcolor{RoyalBlue}{\texttt{(+8.0)}}} & \scriptsize{\textcolor{RoyalBlue}{\texttt{(-4.5)}}} & \scriptsize{\textcolor{RoyalBlue}{\texttt{(-1.7)}}} & \scriptsize{\textcolor{RoyalBlue}{\texttt{(-1.8)}}} & \scriptsize{\textcolor{RoyalBlue}{\texttt{(-3.6)}}} & \scriptsize{\textcolor{RoyalBlue}{\texttt{(-0.7)}}} & \scriptsize{\textcolor{RoyalBlue}{\texttt{(+6.4)}}} & \scriptsize{\textcolor{RoyalBlue}{\texttt{(+6.2)}}} & \scriptsize{\textcolor{RoyalBlue}{\texttt{(+7.8)}}} & \scriptsize{\textcolor{RoyalBlue}{\texttt{(+14.0)}}} & \scriptsize{\textcolor{RoyalBlue}{\texttt{(+6.9)}}} \\
    \bottomrule
  \end{tabular}}
  \vspace{-0.5em}
\end{table*}

\begin{table*}[t!]
  \centering
  \caption{\textbf{Monthly 4-Class F1-Macro Score for PM\textsubscript{2.5} and PM\textsubscript{10} Forecasting (Seasonal Order).} 
  FAKER-Air\textsubscript{GRPO} achieves balanced multi-class discrimination across \texttt{Good}, \texttt{Moderate}, \texttt{Bad}, and \texttt{VeryBad} categories, with substantial improvements over Aurora baseline across all seasons, particularly during winter pollution episodes (Dec--Feb). Values in parentheses indicate absolute improvement over Aurora baseline.}
  \label{tab:monthly_aqi_f1macro}
  \renewcommand{\arraystretch}{1.1}
  
  \resizebox{\textwidth}{!}{
  \begin{tabular}{l|ccc|ccc|ccc|ccc}
    \toprule
    & \multicolumn{3}{c|}{\textbf{Spring (Mar--May)}} & \multicolumn{3}{c|}{\textbf{Summer (Jun--Aug)}} & \multicolumn{3}{c|}{\textbf{Fall (Sep--Nov)}} & \multicolumn{3}{c}{\textbf{Winter (Dec--Feb)}} \\
    \cmidrule(lr){2-4} \cmidrule(lr){5-7} \cmidrule(lr){8-10} \cmidrule(lr){11-13}
    \textbf{Method} & Mar & Apr & May & Jun & Jul & Aug & Sep & Oct & Nov & Dec & Jan & Feb \\
    \midrule
    \multicolumn{13}{c}{\textbf{PM\textsubscript{2.5}} (4-class AQI: \texttt{Good}, \texttt{Moderate}, \texttt{Bad}, \texttt{VeryBad})} \\
    \midrule
    Aurora & 18.7 & 19.0 & 21.2 & 26.2 & 26.6 & 28.5 & 26.5 & 23.3 & 20.5 & 20.0 & 19.9 & 18.0 \\
    \midrule
    FAKER-Air\textsubscript{SFT} & 35.3 & 33.9 & 28.2 & 32.4 & 31.2 & 33.1 & 34.7 & 37.5 & 36.2 & 37.2 & 37.8 & 35.0 \\
    \rowcolor{cyan!5}
    FAKER-Air\textsubscript{GRPO} & 34.6 & 37.5 & 33.8 & 31.7 & 27.4 & 31.2 & 38.1 & 36.8 & 35.9 & 43.8 & 40.5 & 34.6 \\ [-2pt]
    \rowcolor{cyan!5}
    & \scriptsize{\textcolor{RoyalBlue}{\texttt{(+15.9)}}} & \scriptsize{\textcolor{RoyalBlue}{\texttt{(+18.5)}}} & \scriptsize{\textcolor{RoyalBlue}{\texttt{(+12.6)}}} & \scriptsize{\textcolor{RoyalBlue}{\texttt{(+5.5)}}} & \scriptsize{\textcolor{RoyalBlue}{\texttt{(+0.8)}}} & \scriptsize{\textcolor{RoyalBlue}{\texttt{(+2.7)}}} & \scriptsize{\textcolor{RoyalBlue}{\texttt{(+11.6)}}} & \scriptsize{\textcolor{RoyalBlue}{\texttt{(+13.5)}}} & \scriptsize{\textcolor{RoyalBlue}{\texttt{(+15.4)}}} & \scriptsize{\textcolor{RoyalBlue}{\texttt{(+23.8)}}} & \scriptsize{\textcolor{RoyalBlue}{\texttt{(+20.6)}}} & \scriptsize{\textcolor{RoyalBlue}{\texttt{(+16.6)}}} \\
    \midrule
    \multicolumn{13}{c}{\textbf{PM\textsubscript{10}} (4-class AQI: \texttt{Good}, \texttt{Moderate}, \texttt{Bad}, \texttt{VeryBad})} \\
    \midrule
    Aurora & 12.1 & 13.7 & 17.5 & 23.6 & 25.0 & 26.0 & 24.4 & 20.3 & 18.2 & 18.3 & 17.1 & 16.8 \\
    \midrule
    FAKER-Air\textsubscript{SFT} & 38.7 & 37.1 & 32.8 & 36.1 & 33.0 & 34.7 & 39.8 & 43.1 & 40.0 & 41.4 & 40.9 & 37.6 \\
    \rowcolor{cyan!5}
    FAKER-Air\textsubscript{GRPO} & 37.6 & 40.6 & 35.5 & 35.0 & 32.6 & 35.8 & 39.3 & 39.0 & 37.8 & 43.9 & 40.7 & 36.8 \\ [-2pt]
    \rowcolor{cyan!5}
    & \scriptsize{\textcolor{RoyalBlue}{\texttt{(+25.5)}}} & \scriptsize{\textcolor{RoyalBlue}{\texttt{(+26.9)}}} & \scriptsize{\textcolor{RoyalBlue}{\texttt{(+18.0)}}} & \scriptsize{\textcolor{RoyalBlue}{\texttt{(+11.4)}}} & \scriptsize{\textcolor{RoyalBlue}{\texttt{(+7.6)}}} & \scriptsize{\textcolor{RoyalBlue}{\texttt{(+9.8)}}} & \scriptsize{\textcolor{RoyalBlue}{\texttt{(+14.9)}}} & \scriptsize{\textcolor{RoyalBlue}{\texttt{(+18.7)}}} & \scriptsize{\textcolor{RoyalBlue}{\texttt{(+19.6)}}} & \scriptsize{\textcolor{RoyalBlue}{\texttt{(+25.6)}}} & \scriptsize{\textcolor{RoyalBlue}{\texttt{(+23.6)}}} & \scriptsize{\textcolor{RoyalBlue}{\texttt{(+20.0)}}} \\
    \bottomrule
  \end{tabular}}
  \vspace{-0.5em}
\end{table*}

\begin{table*}[t!]
  \centering
  \caption{\textbf{Monthly 4-Class Accuracy for PM\textsubscript{2.5} and PM\textsubscript{10} Forecasting (Seasonal Order).} 
  FAKER-Air\textsubscript{GRPO} demonstrates consistent accuracy improvements across all seasons compared to Aurora, with particularly strong performance during summer clean-air periods (Jun--Aug) where class distributions are more imbalanced. Values in parentheses indicate absolute improvement over Aurora baseline.}
  \label{tab:monthly_aqi_accuracy}
  \renewcommand{\arraystretch}{1.1}
  
  \resizebox{\textwidth}{!}{
  \begin{tabular}{l|ccc|ccc|ccc|ccc}
    \toprule
    & \multicolumn{3}{c|}{\textbf{Spring (Mar--May)}} & \multicolumn{3}{c|}{\textbf{Summer (Jun--Aug)}} & \multicolumn{3}{c|}{\textbf{Fall (Sep--Nov)}} & \multicolumn{3}{c}{\textbf{Winter (Dec--Feb)}} \\
    \cmidrule(lr){2-4} \cmidrule(lr){5-7} \cmidrule(lr){8-10} \cmidrule(lr){11-13}
    \textbf{Method} & Mar & Apr & May & Jun & Jul & Aug & Sep & Oct & Nov & Dec & Jan & Feb \\
    \midrule
    \multicolumn{13}{c}{\textbf{PM\textsubscript{2.5}} (4-class AQI: \texttt{Good}, \texttt{Moderate}, \texttt{Bad}, \texttt{VeryBad})} \\
    \midrule
    Aurora & 24.6 & 28.7 & 34.6 & 43.4 & 48.8 & 48.7 & 45.2 & 34.4 & 29.6 & 25.2 & 23.3 & 22.6 \\
    \midrule
    FAKER-Air\textsubscript{SFT} & 44.0 & 41.3 & 39.8 & 46.4 & 51.3 & 48.4 & 44.3 & 44.5 & 42.9 & 42.3 & 42.5 & 41.4 \\
    \rowcolor{cyan!5}
    FAKER-Air\textsubscript{GRPO} & 40.8 & 42.6 & 44.6 & 49.5 & 51.4 & 49.7 & 52.4 & 45.2 & 41.9 & 44.7 & 41.5 & 38.0 \\ [-2pt]
    \rowcolor{cyan!5}
    & \scriptsize{\textcolor{RoyalBlue}{\texttt{(+16.2)}}} & \scriptsize{\textcolor{RoyalBlue}{\texttt{(+13.9)}}} & \scriptsize{\textcolor{RoyalBlue}{\texttt{(+10.0)}}} & \scriptsize{\textcolor{RoyalBlue}{\texttt{(+6.1)}}} & \scriptsize{\textcolor{RoyalBlue}{\texttt{(+2.6)}}} & \scriptsize{\textcolor{RoyalBlue}{\texttt{(+1.0)}}} & \scriptsize{\textcolor{RoyalBlue}{\texttt{(+7.2)}}} & \scriptsize{\textcolor{RoyalBlue}{\texttt{(+10.8)}}} & \scriptsize{\textcolor{RoyalBlue}{\texttt{(+12.3)}}} & \scriptsize{\textcolor{RoyalBlue}{\texttt{(+19.5)}}} & \scriptsize{\textcolor{RoyalBlue}{\texttt{(+18.2)}}} & \scriptsize{\textcolor{RoyalBlue}{\texttt{(+15.4)}}} \\
    \midrule
    \multicolumn{13}{c}{\textbf{PM\textsubscript{10}} (4-class AQI: \texttt{Good}, \texttt{Moderate}, \texttt{Bad}, \texttt{VeryBad})} \\
    \midrule
    Aurora & 17.5 & 22.3 & 31.9 & 46.0 & 53.1 & 52.0 & 48.4 & 35.1 & 31.1 & 28.7 & 24.1 & 26.1 \\
    \midrule
    FAKER-Air\textsubscript{SFT} & 50.7 & 45.7 & 48.0 & 55.0 & 53.7 & 53.2 & 56.6 & 55.2 & 49.8 & 47.3 & 47.3 & 49.7 \\
    \rowcolor{cyan!5}
    FAKER-Air\textsubscript{GRPO} & 44.9 & 44.9 & 48.3 & 55.3 & 55.0 & 55.5 & 58.9 & 51.6 & 47.8 & 49.0 & 44.7 & 45.9 \\ [-2pt]
    \rowcolor{cyan!5}
    & \scriptsize{\textcolor{RoyalBlue}{\texttt{(+27.4)}}} & \scriptsize{\textcolor{RoyalBlue}{\texttt{(+22.6)}}} & \scriptsize{\textcolor{RoyalBlue}{\texttt{(+16.4)}}} & \scriptsize{\textcolor{RoyalBlue}{\texttt{(+9.3)}}} & \scriptsize{\textcolor{RoyalBlue}{\texttt{(+1.9)}}} & \scriptsize{\textcolor{RoyalBlue}{\texttt{(+3.5)}}} & \scriptsize{\textcolor{RoyalBlue}{\texttt{(+10.5)}}} & \scriptsize{\textcolor{RoyalBlue}{\texttt{(+16.5)}}} & \scriptsize{\textcolor{RoyalBlue}{\texttt{(+16.7)}}} & \scriptsize{\textcolor{RoyalBlue}{\texttt{(+20.3)}}} & \scriptsize{\textcolor{RoyalBlue}{\texttt{(+20.6)}}} & \scriptsize{\textcolor{RoyalBlue}{\texttt{(+19.8)}}} \\
    \bottomrule
  \end{tabular}}
  \vspace{-0.5em}
\end{table*}

%% file: tables/sft_ablation_supp.tex
\begin{table*}[t!]
  \centering
  \caption{\textbf{Ablation Study on SFT for PM\textsubscript{2.5} and PM\textsubscript{10} on long lead time PM forecasting (6-hour intervals up to 120h on 2023).} Integrating OBS with CMAQ and extending temporal accumulation loss ($\mathcal{L}_{\mathrm{TA}}$) to T=4 consistently improves F1-scores across all horizons.}
  \vspace{-0.5em}
  \label{tab:f_score_ablation_split_6h_both}

  \renewcommand{\arraystretch}{1.3}
  \small

  \resizebox{\textwidth}{!}{
  \begin{tabular}{cccc|c|cccccccccccccccccccc}
    \toprule
    \noalign{\vskip 1pt}
    \multicolumn{25}{c}{\textbf{PM\textsubscript{2.5}}} \\
    \noalign{\vskip 1pt}
    \midrule
    OBS & CMAQ & $\mathcal{L}_{\mathrm{TA}}$~{\scriptsize(T=2)} & $\mathcal{L}_{\mathrm{TA}}$~{\scriptsize(T=4)} & Overall & +6h & +12h & +18h & +24h & +30h & +36h & +42h & +48h & +54h & +60h & +66h & +72h & +78h & +84h & +90h & +96h & +102h & +108h & +114h & +120h \\
    \midrule
    \multicolumn{4}{c|}{Aurora Air Pollution~\citep{bodnar2025foundation}} 
      & 16.06 & -- & 48.82 & -- & 40.07 & -- & 17.75 & -- & 15.00 & -- & 4.04 & -- & 4.01 & -- & 1.16 & -- & 1.72 & -- & 0.46 & -- & 1.05 \\
    \noalign{\vskip 2pt}\hline\hline\noalign{\vskip 2pt}
    \yesmark & \nomark & \nomark  & \nomark  
      & 50.74 & 70.03 & 61.53 & 56.35 & 53.83 & 51.57 & 50.55 & 49.87 & 49.47 & 49.03 & 48.79 & 48.57 & 48.47 & 48.27 & 48.15 & 48.01 & 47.95 & 47.82 & 47.76 & 47.68 & 47.65 \\
    \yesmark & \yesmark & \nomark  & \nomark  
      & 54.40 & 66.88 & 62.56 & 60.59 & 58.89 & 56.73 & 55.33 & 54.42 & 53.67 & 52.75 & 52.26 & 52.02 & 51.86 & 51.47 & 51.36 & 51.45 & 51.44 & 51.32 & 51.36 & 51.41 & 51.38 \\
    \yesmark & \yesmark & \yesmark & \nomark  
      & 56.32 & 70.57 & 67.91 & 65.08 & 63.59 & 61.15 & 59.62 & 58.12 & 57.00 & 55.58 & 54.67 & 53.53 & 52.86 & 52.20 & 51.85 & 51.15 & 50.74 & 50.43 & 50.27 & 49.76 & 49.47 \\
      \rowcolor{cyan!5}
    \yesmark & \yesmark & \nomark & \yesmark  
      & 59.90 & 69.92 & 67.53 & 65.50 & 64.48 & 62.75 & 61.83 & 61.00 & 60.39 & 59.51 & 59.05 & 58.59 & 58.23 & 57.71 & 57.44 & 57.15 & 56.94 & 56.65 & 56.54 & 56.29 & 56.21 \\ [-3pt]
    \rowcolor{cyan!5}
    &    &   &  & \scriptsize{\textcolor{RoyalBlue}{\texttt{(+9.16)}}} & \scriptsize{\textcolor{RoyalBlue}{\texttt{(+8.39)}}} & \scriptsize{\textcolor{RoyalBlue}{\texttt{(+6.00)}}} & \scriptsize{\textcolor{RoyalBlue}{\texttt{(+11.67)}}} & \scriptsize{\textcolor{RoyalBlue}{\texttt{(+10.65)}}} & \scriptsize{\textcolor{RoyalBlue}{\texttt{(+11.18)}}} & \scriptsize{\textcolor{RoyalBlue}{\texttt{(+10.26)}}} & \scriptsize{\textcolor{RoyalBlue}{\texttt{(+11.13)}}} & \scriptsize{\textcolor{RoyalBlue}{\texttt{(+10.52)}}} & \scriptsize{\textcolor{RoyalBlue}{\texttt{(+10.72)}}} & \scriptsize{\textcolor{RoyalBlue}{\texttt{(+10.26)}}} & \scriptsize{\textcolor{RoyalBlue}{\texttt{(+10.12)}}} & \scriptsize{\textcolor{RoyalBlue}{\texttt{(+9.76)}}} & \scriptsize{\textcolor{RoyalBlue}{\texttt{(+9.56)}}} & \scriptsize{\textcolor{RoyalBlue}{\texttt{(+9.29)}}} & \scriptsize{\textcolor{RoyalBlue}{\texttt{(+9.39)}}} & \scriptsize{\textcolor{RoyalBlue}{\texttt{(+9.18)}}} & \scriptsize{\textcolor{RoyalBlue}{\texttt{(+8.97)}}} & \scriptsize{\textcolor{RoyalBlue}{\texttt{(+8.86)}}} & \scriptsize{\textcolor{RoyalBlue}{\texttt{(+8.64)}}} & \scriptsize{\textcolor{RoyalBlue}{\texttt{(+8.56)}}}\\
      \noalign{\vskip 1pt}
    \toprule
    \noalign{\vskip 1pt}
    \multicolumn{25}{c}{\textbf{PM\textsubscript{10}}} \\
    \noalign{\vskip 1pt}
    \midrule
    OBS & CMAQ & $\mathcal{L}_{\mathrm{TA}}$~{\scriptsize(T=2)} & $\mathcal{L}_{\mathrm{TA}}$~{\scriptsize(T=4)} & Overall & +6h & +12h & +18h & +24h & +30h & +36h & +42h & +48h & +54h & +60h & +66h & +72h & +78h & +84h & +90h & +96h & +102h & +108h & +114h & +120h \\
    \midrule
    \multicolumn{4}{c|}{Aurora Air Pollution~\citep{bodnar2025foundation}} 
      & 4.73 & -- & 23.34 & -- & 14.08 & -- & 3.38 & -- & 1.78 & -- & 0.22 & -- & 0.23 & -- & 0.02 & -- & 0.04 & -- & 0.00 & -- & 0.00 \\
    \noalign{\vskip 2pt}\hline\hline\noalign{\vskip 2pt}
    \yesmark & \nomark & \nomark & \nomark  
      & 41.63 & 69.94 & 61.12 & 54.92 & 50.80 & 45.71 & 43.08 & 40.91 & 39.65 & 37.91 & 37.11 & 36.32 & 35.99 & 35.24 & 35.16 & 34.85 & 34.80 & 34.38 & 34.37 & 34.15 & 34.12 \\
    \yesmark & \yesmark & \nomark  & \nomark
      & 52.65 & 72.58 & 67.71 & 64.81 & 62.97 & 59.30 & 59.30 & 57.32 & 57.32 & 55.77 & 55.77 & 55.09 & 55.09 & 54.58 & 54.58 & 54.23 & 54.23 & 53.84 & 53.84 & 53.60 & 53.60 \\
    \yesmark & \yesmark & \yesmark  & \nomark
      & 54.60 & 71.17 & 66.68 & 63.86 & 62.17 & 59.44 & 57.92 & 56.27 & 55.16 & 53.80 & 52.79 & 51.61 & 50.94 & 50.29 & 49.86 & 49.15 & 48.87 & 48.47 & 48.13 & 47.52 & 47.36 \\
      \rowcolor{cyan!5}
    \yesmark & \yesmark & \nomark  & \yesmark
      & 57.32 & 69.72 & 66.65 & 64.44 & 63.01 & 61.16 & 59.96 & 58.85 & 57.97 & 56.81 & 56.07 & 55.43 & 54.91 & 54.33 & 53.96 & 53.59 & 53.39 & 53.14 & 52.97 & 52.61 & 52.46 \\ [-3pt]
    \rowcolor{cyan!5}
    &    &   &  & \scriptsize{\textcolor{RoyalBlue}{\texttt{(+15.69)}}} & \scriptsize{\textcolor{RoyalBlue}{\texttt{(+8.60)}}} & \scriptsize{\textcolor{RoyalBlue}{\texttt{(+5.53)}}} & \scriptsize{\textcolor{RoyalBlue}{\texttt{(+13.64)}}} & \scriptsize{\textcolor{RoyalBlue}{\texttt{(+12.21)}}} & \scriptsize{\textcolor{RoyalBlue}{\texttt{(+18.08)}}} & \scriptsize{\textcolor{RoyalBlue}{\texttt{(+16.88)}}} & \scriptsize{\textcolor{RoyalBlue}{\texttt{(+19.20)}}} & \scriptsize{\textcolor{RoyalBlue}{\texttt{(+18.32)}}} & \scriptsize{\textcolor{RoyalBlue}{\texttt{(+19.70)}}} & \scriptsize{\textcolor{RoyalBlue}{\texttt{(+18.96)}}} & \scriptsize{\textcolor{RoyalBlue}{\texttt{(+19.44)}}} & \scriptsize{\textcolor{RoyalBlue}{\texttt{(+18.92)}}} & \scriptsize{\textcolor{RoyalBlue}{\texttt{(+19.26)}}} & \scriptsize{\textcolor{RoyalBlue}{\texttt{(+18.80)}}} & \scriptsize{\textcolor{RoyalBlue}{\texttt{(+18.95)}}} & \scriptsize{\textcolor{RoyalBlue}{\texttt{(+18.59)}}} & \scriptsize{\textcolor{RoyalBlue}{\texttt{(+18.91)}}} & \scriptsize{\textcolor{RoyalBlue}{\texttt{(+18.60)}}} & \scriptsize{\textcolor{RoyalBlue}{\texttt{(+18.49)}}} & \scriptsize{\textcolor{RoyalBlue}{\texttt{(+18.34)}}}\\
      \noalign{\vskip 1pt}
    \toprule
  \end{tabular}}
\end{table*}